\PassOptionsToPackage{table,dvipsnames}{xcolor}
\documentclass[]{style}
\usepackage{amsmath,amssymb,amsfonts}
\usepackage{algorithmic}
\usepackage{textcomp}
\usepackage{xcolor}
\usepackage{graphicx}
\usepackage{booktabs}
\usepackage{multirow}
\usepackage{array}
\usepackage{url}
\usepackage{balance}
\usepackage[edges]{forest}
\usepackage{tikz}
\usetikzlibrary{trees,arrows.meta,calc,positioning,shadows.blur,shapes,fit,backgrounds}

\definecolor{hidden-draw}{RGB}{20,68,106}

\tikzstyle{my-box}=[
    rectangle, draw=hidden-draw, rounded corners,
    text opacity=1, minimum height=1.5em, minimum width=5em,
    inner sep=2pt, align=center, fill opacity=1, line width=0.8pt,
]
\tikzstyle{leaf-head}=[my-box, fill=gray!20, draw=gray!80, text=black]
\tikzstyle{leaf-methods}=[my-box, fill=cyan!15, draw=cyan!60!black, text=black]
\tikzstyle{leaf-datasets}=[my-box, fill=red!15, draw=red!60!black, text=black]
\tikzstyle{leaf-metrics}=[my-box, fill=orange!20, draw=orange!70!black, text=black]
\tikzstyle{modelnode-methods}=[my-box, fill=cyan!8, draw=cyan!50!black, text=black]
\tikzstyle{modelnode-datasets}=[my-box, fill=red!8, draw=red!50!black, text=black]
\tikzstyle{modelnode-metrics}=[my-box, fill=orange!10, draw=orange!60!black, text=black]
\usepackage{xcolor}
\usepackage{graphicx} 
\usepackage{longtable}
\usepackage{pdflscape}
\usepackage{makecell}
\usepackage{rotating}
\usepackage{adjustbox}
\usepackage{threeparttable}
\usepackage{wrapfig}
\usepackage{float}
\usepackage{amsthm}
\usepackage{subcaption}
\usepackage{enumitem}
\usepackage{pifont}
\usepackage[edges]{forest}
\usepackage{tikz}
\usetikzlibrary{arrows.meta,positioning,calc,shapes.symbols,shapes.geometric,shapes.misc}
\usepackage{tabularx}
\usepackage{xltabular}
\usepackage{svg}
\usepackage{fontawesome5}

\theoremstyle{plain}

\theoremstyle{remark}

\definecolor{datalinkcolor}{RGB}{220,110,160}

\newcommand{\hflink}[1]{\href{#1}{%
  \raisebox{-0.14em}{\includegraphics[height=0.82em]{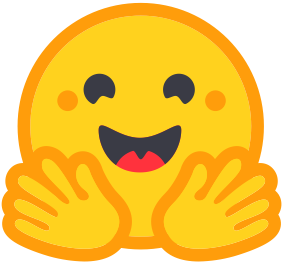}}\,%
  \textcolor{datalinkcolor}{\scriptsize Data}}}

\newcommand{\datalink}[1]{\href{#1}{%
  {\scriptsize\textcolor{datalinkcolor}{\faDatabase}}\,%
  \textcolor{datalinkcolor}{\scriptsize Data}}}
\newcommand{\figref}[1]{Fig.~\ref{#1}}
\newcommand{\tabref}[1]{Table~\ref{#1}}
\newcommand{\secref}[1]{Section~\ref{#1}}

\newcommand{\githublink}[1]{\href{#1}{%
  {\scriptsize\textcolor{black}{\faGithub}}\,%
  \textcolor{black}{\scriptsize Code}}}
\newcommand{\paperlink}[1]{\faIcon{book}\,\href{#1}{\scriptsize Paper}}
\newcommand{\missinglink}{\textcolor{gray}{--}}
\newcolumntype{L}[1]{>{\raggedright\arraybackslash}p{#1}}
\newcolumntype{Y}{>{\raggedright\arraybackslash}X}
\newcommand{\IEEEPARstart}[2]{#1#2}
\newcommand\blfootnote[1]{%
  \begingroup
  \renewcommand\thefootnote{}\footnote{#1}%
  \addtocounter{footnote}{-1}%
  \endgroup
}

\title{{\raisebox{-1.ex}{\includegraphics[height=2em]{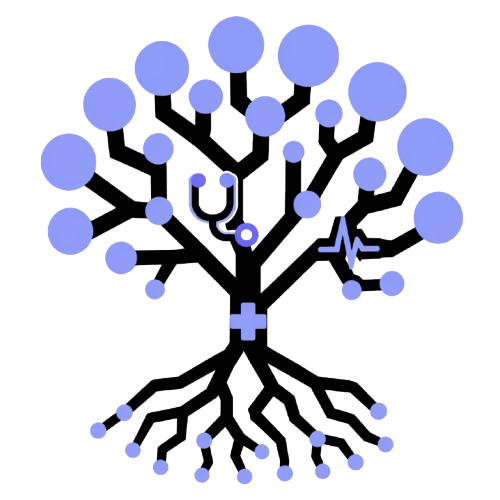}}}\; The Path to Self-Evolving Clinical Systems: Scaling Medical Agents from Assistance to Autonomy
}

\author{Chunzheng Zhu$^{1,*,\ddagger}$}
\author{Lei Tian$^{2,*}$}
\author{Bohan Tan$^{1,*}$}
\author{Ziqi Zhou$^{3,*}$}
\author{Yuxuan Sun$^{4,*}$}

\author{Yijun Wang$^{1}$}
\author{Chengchao Lv$^{1}$}
\author{Yilin Wen$^{2}$}
\author{Yijun He$^{2}$}
\author{Jinghao Lin$^{2}$}
\author{Yihang Chen$^{5}$}

\author{Chee Wei Tan$^{6}$}
\author{Qianshan Wei$^{7}$}
\author{Lei Zhao$^{8}$}
\author{Bin Pu$^{1}$}
\author{Kenli Li$^{1}$}

\author{Yuan Xue$^{9}$}
\author{Jianxin Lin$^{1,\dagger}$}

\affiliation{
\vspace{0.5em}
{\small$*$\;Equal Contribution\;\;$\dagger$\;Corresponding Author\;\;$\ddagger$\;Work done during internship.}

\vspace{0.5em}

{\small $^{1}$Hunan University,\;
$^{2}$ByteDance,\;
$^{3}$Duke University,\;
$^{4}$Westlake University,\;
$^{5}$The University of Hong Kong,\;
$^{6}$Nanyang Technological University,\;
$^{7}$Institute of Automation, Chinese Academy of Sciences,\;
$^{8}$University of Macau,\;
$^{9}$The Ohio State University}

\vspace{1em}

{\faIcon{github}}\,{\url{https://github.com/zhcz328/Awesome-Medical-Agents}}
}

\abstract{
The growing ability of large language models and vision-language models to jointly interpret and reason over images and text is reshaping medical imaging AI, moving it from passive task-specific predictors toward autonomous agents that perceive, reason, plan, remember, and act within real clinical environments. This survey deliberately departs from the capability-first narrative that dominates concurrent medical agent reviews. We begin from the realities of clinical deployment and ask a more basic question: what tasks, contamination-resistant benchmarks, and interactive training environments must a medical agent master before it can be trusted in practice? To enable systematic comparison across heterogeneous systems, we formalize such an agent as a sequential decision process under partial observability and propose a three-level autonomy taxonomy spanning assisted, cooperative, and fully autonomous operation. Around this foundation we organize the entire field along a single \emph{scaling spine} of three orthogonal axes: \emph{framework scaling} over architecture paradigms and tool orchestration, \emph{capability scaling} over the harness-driven loop that realizes perception to action, and \emph{environment scaling} that enriches the tool, data, and interface ecosystem. Within this spine we foreground \emph{clinical environment scaling}, the synthesis of rich tool and data ecosystems with clinical gyms, as the most immediately actionable yet least explored lever for agents that must natively inhabit PACS, EHR, and FHIR ecosystems. Crucially, we frame clinical self-evolution, the capacity of an agent to improve through environment interaction rather than parameter growth alone, as an aspirational frontier rather than a solved capability, and we transfer concrete lessons from general domain self-improving agents, agent gyms, and test-time compute scaling into the imaging setting. Applications across radiology, pathology, ophthalmology, and real hospital workflows are surveyed alongside deployment practice and open risks including hallucination, cascade failures, and fairness. By consolidating over $300$ references with particular emphasis on 2025--2026 advances in self-evolution and agentic environments, this survey charts a roadmap toward trustworthy, self-improving medical imaging systems that are genuinely ready for the clinic.
}

\metadata[Keywords]{Medical imaging agents, vision-language models, self-evolution, clinical scaling, autonomy taxonomy, agentic benchmarks, trustworthy clinical AI, survey.}

\begin{document}

\maketitle

\blfootnote{This survey will be actively maintained and updated together with its companion GitHub repository (\url{https://github.com/zhcz328/Awesome-Medical-Agents}). We welcome feedback, corrections, and contributions from the research community. If you identify relevant work that should be included or have suggestions for improving this survey or the repository, please submit an issue or pull request on GitHub, or contact us at \texttt{zhuchzh@hnu.edu.cn}.}

\onecolumn
\setcounter{tocdepth}{2}

{
\hypersetup{linkcolor=black}
\small

\renewcommand{\baselinestretch}{0.9}\selectfont  
\tableofcontents
}

\section{Introduction}\label{sec:intro}

\IEEEPARstart{T}{he} emergence of foundation models has brought broad, transferable representations to medical imaging, yet these models remain fundamentally passive: given an input they produce a single-pass output with no capacity for iterative reasoning, tool invocation, or contextual adaptation~\citep{zhang2023biomedclip, ma2024segment, lu2024visual}. The current paradigm shift is toward \emph{medical agents}, AI systems that fold perception, reasoning, planning, memory, tool use, and self-reflection into an autonomous loop that interacts with clinical environments~\citep{wang2024survey, xi2025rise, luo2025large, banerjie2025agentic}. Unlike static models that map inputs to outputs, such agents decompose complex clinical tasks, invoke specialized tools (e.g., segmentation models, report generators, retrieval databases), maintain working memory of patient history, and iteratively refine decisions from intermediate feedback~\citep{yao2022react, schick2023toolformer, shen2023hugginggpt}, mirroring the trajectory in natural language processing where LLM based agents~\citep{achiam2023gpt, touvron2023llama, grattafiori2024llama} showed that tool augmented reasoning could tackle tasks far beyond single-pass inference.

Several converging factors make medical agents both feasible and necessary. First, the maturation of multimodal large language models (MLLMs) such as \textsc{MedGemma}~\cite{sellergren2025medgemma}, \textsc{Lingshu}~\cite{xu2025lingshu}, \textsc{HealthGPT}~\cite{lin2025healthgpt}, \textsc{InternVL3}~\cite{zhu2025internvl3}, \textsc{Qwen3-VL}~\cite{bai2025qwen3}, and \textsc{RadVLM}~\cite{deperrois2025radvlm} supplies a reasoning backbone that jointly processes images and text~\cite{ye2025multimodal}, while standardized tool interfaces such as the Model Context Protocol~\cite{hou2025model, ehtesham2025survey, radosevich2025mcp} and function calling APIs~\cite{qin2024toolllm, patil2025berkeley} let this backbone orchestrate diverse specialist models. Second, core reasoning capabilities have advanced from single-shot perception toward more deliberate, self-corrective inference. Representative progress includes dynamic visual attention that lets agents acquire progressive diagnostic focus mimicking expert gaze, and causal reasoning with structured self-reflection that helps them identify and correct spurious visual correlations~\cite{zhu2026medeyes, lin2026medcausalx}. Together these trends indicate that trustworthy multi-step medical reasoning is becoming attainable rather than remaining aspirational. Third, clinical demand for integrated workflows, where a single consultation may require reading a CT scan, cross referencing prior studies, generating a structured report, and flagging follow up actions, calls for an agentic multistep paradigm that transcends what any individual model can provide~\cite{bluethgen2025agentic, dietrich2025agentic}. Fourth, the recently articulated principle of \emph{environment scaling}~\cite{fang2025towards, huang2025scaling} posits that agent capability can be amplified by enriching the tool and data environment rather than solely increasing model parameters~\cite{snell2024scaling, sardana2023beyond}, a uniquely scalable path for agents that operate in the tool rich clinical ecosystems exposed by PACS, EHR, and FHIR interfaces. We elevate this last principle from an incidental factor into one of the three organizing axes of the survey, the \emph{scaling spine} formalized in \secref{sec:spine}, and argue that the clinical environment itself is the most immediately actionable lever for advancing medical agent capabilities.

\textbf{Contributions of this survey.} This paper makes the following contributions:
\begin{itemize}[leftmargin=*,noitemsep]
\renewcommand\labelitemi{$\diamond$}
    \item We adopt a \emph{real environment first} organizing principle: rather than leading with a generic capability or per tool taxonomy as in concurrent medical agent surveys, we begin from the point of deployment, surveying the clinical scenarios, contamination-resistant benchmarks, and interactive training environments a real-world medical agent must master.
\item We formalize medical agents as six interacting cognitive modules operating under partial observability, and propose a three-level autonomy taxonomy that unifies the diverse literature.
\item We introduce a \emph{scaling spine} threading the entire survey, comprising \emph{framework scaling} (\secref{sec:arch}), \emph{capability scaling} (the harness-driven cognitive loop, \secref{sec:core}), and \emph{environment scaling} (\secref{sec:train}), presented upfront in \secref{sec:spine} as a global map.
    \item We elevate \emph{self-evolving systems} from a buried capability bullet to the destination of the spine, synthesizing general domain advances such as self-improving agents, agent gyms, and test-time compute scaling, and tracing their transfer into medical imaging.
    \item We review clinical applications from specialty departments to real hospital workflows, alongside deployment practice and a risk centered analysis spanning hallucination, cascade failures, and fairness.
\end{itemize}

\textbf{Relation to concurrent surveys.} Several recent surveys review medical agents from a capability centric angle, organizing the literature by cognitive module and by tool before mapping systems onto hospital departments, a skeleton that mirrors the generic agent architecture literature~\citep{zhang2025landscape, wang2024survey} and treats evaluation as a downstream appendix. Our survey departs in three ways. First, we make real environment scenarios, benchmarks, and interactive training environments the \emph{entry point} rather than the closing chapter, since real-world readiness is the operative question for medical agents. Second, we recast the per tool and per module breakdown as instances of a unifying scaling spine, reading tool use and capability modules through how they scale toward autonomy rather than duplicating prior taxonomies. Third, we foreground self-evolution and environment scaling, drawing transferable lessons from general domain self-improving agents, as the defining frontier for agents facing drifting clinical distributions. Across scope and methodology, representative recent surveys share one blind spot our positioning makes concrete: none couples a disclosed retrieval protocol with an environment centered scaling axis.

\textbf{Survey methodology and scope.} 
 Over January 2022 to June 2026, the window in which large language model driven agency became technically feasible for imaging, we queried major literature databases and the proceedings of leading computer vision, machine learning, and medical informatics venues, conjoining an \emph{agent} axis (\texttt{agent}, \texttt{agentic}, \texttt{tool use}, \texttt{multi-agent}, \texttt{autonomous}), a \emph{medical imaging} axis (\texttt{radiology}, \texttt{pathology}, \texttt{CT}, \texttt{MRI}, \texttt{X ray}, \texttt{report generation}), and a \emph{foundation model} axis (\texttt{large language model}, \texttt{foundation model}, \texttt{vision-language}). The search returned $1{,}174$ records and $942$ after de-duplication; three reviewers then independently screened titles and abstracts against a single inclusion criterion, an autonomous loop that perceives, reasons, and acts on an imaging or clinical task, with a fourth arbitrating disagreements, leaving over 300 studies that a snowball pass supplemented with the latest 2026 entries. We deliberately retain a few non imaging agents, namely electronic record operation and therapeutic reasoning, \emph{only} where they offer an architectural lesson that transfers to imaging, and flag such cases in \secref{sec:benchmarks} and \secref{sec:app}. Guided by this corpus, the survey unfolds along the scaling spine, from real environment scenarios and benchmarks through framework, capability, and environment scaling to self-evolving systems and their clinical deployment.


\section{Preliminaries and Taxonomy}\label{sec:prelim}


\subsection{Formal Definition of Medical Agents}
\label{sec:formal}

\begin{wrapfigure}{r}{0.5\textwidth}
    \centering
    \vspace{-40pt}
    \includegraphics[width=0.5\textwidth]{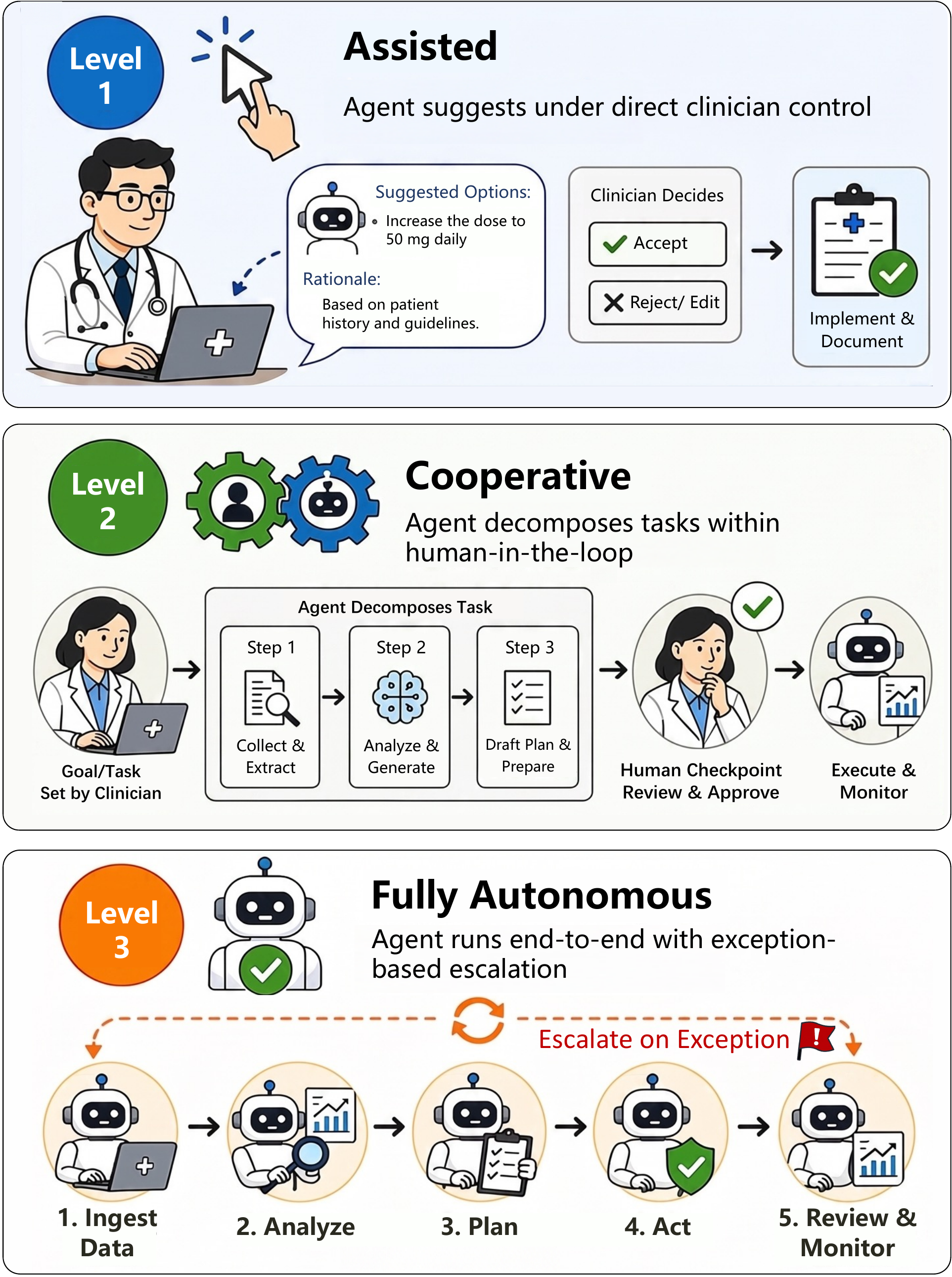}
\caption{\textbf{Three-level autonomy taxonomy for medical agents.} Level~1 (Assisted): the agent suggests under direct clinician control. Level~2 (Cooperative): it decomposes tasks and invokes tools within a human-in-the-loop framework. Level~3 (Fully Autonomous): it runs end-to-end workflows with exception-based human escalation.}
\label{fig:taxonomy}
    \vspace{-20pt}
\end{wrapfigure}
We define a \emph{medical agent} as a computational system $\mathcal{A} = (\mathcal{P}, \mathcal{R}, \mathcal{L}, \mathcal{M}, \mathcal{T}, \mathcal{F})$ composed of six functional modules: perception $\mathcal{P}$, reasoning $\mathcal{R}$, planning $\mathcal{L}$, memory $\mathcal{M}$, tool use $\mathcal{T}$, and action reflection $\mathcal{F}$~\citep{wang2024survey, xi2025rise, russell2010artificial}. A defining feature of this setting is \emph{partial observability}: the agent never observes the patient's underlying condition directly, but only indirect and incomplete evidence such as images, prior reports, and tool outputs, from which it must maintain and progressively refine an internal estimate of the true clinical state. The six modules jointly realize this process: perception and memory build and retain a belief over the latent condition, reasoning and planning turn that belief into actions such as tool calls and report operations, tool use executes them against the clinical environment, and action reflection revises the belief from returned feedback. This formulation makes explicit that medical agency is fundamentally a problem of acting under uncertainty and incomplete evidence rather than single-pass prediction.

These six modules are not an end in themselves but the per agent substrate that the \emph{capability} axis of the spine scales, while the environment $\mathcal{E}$ becomes the explicit lever of its third axis. Read through this partial-observability lens, the three scaling axes of the spine correspond to enlarging distinct components of the decision process: framework scaling widens the space of available actions by adding cooperating agents, capability scaling deepens the mapping from the agent's state estimate to its actions through a richer runtime harness, and environment scaling enriches both the evidence the agent can observe and the objectives it is scored against by connecting more tools, data sources, and interfaces, which is precisely why it is the most immediately actionable lever for agents that already inhabit tool rich clinical ecosystems.

\subsection{Three Level Autonomy Taxonomy}

Drawing on the autonomy levels established in autonomous driving~\citep{on2021taxonomy} and surgical robotics~\citep{yang2017medical,haidegger2019autonomy}, and informed by recent taxonomies of AI agent autonomy~\citep{feng2025levels}, we propose a three-level taxonomy for medical agents, shown in Fig.~\ref{fig:taxonomy}:

\textbf{Level~1: Assisted Autonomy.} At this level, the agent serves as an intelligent assistant that enhances clinician efficiency without independent decision making. The agent responds to explicit queries, generates candidate findings, or pre fills report templates, but all decisions require clinician confirmation~\citep{chen2024chexagent, chaves2024towards}. Examples include interactive segmentation tools with prompt based refinement~\citep{ma2024segment, liu2026medsam} and report drafting assistants~\citep{bannur2024maira}.

\textbf{Level~2: Cooperative Autonomy.} The agent proactively decomposes tasks, invokes specialized tools, and presents synthesized results but operates within a human-in-the-loop framework~\citep{xu2025multiagentreasoningsystemscollaborative, tang2024medagents}. The clinician retains override authority and reviews agent outputs before clinical action. Multiagent diagnostic pipelines that perform differential diagnosis and flag critical findings exemplify this level~\citep{kim2024mdagents, wang2025medagent, rose2025meddxagent, zhu2026medagentboard}.

\textbf{Level~3: Fully Autonomous Operation.} The agent independently executes end-to-end clinical workflows, from image acquisition quality assessment through diagnosis to report generation and follow up scheduling, with minimal human oversight, subject to regulatory guardrails and exception triggered human escalation~\citep{leong2026autonomous, wang2025systematic}. While no current system fully achieves Level~3, emerging frameworks such as autonomous radiology pipelines with closed-loop quality control~\citep{tzanis2025maistro, roschewitz2026radagent} and self-evolving agent ecosystems~\citep{fan2026evolving, shen2026evo} represent significant steps in this direction.

\subsection{The Scaling Spine: Framework, Capability, and Environment Scaling}\label{sec:spine}

The literature on medical agents is easy to enumerate but hard to organize, because architecture papers, cognitive module papers, and training papers each adopt their own vocabulary and rarely state how they compound. We therefore commit upfront to a single organizing axis, the \emph{scaling spine}, and read the entire survey through it. The spine asserts that the real-world readiness of an agent $\mathcal{A}$ is amplified along three logically distinct and largely independent directions, which we write compactly as:
\begin{equation}\label{eq:spine}
    \mathrm{Readiness}(\mathcal{A}) \;\approx\; f\big(\,\underbrace{N_{\text{topo}}}_{\text{framework}},\;\; \underbrace{D_{\text{loop}}}_{\text{capability}},\;\; \underbrace{|\mathcal{E}|}_{\text{environment}}\,\big),
\end{equation}
where $N_{\text{topo}}$ denotes the structural richness of the agent topology (number and specialization of cooperating agents and the orchestration that binds them), $D_{\text{loop}}$ denotes the depth of the cognitive loop wrapped around a fixed backbone (the runtime harness and the six modules it cycles), and $|\mathcal{E}|$ denotes the size and quality of the reachable environment (tools, data sources, and interaction interfaces). The three are complementary rather than substitutable, since a richer topology cannot compensate for a shallow loop, and neither helps an agent that cannot reach the PACS, EHR, or guideline databases it needs, which is precisely why we treat them as orthogonal axes of one spine rather than as competing schools.

\begin{tcolorbox}[
  colback=secblue!5,
  colframe=secblue!50,
  colbacktitle=secblue!50,
  coltitle=black,
  title={\textbf{\textcolor{secblue}{The Scaling Spine:} A Core Map of This Survey}},
  boxrule=5pt,
  arc=5pt,
  drop shadow,
  parbox=false,
  before skip=5pt,
  after skip=10pt,
  left=5pt,
  right=20pt,
]
\begin{itemize}[leftmargin=*,noitemsep]
\renewcommand\labelitemi{$\diamond$}
    \item \textbf{Framework scaling}: scales the topology $N_{\text{topo}}$, from a single tool augmented agent through role-based teams to orchestration topologies that mirror a tumour board's division of labor.
    \item \textbf{Capability scaling}: scales the loop depth $D_{\text{loop}}$, deepening the harness-driven perceive, reason, plan, act, and reflect cycle around a backbone of fixed size rather than enlarging the backbone itself.
    \item \textbf{Environment scaling}: scales the reachable environment $|\mathcal{E}|$, enriching tools, data, and interfaces, and realized over time through the adaptation ladder whose destination is a self-evolving agent.
\end{itemize}
\end{tcolorbox}

\begin{figure*}[!t]
\centering
\includegraphics[width=\textwidth]{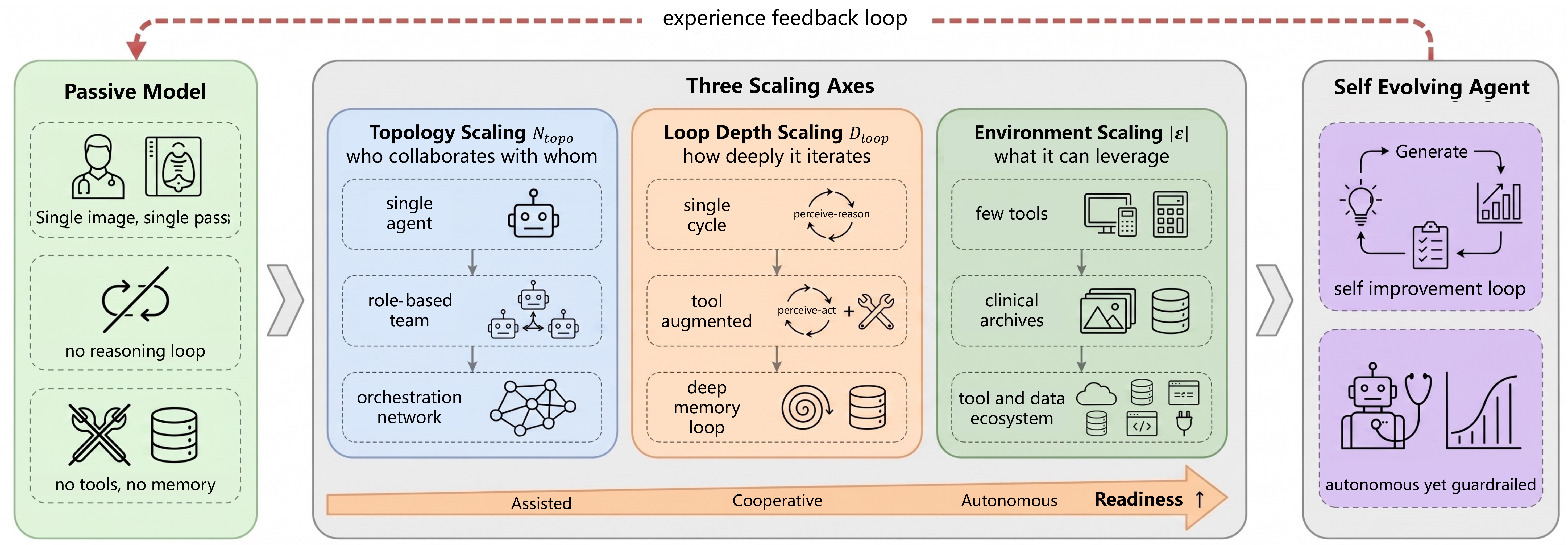}
\caption{\textbf{The scaling spine carries a passive single-pass model to a self-evolving medical agent along three orthogonal axes.} A passive foundation model (left) reads a single image in one forward pass with no reasoning loop, no tools, and no memory. Three complementary scaling axes then lift it toward autonomy: framework scaling $N_{\text{topo}}$ enriches \emph{who collaborates with whom} (single agent, role-based team, orchestration network); capability scaling $D_{\text{loop}}$ deepens \emph{how far it iterates} (single perceive-reason cycle, tool augmented perceive-act, deep memory loop); and environment scaling $|\mathcal{E}|$ widens \emph{what tools and data it can leverage} (few tools, clinical archives, a full tool and data ecosystem). Readiness rises from assisted through cooperative to autonomous as the three axes advance together, whereas scaling any single axis in isolation saturates. The destination (right) is a self-evolving agent that runs a self-improvement loop yet remains guardrailed, closed by an experience feedback loop that returns deployment outcomes as supervision. The axes are complementary rather than substitutable, which is why we treat them as one spine rather than competing schools.}
\label{fig:scaling}
\end{figure*}

The spine lifts a passive single-pass model toward autonomy along three orthogonal axes, which we summarize in \figref{fig:scaling} before elaborating each in turn. 
The first axis, \emph{framework scaling} (\secref{sec:arch}), grows capability by enriching the wiring among agents, moving from a single tool augmented agent to role-based multi-agent teams and finally to explicit orchestration topologies, the level at which a tumour board style division of labor becomes expressible. The second axis, \emph{capability scaling} (\secref{sec:core}), grows capability by deepening the cognitive loop around a backbone of fixed size, and our central claim there is that competence tracks the engineering of the perceive, reason, plan, act, and reflect loop more than the raw parameter count. The third axis, \emph{environment scaling} (\secref{sec:train}), grows capability by enlarging $\mathcal{E}$ itself, and it is the axis we foreground because medical agents already inhabit unusually tool rich clinical ecosystems, so that connecting a new specialist model, archive, or FHIR endpoint often buys more readiness per unit effort than another round of finetuning~\citep{fang2025towards, huang2025scaling, hou2025model}. Crucially, environment scaling is realized progressively over time through the adaptation ladder of \secref{sec:train}, from prompting through verifiable reward to self-play, whose destination is the \emph{self-evolving} agent that keeps expanding its own $\mathcal{E}$ in place. We deliberately distinguish this spine from the two conventional scaling laws it complements, namely parameter scaling and test-time compute scaling, both of which act on a single agent in isolation and neither of which captures the structural and environmental leverage that defines an agentic system in realistic clinical settings.

\section{Real Environment Clinical Scenarios, Benchmarks, and Training Environments}\label{sec:bench}

We open the technical body of this survey from the deployment destination rather than from the agent internals, because the operative question for medical agents is not whether a capability module exists but whether the assembled system survives contact with a real clinical environment~\citep{jiang2025medagentbench, schmidgall2024agentclinic, yan2026clinicallab}. This inversion is deliberate. A real environment imposes constraints that static accuracy metrics never expose: multiturn interaction with patients and ordering clinicians, partial and streaming observations from PACS and EHR, latency and cost budgets, distributional drift across sites and scanners, and the requirement that every action be auditable~\citep{bedi2025medhelm, arora2025healthbench, gu2025medagentaudit}. We therefore organize this section around three concentric layers that an agent must traverse on the way to deployment, namely the \emph{scenarios} it is expected to act in, the \emph{benchmarks} that measure whether it acts correctly, and the \emph{training environments} (clinical gyms and simulators) in which it can practice and improve before touching a patient. Crucially, these three layers are not a closing appendix to a capability catalog but the scaffolding from which the rest of the survey reads back: architecture, cognition, and training all exist to close the gap between an agent and the environment defined here.

\subsection{Real Environment Clinical Scenarios}\label{sec:scenarios}

A useful taxonomy of medical agent scenarios is organized by how tightly the agent is coupled to the live clinical loop rather than by anatomical specialty~\citep{schmidgall2024agentclinic, yan2026clinicallab, jiang2025medagentbench}. At one end sit \emph{single shot consultative} scenarios, in which the agent answers a bounded imaging query (a visual question, a finding classification, a draft report sentence) with the clinician fully in control~\citep{lau2018dataset, he2020pathvqa}. The middle band comprises \emph{interactive multiturn} scenarios, where the agent must request additional views, query prior studies, reconcile contradictory evidence, and explain its reasoning over several turns, mirroring how a radiologist or an MDT actually works~\citep{schmidgall2024agentclinic, almansoori2025self, feng2026doctoragent}. At the far end lie \emph{streaming workflow} scenarios, in which the agent participates in a continuously running hospital process: real-time triage and reprioritization of worklists for critical findings such as pulmonary embolism or intracranial hemorrhage~\citep{khosravi2026agentic}, longitudinal follow up across imaging time points~\citep{yu2026agenticmemorylearningunified, sun2026linking}, and end-to-end report to order pipelines under FHIR~\citep{jiang2025medagentbench, shi2024ehragent}. The defining property of the right hand scenarios, and the reason they are underserved by conventional benchmarks, is that the environment changes in response to the agent's own actions, so correctness is a property of an interaction trajectory rather than of a single output. This action conditioned, partially observable structure is exactly what our formulation in \secref{sec:prelim} is designed to capture, and it sets the requirements that the architectures (\secref{sec:arch}), capabilities (\secref{sec:core}), and training regimes (\secref{sec:train}) must satisfy.

\begin{figure}[t]
\centering
\includegraphics[width=\columnwidth] {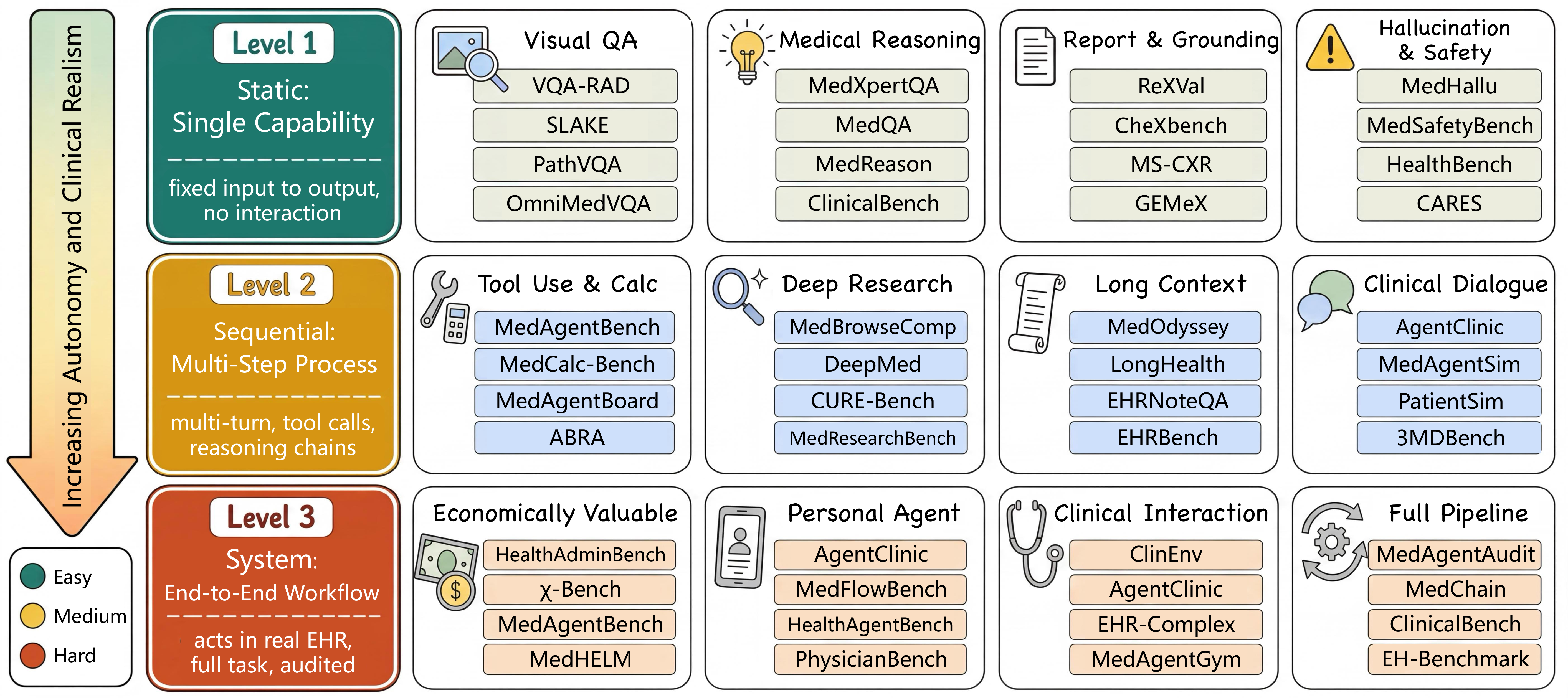}
\caption{\textbf{Benchmark taxonomy for medical agent evaluation.} We organize representative benchmarks along three levels of increasing autonomy and clinical realism: Level~1 (static, single-capability) tests isolated perception and reasoning under clean inputs; Level~2 (sequential, multi-step) requires tool invocation, evidence retrieval, long-context reasoning, and sustained clinical dialogue; Level~3 (system, end-to-end workflow) demands autonomous operation inside real or simulated EHR environments with full-pipeline auditing.}
\label{fig:benchmarks}
\end{figure}

\begin{table*}[htpb]
\centering
\caption{\textbf{Comprehensive Catalog of Benchmarks for Medical Agent Readiness.}
Benchmarks organized along three levels of increasing autonomy and clinical realism.}
\label{tab:benchmarks}
\scriptsize
\setlength{\tabcolsep}{1.4pt}
\renewcommand{\arraystretch}{1}
\begin{tabularx}{\textwidth}{@{}L{0.8cm}L{3.1cm}YL{1.9cm}L{1.0cm}L{1.0cm}@{}}
\toprule
\textbf{Year} & \textbf{Benchmark} & \textbf{Key Contribution} & \textbf{Venue} & \textbf{Paper} & \textbf{Data/Code} \\
\midrule

\multicolumn{6}{l}{\textbf{Level 1: Isolated Single Capabilities}} \\
\midrule

2018 & VQA-RAD~\citep{lau2018dataset}
& Clinician generated naturalistic radiology visual question answering
& Sci. Data
& \paperlink{https://www.nature.com/articles/sdata2018251}
& \hflink{https://huggingface.co/datasets/flaviagiammarino/vqa-rad} \\

2019 & PubMedQA~\citep{jin2019pubmedqa}
& Biomedical yes no maybe reasoning over abstracts
& EMNLP 2019
& \paperlink{https://arxiv.org/abs/1909.06146}
& \hflink{https://huggingface.co/datasets/qiaojin/PubMedQA} \\

2020 & PathVQA~\citep{he2020pathvqa}
& Pathology VQA, 32K questions over 5K images
& arXiv
& \paperlink{https://arxiv.org/abs/2003.10286}
& \hflink{https://huggingface.co/datasets/flaviagiammarino/path-vqa} \\

2021 & MedQA~\citep{jin2021disease}
& Free form licensing exam open domain question answering
& AAAI 2021
& \paperlink{https://arxiv.org/abs/2009.13081}
& \hflink{https://huggingface.co/datasets/GBaker/MedQA-USMLE-4-options} \\

2021 & SLAKE~\citep{liu2021slake}
& Bilingual knowledge enhanced radiology VQA with semantic labels
& ISBI 2021
& \paperlink{https://arxiv.org/abs/2102.09542}
& \hflink{https://huggingface.co/datasets/mdwiratathya/SLAKE-vqa-english} \\

2022 & MedMCQA~\citep{pal2022medmcqa}
& Multisubject entrance exam medical multiple choice QA
& CHIL 2022
& \paperlink{https://arxiv.org/abs/2203.14371}
& \hflink{https://huggingface.co/datasets/openlifescienceai/medmcqa} \\

2022 & MS-CXR~\citep{boecking2022making}
& Radiologist phrase grounding boxes for chest radiographs
& ECCV 2022
& \paperlink{https://arxiv.org/abs/2204.09817}
& \datalink{https://physionet.org/content/ms-cxr} \\

2023 & ReXVal~\citep{yu2023radiology}
& Radiologist error annotations for report evaluation
& Patterns
& \paperlink{https://physionet.org/content/rexval-dataset/1.0.0/}
& \datalink{https://physionet.org/content/rexval-dataset} \\

2024 & CheXbench~\citep{chen2024chexagent}
&  Chest X-ray perception and report understanding, 8 tasks
& arXiv
& \paperlink{https://arxiv.org/abs/2401.12208}
& \githublink{https://github.com/Stanford-AIMI/CheXagent} \\

2024 & OmniMedVQA~\citep{hu2024omnimedvqa}
& Large scale medical VQA, 118K images, 12 modalities
& CVPR 2024
& \paperlink{https://arxiv.org/abs/2402.09181}
& \hflink{https://huggingface.co/datasets/foreverbeliever/OmniMedVQA} \\

2024 & MedSafetyBench~\citep{han2024medsafetybench}
& Medical safety benchmark, 1{,}800 harmful request demonstrations
& NeurIPS 2024
& \paperlink{https://arxiv.org/abs/2403.03744}
& \githublink{https://github.com/AI4LIFE-GROUP/med-safety-bench} \\

2024 & CARES~\citep{xia2024cares}
& Med LVLM trustworthiness across five reliability dimensions
& NeurIPS 2024
& \paperlink{https://arxiv.org/abs/2406.06007}
& \githublink{https://github.com/richard-peng-xia/CARES} \\

2024 & ClinicalBench~\citep{chen2024clinicalbench}
& Tests whether LLMs surpass classical clinical prediction models
& arXiv
& \paperlink{https://arxiv.org/abs/2411.06469}
& \githublink{https://github.com/canyuchen/ClinicalBench} \\

2025 & MedXpertQA~\citep{zuo2025medxpertqa}
& Expert level reasoning, 4{,}460 text and multimodal questions
& ICML 2025
& \paperlink{https://arxiv.org/abs/2501.18362}
& \hflink{https://huggingface.co/datasets/TsinghuaC3I/MedXpertQA} \\

2025 & MedHallu~\citep{pandit2025medhallu}
& First medical hallucination detection benchmark, 10K pairs
& arXiv
& \paperlink{https://arxiv.org/abs/2502.14302}
& \hflink{https://huggingface.co/datasets/UTAustin-AIHealth/MedHallu} \\

2025 & HealthBench~\citep{arora2025healthbench}
& Physician rubric graded multiturn health conversations, 5K cases
& arXiv
& \paperlink{https://arxiv.org/abs/2505.08775}
& \hflink{https://huggingface.co/datasets/openai/healthbench} \\

2025 & MedReason~\citep{wu2025medreason}
&  KG-grounded medical reasoning chains, 32K QA
& arXiv
& \paperlink{https://arxiv.org/abs/2504.00993}
& \githublink{https://github.com/UCSC-VLAA/MedReason} \\

2025 & GEMeX~\citep{liu2025gemex}
& Grounded, explainable chest X-ray VQA (1.6M questions) 
& ICCV 2025
& \paperlink{https://arxiv.org/abs/2411.16778}
& \hflink{https://www.med-vqa.com/GEMeX} \\

2026 & RADAR~\citep{sun2026radar}
& Multimodal 3D image based radiology report discrepancy review
& arXiv
& \paperlink{https://arxiv.org/abs/2603.06681}
& \missinglink \\

\midrule
\multicolumn{6}{l}{\textbf{Level 2: Tool Augmented and Interactive Agency}} \\
\midrule

2024 & AgentClinic~\citep{schmidgall2024agentclinic}
& Multimodal multiagent simulated clinical encounters with tools
& ICLR 2025
& \paperlink{https://arxiv.org/abs/2405.07960}
& \githublink{https://github.com/SamuelSchmidgall/AgentClinic} \\

2024 & MedCalc-Bench~\citep{khandekar2024medcalc}
& Benchmarks LLMs on clinical calculator medical computation
& NeurIPS 2024
& \paperlink{https://arxiv.org/abs/2406.12036}
& \hflink{https://huggingface.co/datasets/ncbi/MedCalc-Bench-v1.0} \\

2024 & EHRNoteQA~\citep{kweon2024ehrnoteqa}
& 962 QA pairs over MIMIC-IV discharge summaries, 8 clinical topics
& NeurIPS 2024
& \paperlink{https://arxiv.org/abs/2402.16040}
& \datalink{https://physionet.org/content/ehr-notes-qa-llms} \\

2025 & MedAgentBench~\citep{jiang2025medagentbench}
& FHIR compliant virtual EHR, 300 clinician authored agentic tasks
& arXiv
& \paperlink{https://arxiv.org/abs/2501.14654}
& \githublink{https://github.com/stanfordmlgroup/medagentbench} \\

2025 & MedR-Bench~\citep{qiu2025quantifying}
& 1{,}453 real cases quantifying medical reasoning quality
& Nat. Commun.
& \paperlink{https://arxiv.org/abs/2503.04691}
& \githublink{https://github.com/MAGIC-AI4Med/MedRBench} \\

2025 & MedAgentsBench~\citep{tang2025medagentsbench}
& Hard multistep clinical reasoning for thinking models
& arXiv
& \paperlink{https://arxiv.org/abs/2503.07459}
& \hflink{https://huggingface.co/datasets/super-dainiu/MedicalAgentsBench} \\

2025 & 3MDBench~\citep{Sviridov_2025}
& Multimodal multiagent doctor patient dialogue, 34 diagnoses
& arXiv
& \paperlink{https://arxiv.org/abs/2504.13861}
& \hflink{https://huggingface.co/datasets/univanxx/3mdbench} \\

2025 & MedAgentBoard~\citep{zhu2026medagentboard}
& Multiagent versus single LLM across four clinical tasks
& NeurIPS 2025
& \paperlink{https://arxiv.org/abs/2505.12371}
& \githublink{https://github.com/yhzhu99/medagentboard} \\

2025 & MedBrowseComp~\citep{chen2025medbrowsecomp}
& First multihop medical web browsing agent benchmark
& arXiv
& \paperlink{https://arxiv.org/abs/2505.14963}
& \hflink{https://huggingface.co/datasets/AIM-Harvard/MedBrowseComp} \\

2025 & BFCL~\citep{patil2025berkeley}
& Live executable benchmark for function and tool calling
& ICML 2025
& \paperlink{https://openreview.net/forum?id=2GmDdhBdDk}
& \hflink{https://huggingface.co/datasets/gorilla-llm/Berkeley-Function-Calling-Leaderboard} \\

2025 & MedOdyssey~\citep{fan2025medodyssey}
& Medical long context benchmark up to 200K tokens, 7 length levels
& NAACL 2025
& \paperlink{https://arxiv.org/abs/2406.15019}
& \githublink{https://github.com/JOHNNY-fans/MedOdyssey} \\

2025 & LongHealth~\citep{adams2025longhealth}
& 400 questions over 20 long fictional patient clinical documents
& J. Healthc. Inform. Res.
& \paperlink{https://arxiv.org/abs/2401.14490}
& \githublink{https://github.com/kbressem/LongHealth} \\

2025 & R2MED~\citep{zhang2025r2med}
&  Reasoning-based medical retrieval benchmark (876 queries, 3 tasks)
& arXiv
& \paperlink{https://arxiv.org/abs/2505.14558}
& \githublink{https://github.com/R2MED/R2MED} \\

2025 & CURE-Bench~\citep{cofala2025medai}
&     Therapeutic agent reasoning evaluation
& arXiv
& \paperlink{https://arxiv.org/abs/2512.11682}
& \githublink{https://github.com/mims-harvard/CURE-Bench} \\

2025 & MedAgentSim~\citep{almansoori2025self}
& Self-evolving multi-agent clinical simulation with dialogue
& MICCAI 2025
& \paperlink{https://arxiv.org/abs/2503.22678}
& \githublink{https://github.com/MAXNORM8650/MedAgentSim} \\

2025 & PatientSim~\citep{kyung2026patientsim}
& Persona driven patient simulator, 37 personas from MIMIC-IV
& NeurIPS 2025
& \paperlink{https://arxiv.org/abs/2505.17818}
& \githublink{https://github.com/dek924/PatientSim} \\

2026 & MedDialogRubrics~\citep{gong2026meddialogrubrics}
& Multiturn diagnostic consultation, 5{,}200 cases, 60K+ rubrics
& arXiv
& \paperlink{https://arxiv.org/abs/2601.03023}
& \missinglink \\

2026 & AgentRx~\citep{jorf2026agentrx}
& Single vs multi-agent LLMs for multimodal clinical prediction
& arXiv
& \paperlink{https://arxiv.org/abs/2605.10286}
& \missinglink \\

2026 & ABRA~\citep{maksudov2026abra}
& Radiology agent with OHIF/Orthanc, 21 tools, 655 tasks
& arXiv
& \paperlink{https://arxiv.org/abs/2605.11224}
& \githublink{https://github.com/Luab/ABRA} \\

2026 & AutoMedBench~\citep{liu2026automedbench}
& Workflow aware benchmark for autonomous medical AI research, five stage agentic pipeline over imaging and multimodal tasks
& arXiv
& \paperlink{https://arxiv.org/abs/2606.01961}
& \githublink{https://github.com/automedbench/automedbench} \\

2026 & DeepMed~\citep{wang2026deepmed}
& Medical deep research agent via multi-hop med-search, turn-controlled training
& ACL 2026 Findings
& \paperlink{https://aclanthology.org/2026.findings-acl.904}
& \missinglink \\

2026 & MedResearchBench~\citep{tan2026medresearchbench}
& AI research agents on clinical medical research, multi-domain tasks
& medRxiv
& \paperlink{https://www.medrxiv.org/content/10.64898/2026.03.30.26349749v1}
& \missinglink \\

2026 & EHRBench~\citep{xie2026ehrbench}
& Automated EHR grounded clinical decision-making benchmark
& arXiv
& \paperlink{https://arxiv.org/abs/2605.30637}
& \missinglink \\

\midrule
\multicolumn{6}{l}{\textbf{Level 3: System Level Clinical Operation}} \\
\midrule

2024 & EHRAgent~\citep{shi2024ehragent}
& Code empowered agent for multitabular EHR reasoning
& EMNLP 2024
& \paperlink{https://arxiv.org/abs/2401.07128}
& \githublink{https://github.com/wshi83/EhrAgent} \\

2025 & ClinicalLab~\citep{yan2026clinicallab}
& Multidepartmental benchmark, 24 departments, 150 diseases
& NeurIPS 2025
& \paperlink{https://arxiv.org/abs/2406.13890}
& \githublink{https://github.com/WeixiangYAN/ClinicalLab} \\

2025 & MedChain~\citep{liu2026medchain}
& Sequential clinical workflow benchmark, 12K cases, five stages
& NeurIPS 2025
& \paperlink{https://arxiv.org/abs/2412.01605}
& \githublink{https://github.com/ljwztc/MedChain} \\

2025 & MedHELM~\citep{bedi2025medhelm}
& Holistic medical LLM eval, 121 clinician validated tasks
& arXiv
& \paperlink{https://arxiv.org/abs/2505.23802}
& \githublink{https://github.com/stanford-crfm/helm} \\

2025 & MedAgentGym~\citep{xu2025medagentgym}
& Code based medical reasoning training env, 72K tasks
& arXiv
& \paperlink{https://arxiv.org/abs/2506.04405}
& \githublink{https://github.com/wshi83/MedAgentGym} \\

2025 & FHIR-AgentBench$^{\dagger}$~\citep{lee2025fhir}
& Agents on 2{,}931 HL7 FHIR interoperable EHR tasks
& arXiv
& \paperlink{https://arxiv.org/abs/2509.19319}
& \githublink{https://github.com/glee4810/FHIR-AgentBench} \\

2025 & MedAgentAudit~\citep{gu2025medagentaudit}
& Audits collaborative failure modes in medical multiagent system s
& arXiv
& \paperlink{https://arxiv.org/abs/2510.10185}
& \githublink{https://github.com/MedX-PKU/MedAgentAudit} \\

2026 & MedOpenClaw~\citep{shen2026medopenclaw}
& Auditable full study agents over uncurated clinical imaging
& arXiv
& \paperlink{https://arxiv.org/abs/2603.24649}
& \missinglink \\

2026 & MedFlowBench~\citep{shen2026medopenclawmedflowbenchauditingmedical}
& Full-study VLM benchmark for imaging viewers (3D Slicer, QuPath) and radiology/pathology workflows
& arXiv
& \paperlink{https://arxiv.org/abs/2603.24649}
& \missinglink \\

2026 & PhysicianBench~\citep{liu2026physicianbench}
& Agents in real EHR, long-horizon tasks, 670 grounded checkpoints
& arXiv
& \paperlink{https://arxiv.org/abs/2605.02240}
& \githublink{https://github.com/HealthRex/PhysicianBench} \\

2026 & CHI-Bench$^{\dagger}$~\citep{chen2026chi}
& End to end long-horizon policy rich healthcare workflow automation
& arXiv
& \paperlink{https://arxiv.org/abs/2605.16679}
& \githublink{https://github.com/actava-ai/chi-bench} \\

2026 & {\fontsize{8}{9.5}\selectfont
MedCase-Structured$^{\dagger}$~\citep{muti2026medcase}}
& Text-to-FHIR dataset for reasoning in realistic structured EHR
& ICML 2026 WS
& \paperlink{https://arxiv.org/abs/2605.30295}
& \hflink{https://huggingface.co/datasets/system-technologies/MedCase-Structured} \\

2026 & ClinEnv~\citep{lu2026clinenv}
& Interactive long-horizon inpatient EHR simulation
& arXiv
& \paperlink{https://arxiv.org/abs/2606.02568}
& \missinglink \\

2026 & HealthAdminBench~\citep{bedi2026healthadminbench}
& Computer-use agents on 135 healthcare administration tasks, 4 GUIs
& arXiv
& \paperlink{https://arxiv.org/abs/2604.09937}
& \githublink{https://github.com/som-shahlab/health-admin-bench} \\

2026 & HealthAgentBench~\citep{liu2026healthagentbench}
& 54 agentic healthcare tasks, 7 environments across patient journey
& arXiv
& \paperlink{https://arxiv.org/abs/2606.31179}
& \githublink{https://microsoft.github.io/HealthAgentBench} \\

2026 & EHR-Complex~\citep{qiao2026ehr}
& Interactive clinical database reasoning on MIMIC-IV, 52K tasks
& arXiv
& \paperlink{https://arxiv.org/abs/2606.23301}
& \missinglink \\

\bottomrule
\end{tabularx}
\end{table*}

\subsection{Benchmarks for Real World Readiness}
\label{sec:benchmarks}

To gauge how far medical agents are from deployment, we organize existing benchmarks along an axis of increasing autonomy and clinical realism, spanning three levels from isolated single capabilities, through tool augmented and dialogue driven agentic behavior, to full system level clinical operation, as summarized in \figref{fig:benchmarks} and detailed in \tabref{tab:benchmarks}. This layering matters because a model that scores well on a static perception probe may still collapse once it must call tools, retain long records, or close a multi-turn clinical loop, so readiness should be read level by level rather than from any single aggregate score, and the rapid arrival of 2026 benchmarks at the upper levels shows that the field is actively migrating its measurement effort from isolated competence toward sustained clinical operation.

At Level 1 the focus is whether a model masters isolated competence under clean inputs. Visual understanding is probed by VQA-RAD~\cite{lau2018dataset}, SLAKE~\cite{liu2021slake}, PathVQA~\cite{he2020pathvqa} and OmniMedVQA~\cite{hu2024omnimedvqa}, which together cover radiology, pathology and cross modality question answering. Knowledge grounded reasoning is measured by MedXpertQA~\cite{zuo2025medxpertqa}, MedQA~\cite{jin2021disease}, MedReason~\cite{wu2025medreason} and ClinicalBench~\cite{chen2024clinicalbench}, ranging from licensing style examinations to knowledge graph grounded reasoning chains and a direct test of whether language models can rival classical clinical prediction models. Report quality and visual grounding are assessed by ReXVal~\cite{yu2023radiology}, CheXbench~\cite{chen2024chexagent}, MS-CXR~\cite{boecking2022making} and GEMeX~\cite{liu2025gemex}, which move evaluation beyond surface overlap toward radiologist adjudicated error grading, phrase-level box grounding and large-scale groundable and explainable chest X-ray question answering, and the recent RADAR~\cite{sun2026radar} pushes this further by pairing a 3D study with a preliminary report and candidate edits so that a model is judged on whether it can flag genuine report discrepancies rather than merely describe an image. Crucially, trustworthiness is stressed here through MedHallu~\cite{pandit2025medhallu}, MedSafetyBench~\cite{han2024medsafetybench}, the safety split of HealthBench~\cite{arora2025healthbench} and CARES~\cite{xia2024cares}, since a hallucinated finding is more dangerous than a missed one in clinical practice.

Level 2 raises the bar from knowing to acting, asking whether an agent can wield external tools, search for evidence, sustain long context and hold a clinical dialogue. Tool use and quantitative competence are evaluated by MedAgentBench~\cite{jiang2025medagentbench}, MedCalc-Bench~\cite{khandekar2024medcalc}, MedAgentBoard~\cite{zhu2026medagentboard} and the general-purpose BFCL~\cite{patil2025berkeley}, which expose whether a model can invoke calculators and structured APIs rather than guess, while the 2026 ABRA benchmark~\cite{maksudov2026abra} makes this concrete for imaging by driving an OHIF viewer and an Orthanc DICOM server through twenty one function calling tools. Evidence seeking and research level synthesis are captured by MedBrowseComp~\cite{chen2025medbrowsecomp}, DeepMed~\cite{wang2026deepmed}, MedResearchBench~\cite{tan2026medresearchbench} and the reasoning driven retrieval benchmark R2MED~\cite{zhang2025r2med}, which test whether an agent can locate, filter and synthesize clinical evidence under realistic web and literature noise and even carry out end-to-end medical research tasks. Long context fidelity, essential for reading an entire patient history, is probed by the medical specific MedOdyssey~\cite{fan2025medodyssey}, LongHealth~\cite{adams2025longhealth}, EHRNoteQA~\cite{kweon2024ehrnoteqa} and EHRBench~\cite{xie2026ehrbench}, which reveal sharp degradation once an agent must reason over full discharge summaries and longitudinal records rather than short snippets. Interactive clinical competence is then measured by AgentClinic~\cite{schmidgall2024agentclinic}, the self-evolving simulation MedAgentSim~\cite{almansoori2025self}, the persona driven simulator PatientSim~\cite{kyung2026patientsim} and 3MDBench~\cite{Sviridov_2025}, and the 2026 MedDialogRubrics suite~\cite{gong2026meddialogrubrics} sharpens this evaluation with 5{,}200 patient cases and more than sixty thousand fine-grained rubrics, so that multi-turn diagnostic dialogue is graded against explicit clinical criteria rather than a single reference answer.

Level 3 targets system level realism, asking whether an agent can operate autonomously inside the messy environment of real clinical and economic workflows. Economically valuable, deployment facing capability is quantified by HealthAdminBench~\cite{bedi2026healthadminbench}, the long-horizon policy rich $\chi$-Bench~\cite{chen2026chi}, the FHIR compliant MedAgentBench~\cite{jiang2025medagentbench} and the holistic MedHELM~\cite{bedi2025medhelm}, which connect raw capability to administrative, economic and clinician validated utility. Personal and longitudinal agency is stressed by AgentClinic~\cite{schmidgall2024agentclinic}, the full study imaging workflow benchmark MedFlowBench~\cite{shen2026medopenclawmedflowbenchauditingmedical}, the unified agentic healthcare suite HealthAgentBench~\cite{liu2026healthagentbench} and PhysicianBench~\cite{liu2026physicianbench}, which ground long-horizon tasks in real systems with execution verified checkpoints. Closed loop clinical interaction with live records and sandboxes is measured by ClinEnv~\cite{lu2026clinenv}, AgentClinic~\cite{schmidgall2024agentclinic}, the interactive database benchmark EHR-Complex~\cite{qiao2026ehr} and the code-based training environment MedAgentGym~\cite{xu2025medagentgym}, in which an agent must consult sub agents, query large clinical databases and execute SQL or Python against a MIMIC-IV substrate. Full pipeline operation and auditing across an entire patient encounter are evaluated by MedAgentAudit~\cite{gu2025medagentaudit}, the five stage sequential workflow MedChain~\cite{liu2026medchain}, ClinicalBench~\cite{chen2024clinicalbench} and the ophthalmic hallucination oriented EH-Benchmark~\cite{pan2025eh}, which probe traceable behavior and error propagation across staged clinical decision making. The persistent and widening gap from Level 1 to Level 3, made starker by the 2026 long-horizon benchmarks, indicates that current systems remain competent perceivers yet immature operators, motivating the multi-axis evaluation we develop next.

\subsection{Evaluation Metrics Beyond Accuracy}
\label{sec:metrics}
A persistent gap between leaderboard numbers and bedside trust is that most medical agent benchmarks still collapse performance into a single accuracy or exact match score, whereas real deployment failures rarely look like a wrong final answer alone. We therefore organize evaluation along five complementary axes that together characterize whether an agent is ready for the real world. The first axis is reasoning quality, which asks not only whether the conclusion is correct but whether the intermediate trajectory is faithful, grounded, and free of unsupported leaps. Recent work that audits reasoning traces step by step shows that final answer correctness routinely coexists with logically broken or hallucinated intermediate steps, so a faithful trajectory must be scored as a first-class quantity rather than assumed from the outcome \citep{lee2025evaluating}. In the clinical setting this matters acutely, because a report that reaches a plausible impression through an unsupported observation is more dangerous than an obviously wrong one, and metrics such as factual entailment over extracted findings \citep{ostmeier2024green, jain2021radgraph} and expert adjudicated error grading \citep{yu2023radiology} were introduced precisely to expose this failure mode.

The second axis is efficiency, which captures the cost of reaching an answer in tool calls, tokens, wall clock latency, and monetary spend. As agents are allowed to think longer and call more tools, accuracy and cost no longer move together: a careful study of how to allocate test-time compute across an agent rollout shows that naive scaling of steps yields diminishing and sometimes negative returns once redundant tool calls and repeated self-verification dominate \citep{zhu2025scaling}. A deployment metric thus reports accuracy at a fixed budget and the marginal accuracy per tool call, rather than the unconstrained ceiling. The third axis is robustness, meaning stability under distribution shift, adversarial or noisy inputs, and missing context, which cross-site and stress-oriented benchmarks make measurable \citep{xia2024cares, hu2024omnimedvqa}. The fourth axis is user experience, covering the agent’s ability to ask clarifying questions, to defer when uncertain, and to communicate calibrated confidence, properties that simulation and interaction benchmarks surface far better than static question answering \citep{schmidgall2024agentclinic, Sviridov_2025}. The fifth axis is safety, which spans refusal of unsafe requests, hallucination rate, and harm avoidance, quantified by dedicated safety and hallucination suites \citep{han2024medsafetybench, pandit2025medhallu, arora2025healthbench}. Reporting all five axes jointly, rather than an accuracy figure, is what turns a benchmark result into evidence of clinical readiness.

\subsection{Training Environments and Clinical Gyms}
\label{sec:gyms}
Evaluation alone does not produce capable agents, and the trajectory of general-purpose agents suggests that the binding constraint is the availability of rich, interactive training environments rather than model scale by itself. In the general domain, the move from single turn supervised data to large-scale interactive reinforcement learning has been enabled by frameworks that run many concurrent, multi-turn, multi task rollouts and stabilize learning across heterogeneous environments \citep{zhang2025agentrl}, which is precisely the recipe that lifts agents from imitating demonstrations to discovering recovery strategies under their own mistakes. This general trend motivates building the clinical analog, a training environment where a medical agent can act, observe consequences, and be rewarded, rather than only being graded on a fixed test set.

The clinical analog is beginning to take shape along three lines. The first line repurposes existing tool and execution benchmarks as gyms, turning their verifiable success signals into dense rewards for policy optimization, as seen in agent training oriented medical suites that expose electronic record queries, calculators, and retrieval as actionable tools \citep{xu2025medagentgym, shi2024ehragent, khandekar2024medcalc}. The second line builds high fidelity simulation environments in which a language model plays the patient, the colleague, or the downstream system, so that an agent can be trained on full diagnostic dialogues and multi-step workups without touching protected data \citep{schmidgall2024agentclinic, Sviridov_2025, yan2026clinicallab}. The third line connects agents to realistic clinical infrastructure such as standardized record interfaces and ordering systems, which both raises the ceiling of what can be practiced and forces the agent to respect the constraints of real workflows \citep{jiang2025medagentbench, lee2025fhir}. Read together, these efforts indicate that the field is shifting from static medical question answering toward closed-loop clinical gyms, and that the agents which prove ready for the real world will be the ones trained inside environments that resemble it.

\begin{figure*}[!t]
\centering
\includegraphics[width=1\textwidth]{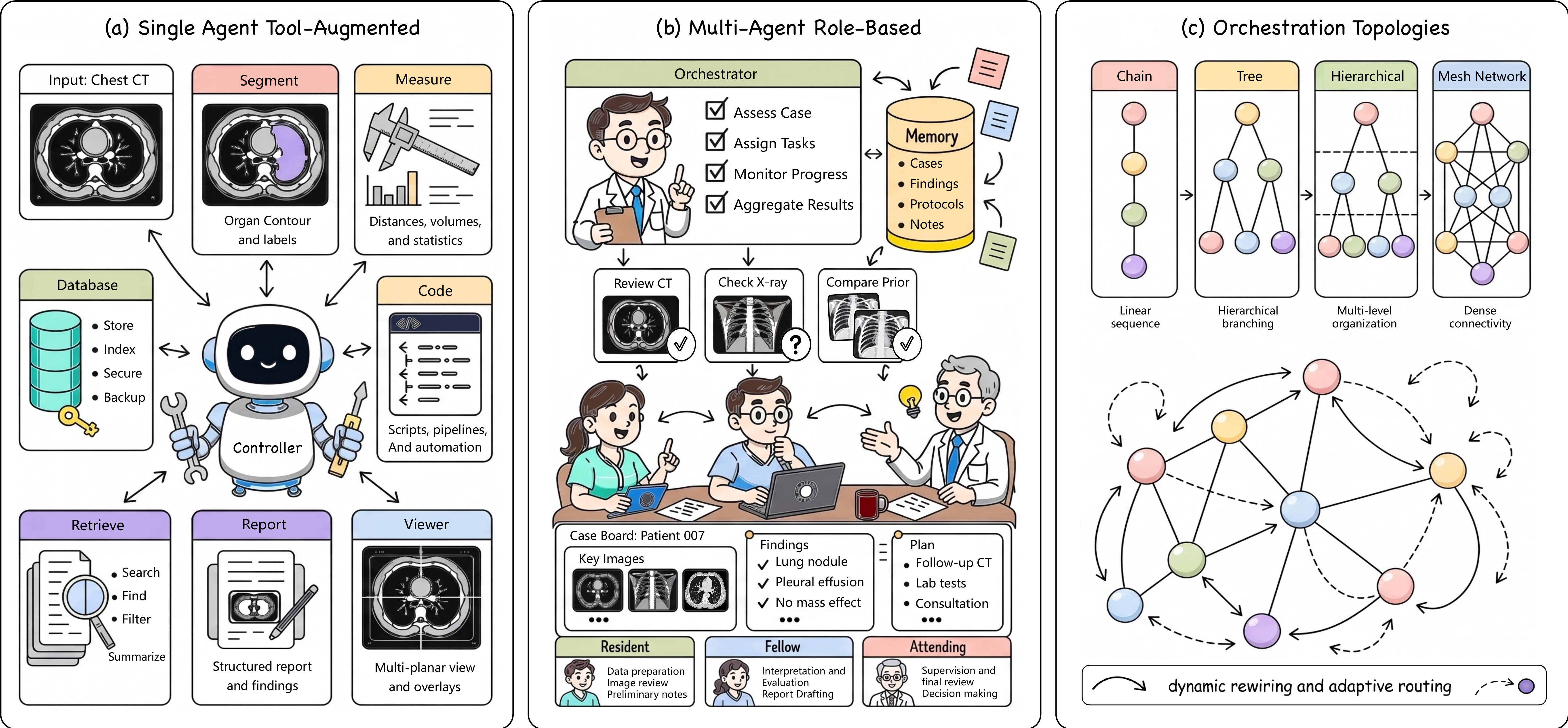}
\caption{\textbf{Three stages of architectural scaling for medical agents, ordered by where coordination is decided.} (a)~Single agent tool augmented: one controller decides everything and calls an external tool repertoire. (b)~Multi agent role-based: coordination is fixed at design time through specialized roles under an orchestrator that mirrors clinical division of labor. (c)~Orchestration topologies: coordination itself becomes a learnable, query adaptive object. Two cross cutting design choices recur across all stages, namely whether actions are tool calls or executable code, and whether the system is static or revises itself from experience.}
\label{fig:arch}
\end{figure*}


\section{Agent Framework Scaling: From Tools to Topologies}\label{sec:arch}

We organize this design space along one axis, \textbf{where coordination is decided}, that is, who chooses the next action and how that choice is distributed across one or more reasoning processes. Three stages emerge in order of increasing structural complexity, a \textit{single agent tool augmented} controller, a \textit{multi-agent role-based} team, and an explicit \textit{orchestration topology} in which the interaction graph itself becomes an object of optimization \citep{zhu2026llm}. Two further choices cut across all three stages and are better treated as orthogonal dimensions than as separate categories. The first is the \textit{action interface}, whether an agent acts through discrete tool calls or by emitting executable code, a distinction traceable to the program aided reasoning of \textsc{PAL} \citep{gao2023pal} and the tool learning of \textsc{Toolformer} \citep{schick2023toolformer}. The second is \textit{adaptability}, whether the system is static at deployment or revises its own behaviour, prompts, memory, or structure over time. \figref{fig:arch} summarizes the three stages together with these two cross cutting choices, which we use to situate medical imaging systems against their general-purpose antecedents.

\subsection{Single Agent Tool Augmented}\label{sec:single}

The simplest and still dominant design wraps a single multimodal backbone with a curated toolbox, so that all coordination is internal to one reasoning loop. The template is the interleaving of reasoning and acting in \textsc{ReAct} \citep{yao2022react}, where the model alternates between a thought, a tool invocation, and an observation, refined by chain of thought \citep{wei2022chain}, least to most decomposition \citep{zhou2022least}, and verbal self-correction in \textsc{Reflexion} \citep{shinn2023reflexion}. In the visual domain the same loop is instantiated either by tool routing, as in \textsc{HuggingGPT} \citep{shen2023hugginggpt} and \textsc{MM-ReAct} \citep{yang2023mm}, or by program synthesis over visual primitives, as in \textsc{ViperGPT} \citep{suris2023vipergpt} and \textsc{VisProg} \citep{gupta2023visual}, which already anticipate the code as action interface below. Medical instantiations specialize this template to clinical perception and measurement. \textsc{CheXagent} \citep{chen2024chexagent} couples a vision-language backbone with chest radiograph reasoning , \textsc{MedRAX} \citep{fallahpour2025medrax} equips a single controller with segmentation, detection, and report grounding tools, \textsc{MMedAgent} \citep{li2024mmedagent} learns to select among heterogeneous medical tools, and \textsc{CXR-Agent} and related controllers add report level retrieval and verification \citep{lou2025cxragent, madavan2025med, fang2025towards}. This design remains competitive because a single agent avoids the communication overhead and error propagation of multi party deliberation, so when the toolbox already covers the clinical subtasks a tight perception, measurement, and reporting loop is both accurate and auditable.

The action interface is the most consequential variation within this stage. Rather than emitting a fixed schema tool call, an agent may write and execute code, which lets it compose unforeseen operations, manipulate intermediate tensors, and interact with real software environments. \textsc{EHRAgent} demonstrates code as action over structured clinical records \citep{shi2024ehragent}, \textsc{MedOpenCLAW} drives a real 3D Slicer environment so that the agent operates the same software a radiologist would \citep{shen2026medopenclaw}, and environment scaling efforts build executable clinical sandboxes in which such agents can be trained and stress tested \citep{song2026envscaler, dong2026agent}. The motivation is expressivity under partial tool coverage, since executable code turns a closed tool menu into an open action space, at the cost of requiring sandboxing and verification. This shift also enables reinforcement style improvement of the agent itself, as in software engineering oriented reward learning \citep{wei2026swe} and the radiology controller \textsc{RadAgent} \citep{roschewitz2026radagent}, which points toward the adaptability dimension we revisit later.

\subsection{Multi Agent Role Based}\label{sec:multi}

When a task spans several specialties or demands explicit cross checking, a single loop becomes a bottleneck, and coordination is lifted into a team of role specialized agents. Two precedents establish the motivation. Role based software teams such as \textsc{MetaGPT} encode standard operating procedures so that specialization reduces error \citep{hong2024metagpt}, while studies of scale show that adding agents helps only when their interaction is well structured rather than merely numerous \citep{guo2024large, qian2025scaling}. Medicine adopts this paradigm because it mirrors a tumour board or a reading room, where a resident, a fellow, and an attending contribute graded expertise and a consensus is negotiated. \textsc{MDAgents} adaptively decides whether a query warrants a solo responder or a full panel \citep{kim2024mdagents}, \textsc{MedAgents} and \textsc{MDTeamGPT} structure multi role diagnostic discussion \citep{tang2024medagents, chen2025mdteamgpt}, and \textsc{M3Builder} and \textsc{MARCH} extend role decomposition to imaging workflow construction and multimodal reasoning \citep{feng2025m, lin2026march}. The justification is calibrated deliberation, since distributing the diagnosis across roles makes disagreement explicit and reviewable, which matters more clinically than raw throughput.

Two practical concerns shape this stage. The first is reliability, since more agents introduce more failure modes, and recent benchmarks caution that multi-agent gains are uneven and sometimes illusory once cost and consistency are accounted for \citep{zhu2026medagentboard, elboardy2025medical, froger2025scaling}. The second is persistent memory, since a clinical team accumulates shared context across a case and across cases, which motivates structured cross agent memory such as \textsc{G-Memory} \citep{yang2026graph} and longitudinal medical agent records \citep{ge2026clinicalagents, kim2025towards}. Memory is also the natural entry point for adaptability at the team level, since revising what is stored and retrieved lets a fixed set of roles improve without changing their wiring, connecting this stage to the self-revising systems below.

\subsection{Orchestration Topologies}\label{sec:topo}

The third stage makes the interaction structure itself a first-class design variable. Instead of fixing a chain, a star, or a debate a priori, the communication graph among agents and tools is searched, learned, and continually rewired at runtime, which subsumes the earlier stages as special cases of a single more general formulation. The reasoning topology lineage from chains \citep{wei2022chain} to trees \citep{yao2023tree} and graphs of thought \citep{besta2024graph} already showed within a single agent that the shape of deliberation, not only its content, governs performance, and orchestration generalizes this insight from one reasoner to entire populations of agents. \textsc{GPTSwarm} represents a multi-agent system as an optimizable graph whose edges are tuned for the task \citep{zhuge2024gptswarm}, \textsc{DyLAN} activates and prunes agents according to their contribution \citep{liu2023dynamic}, and the agentic supernet of \textsc{MaAS} allocates topology and compute as a function of query difficulty so that easy cases stay cheap and hard cases recruit more structure \citep{zhang2025multi}. Beyond selection from a fixed space, recent work generates topologies on the fly by graph diffusion over candidate structures \citep{jiang2025dynamic}, and surveys frame the field as a decisive movement from static templates toward dynamic runtime graphs \citep{yue2026static, kulkarni2026benchmarking}. The motivation is difficulty adaptive allocation, since a clinical deployment faces a long tail of cases whose required reasoning depth varies by orders of magnitude, and a single fixed topology must either overspend on the easy majority or underserve the hard minority.

This stage is also where adaptability becomes structural rather than parametric. A system that rewires its own graph, recruits new roles, or retires ineffective ones is in effect editing its own architecture, the strongest form of the self-evolving behaviour previewed earlier through code-level self-improvement and memory-level revision. Work on evolving agent populations and self-organizing collectives makes this explicit \citep{fan2026evolving, li2024agent}, and we read it as the natural endpoint of the coordination axis, since once the question of where coordination is decided is itself delegated to the system, the boundary between using an agent and designing one dissolves. For medical imaging this is double edged, since dynamic structure promises efficiency and robustness across heterogeneous workloads, yet complicates the auditability and reproducibility that safe deployment demands, a tension we return to under evaluation and governance.
\begin{figure*}[t]
    \centering
    \vspace{-3mm}
    \begin{adjustbox}{max width=0.98\textwidth}
        \begin{forest}
            forked edges,
            for tree={
                grow=east, reversed=true, anchor=base west, parent anchor=east,
                child anchor=west, base=left, font=\normalsize, rectangle,
                draw=hidden-draw, rounded corners, align=left, minimum width=1em,
                edge+={darkgray, line width=1pt}, s sep=3pt, inner xsep=0pt,
                inner ysep=3pt, line width=0.8pt,
                ver/.style={rotate=90, child anchor=north, parent anchor=south, anchor=center},
            },
            [
                Capability Scaling, leaf-head, ver
                [
                    Perception\\(\S\ref{sec:perception}), leaf-datasets, text width=8.0em
                    [Visual Encoders, leaf-datasets, text width=15.5em
                        [BiomedCLIP~\cite{zhang2023biomedclip}{,} RAD-DINO~\cite{perezrad}{,} CONCH~\cite{lu2024visual}{,} RadVLM~\cite{deperrois2025radvlm}, modelnode-datasets, text width=35em]]
                    [Multiscale Processing, leaf-datasets, text width=15.5em
                        [FPN~\cite{lin2017feature}{,} PathChat~\cite{chen2024towards}{,} PathFinder~\cite{ghezloo2025pathfinder}, modelnode-datasets, text width=35em]]
                    [Visual Tokenization, leaf-datasets, text width=15.5em
                        [LLaVA-Med~\cite{li2023llava}{,} BTB3D~\cite{hamamci2026better}{,} InternVL~\cite{chen2024internvl}, modelnode-datasets, text width=35em]]
                ]
                [
                    Reasoning\\(\S\ref{sec:reasoning}), leaf-metrics, text width=8.0em
                    [Chain of Thought, leaf-metrics, text width=15.5em
                        [CoT~\cite{wei2022chain}{,} Med-Gemini~\cite{saab2024capabilities}{,} CARE~\cite{du2026care}{,} MedVLM-R1~\cite{pan2025medvlm}, modelnode-metrics, text width=35em]]
                    [Differential Diagnosis, leaf-metrics, text width=15.5em
                        [MDAgents~\cite{kim2024mdagents}{,} MACD~\cite{li2025macd}{,} MedDxAgent~\cite{rose2025meddxagent}, modelnode-metrics, text width=35em]]
                    [Search \& Verification, leaf-metrics, text width=15.5em
                        [ToT~\cite{yao2023tree}{,} MedAgent-Zero~\cite{tang2024medagents}{,} ClinicalAgents~\cite{ge2026clinicalagents}{,} TTC~\cite{snell2024scaling}, modelnode-metrics, text width=35em]]
                ]
                [
                    Planning\\(\S\ref{sec:planning}), leaf-methods, text width=8.0em
                    [ReAct Planning, leaf-methods, text width=15.5em
                        [ReAct~\cite{yao2022react}{,} MedRAX~\cite{fallahpour2025medrax}{,} MedOpenClaw~\cite{shen2026medopenclaw}, modelnode-methods, text width=35em]]
                    [Plan and Solve, leaf-methods, text width=15.5em
                        [Plan-Solve~\cite{wang2023plan}{,} MEDCO~\cite{wei2024medco}{,} MDTeamGPT~\cite{chen2025mdteamgpt}, modelnode-methods, text width=35em]]
                    [Hierarchical \& Adaptive, leaf-methods, text width=15.5em
                        [MAISTRO~\cite{tzanis2025maistro}{,} MACRO~\cite{fan2026evolving}{,} EnvScaling~\cite{fang2025towards}, modelnode-methods, text width=35em]]
                ]
                [
                    Memory\\(\S\ref{sec:memory}), leaf-datasets, text width=8.0em
                    [Cross-Case Memory, leaf-datasets, text width=15.5em
                        [Synapse~\cite{zheng2024synapse}{,} AWM~\cite{wang2024agent}{,} A-MEM~\cite{xu2026mem}, modelnode-datasets, text width=35em]]
                    [Retrieval Augmentation, leaf-datasets, text width=15.5em
                        [RAG~\cite{lewis2020retrieval}{,} RULE~\cite{xia2024rule}{,} MM-RAG~\cite{karim2025multimodal}, modelnode-datasets, text width=35em]]
                    [Knowledge Graph Grounding, leaf-datasets, text width=15.5em
                        [MedReason~\cite{wu2025medreason}{,} MedKA~\cite{deng2025medka}{,} GraphRAG~\cite{klila2026injecting}, modelnode-datasets, text width=35em]]
                ]
                [
                    Tool Use\\(\S\ref{sec:tooluse}), leaf-metrics, text width=8.0em
                    [Protocols (MCP), leaf-metrics, text width=15.5em
                        [MCP~\cite{hou2025model}{,} MCP-FHIR~\cite{ehtesham2025enhancing}{,} MCP Security~\cite{hasan2025model}, modelnode-metrics, text width=35em]]
                    [Selection \& Scaling, leaf-metrics, text width=15.5em
                        [ToolLLM~\cite{qin2024toolllm}{,} Env Scaling~\cite{fang2025towards}{,} EnvScaler~\cite{song2026envscaler}, modelnode-metrics, text width=35em]]
                    [Composition \& Discovery, leaf-metrics, text width=15.5em
                        [CREATOR~\cite{qian2023creator}{,} MedOrch~\cite{he2025medorch}{,} MedOpenClaw~\cite{shen2026medopenclaw}, modelnode-metrics, text width=35em]]
                ]
                [
                    Action \& \\ Reflection (\S\ref{sec:reflection}), leaf-methods, text width=8.0em
                    [Reflexion, leaf-methods, text width=15.5em
                        [Reflexion~\cite{shinn2023reflexion}{,} ReflecTool~\cite{liao2025reflectool}, modelnode-methods, text width=35em]]
                    [Self Refine, leaf-methods, text width=15.5em
                        [Self-Refine~\cite{madaan2023self}{,} Evo-MedAgent~\cite{shen2026evo}{,} MACRO~\cite{fan2026evolving}, modelnode-methods, text width=35em]]
                    [Verification \& Audit, leaf-methods, text width=15.5em
                        [MDAgents~\cite{kim2024mdagents}{,} GREEN~\cite{ostmeier2024green}{,} MedAgentAudit~\cite{gu2025medagentaudit}, modelnode-methods, text width=35em]]
                ]
                [
                    Loop Engineering\\(\S\ref{sec:harness}), leaf-methods, text width=8.0em
                    [Agent Loop \& ACI, leaf-methods, text width=15.5em
                        [ReAct~\cite{yao2022react}{,} SWE-agent~\cite{yang2024swe}{,} CodeAct~\cite{wang2024executable}{,} NL Harness~\cite{pan2026natural}, modelnode-methods, text width=35em]]
                    [Scheduling \& Isolation, leaf-methods, text width=15.5em
                        [AutoGen~\cite{wu2023autogen}{,} MetaGPT~\cite{hong2024metagpt}{,} OpenHands~\cite{wang2025openhands}{,} AIOS~\cite{mei2024aios}, modelnode-methods, text width=35em]]
                    [Tool Routing \& Interop, leaf-methods, text width=15.5em
                        [MCP~\cite{hou2025model}{,} Interop Survey~\cite{ehtesham2025survey}{,} Harness Survey~\cite{meng2026agent}, modelnode-methods, text width=35em]]
                    [{Context, Memory \& Governance}, leaf-methods, text width=15.5em
                        [Context Eng.~\cite{anthropic2025contextengineering}{,} MemGPT~\cite{packer2023memgpt}{,} GuardAgent~\cite{xiang2024guardagent}{,} VeriGuard~\cite{miculicich2025veriguard}, modelnode-methods, text width=35em]]
                ]
                [
                    Self-Evolving\\Systems (\S\ref{sec:skills}), leaf-datasets, text width=8.0em
                    [Skill Acquisition, leaf-datasets, text width=15.5em
                        [Voyager~\cite{wang2023voyager}{,} ExpeL~\cite{zhao2024expel}{,} AWM~\cite{wang2024agent}{,} CREATOR~\cite{qian2023creator}, modelnode-datasets, text width=35em]]
                    [Skill Retrieval \& Composition, leaf-datasets, text width=15.5em
                        [Synapse~\cite{zheng2024synapse}{,} Self-Discover~\cite{zhou2024self}{,} TroVE~\cite{wang2024trove}{,} Agent Skills~\cite{zhang2025equipping}, modelnode-datasets, text width=35em]]
                    [System-Level Evolution, leaf-datasets, text width=15.5em
                        [Agent Hospital~\cite{li2024agent}{,} Evo-MedAgent~\cite{shen2026evo}{,} Self-Evolving Survey~\cite{fang2025comprehensive}, modelnode-datasets, text width=35em]]
                    [Governance \& Safety, leaf-datasets, text width=15.5em
                        [VeriGuard~\cite{miculicich2025veriguard}{,} MedAgentAudit~\cite{gu2025medagentaudit}{,} CodeAct~\cite{wang2024executable}, modelnode-datasets, text width=35em]]
                ]
            ]
        \end{forest}
    \end{adjustbox}
    \caption{Taxonomy of the capability scaling stack. Six cognitive modules realize the agent's intelligence as stations on the loop; loop engineering binds them into an autonomous execution cycle; and the self-evolving systems layer shows how compounding experience across patients turns a static pipeline into one that keeps improving.}
    \label{fig:capability-taxonomy}
\end{figure*}
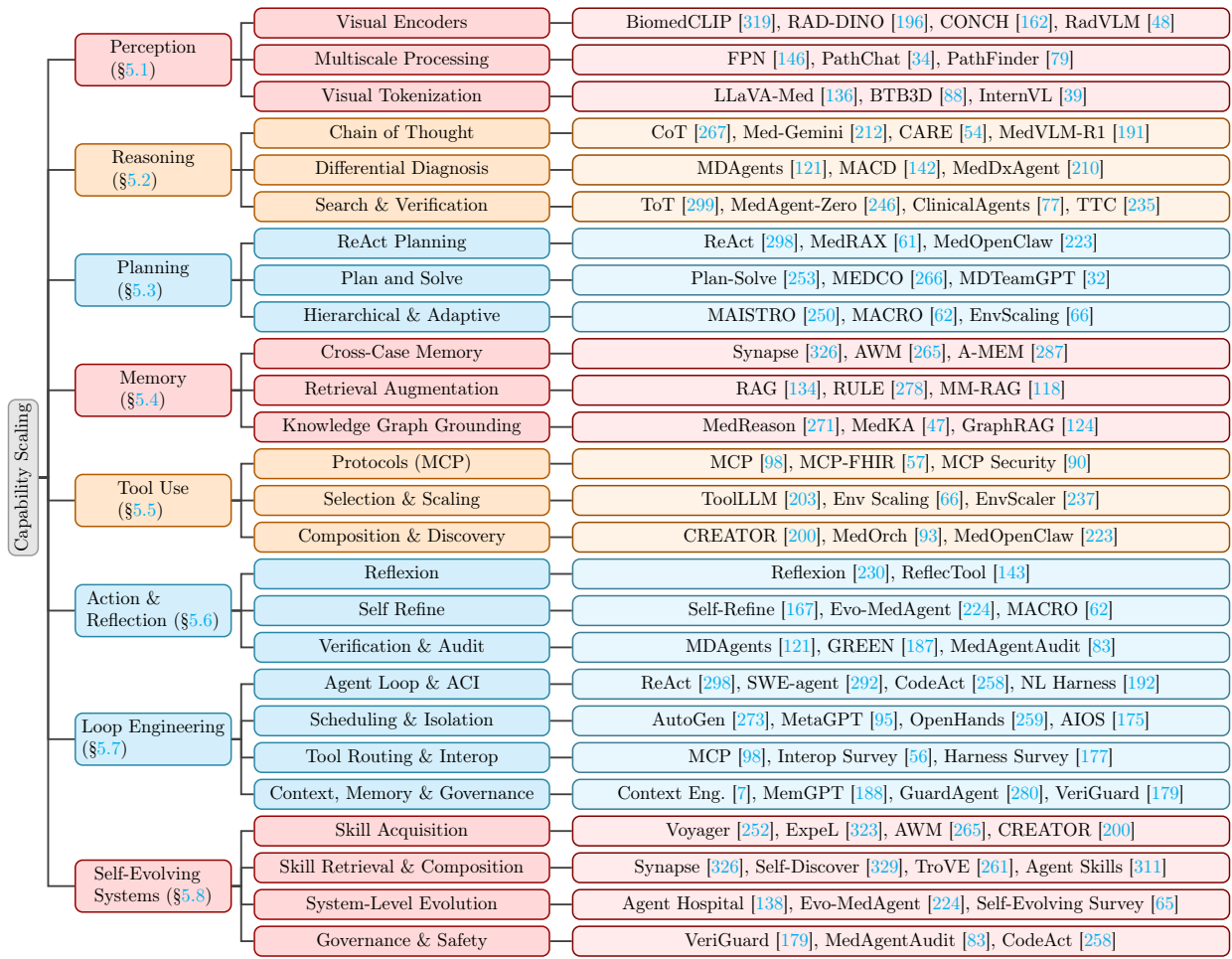


\section{Capability Scaling: The Harness-Driven Cognitive Loop}
\label{sec:core}

\begin{tcolorbox}[
  colback=secblue!5,
  colframe=secblue!50,
  colbacktitle=secblue!50,
  coltitle=black,
  title={\textbf{\textcolor{secblue}{Capability:} Three Layers of the Cognitive Loop}},
  boxrule=5pt,
  arc=5pt,
  drop shadow,
  parbox=false,
  before skip=5pt,
  after skip=10pt,
  left=5pt,
  right=20pt,
]
\begin{itemize}[leftmargin=*,noitemsep]
\renewcommand\labelitemi{$\diamond$}
    \item \textbf{Cognitive modules} (Sections~\ref{sec:perception} to~\ref{sec:reflection}): realize the agent's intelligence as six stations on the loop, perception, reasoning, planning, memory, tool use, and reflection, each revisited every turn so that a finding detected once stays a stable anchor for later decisions.
    \item \textbf{Loop engineering} (Section~\ref{sec:harness}): engineers the runtime that binds these modules into an autonomous perceive, reason, plan, act, and reflect cycle through scaffolding, context management, and runtime governance, transforming passive components into a system that runs itself.
    \item \textbf{Self-evolving systems} (Section~\ref{sec:skills}): emerge as the loop compounds experience across patients, consolidating recurring behavior into durable procedures that turn a static diagnostic pipeline into a system that keeps improving without weight updates.
\end{itemize}
\end{tcolorbox}

A medical agent is defined less by any single faculty than by how its faculties cycle together around the patient. A radiologist does not detect a lesion, reason about it, order a follow up, and reflect on the report as four disconnected acts; these stages form one continuous loop, and a clinically useful agent must reproduce that loop rather than a bag of isolated skills. We therefore organize this section in three layers. We first introduce the six cognitive modules that realize the agent's intelligence, perception, reasoning, planning, memory, tool use and reflection, each revisited every turn so that a finding detected once stays a stable anchor for later decisions. We then formalize the runtime substrate, loop engineering, that binds these modules into an autonomous execution cycle, because recent agentic systems repeatedly show that the engineering of the loop, rather than the scale of the underlying model, governs whether a clinical task actually closes \citep{yang2024swe, jimenez2024swe}. Finally we show how loops that compound experience across patients give rise to self-evolving systems, turning a static diagnostic pipeline into one that keeps improving as it sees more patients. \figref{fig:capability-taxonomy} organizes the literature along exactly these three layers.

\subsection{Perception}
\label{sec:perception}

Perception is the entry point, where raw imaging signals are encoded into semantic representations that reasoning can consume, and in medicine it must bridge profoundly heterogeneous modalities, since one case may couple a chest radiograph, a whole slide pathology image and free text findings within a single decision. Encoders tuned to the medical domain rather than borrowed wholesale from natural images are the practical foundation here, whether through contrastive image text pretraining on biomedical corpora \citep{zhang2023biomedclip}, self-supervised radiographic backbones that learn anatomy without labels \citep{perezrad}, pathology specific representations distilled from histopathology at scale \citep{lu2024visual}, or radiology vision-language models that align images with structured reports \citep{deperrois2025radvlm}. Multimodal clinical benchmarks expose how brittle agents become when perception fails to ground these signals in a shared representation \citep{schmidgall2024agentclinic}, which is why the design question is not only how accurately a finding is detected once but how stably its representation persists as the agent gathers more views, so that later reasoning and reflection operate on a consistent semantic anchor for the same lesion rather than on a representation that drifts across iterations.

\subsection{Reasoning}
\label{sec:reasoning}

Reasoning converts grounded perceptions into clinical decision logic, and its quality hinges on the structure imposed on the inference rather than on isolated chains of thought. Self composed reasoning structures let an agent assemble the inference scaffold best suited to the case at hand \citep{zhou2024self}, and in medicine this structure is often distributed across collaborating roles, where an adaptive ensemble of specialist agents deliberates before committing to a differential diagnosis \citep{kim2024mdagents}. Crucially, a clinical inference is never terminal, since each diagnostic hypothesis is a provisional commitment that action and reflection may later overturn once a new sequence or a discordant lab returns, which is why we treat reasoning as a recurring step rather than a single forward pass.

\subsection{Planning}
\label{sec:planning}

Planning decomposes a clinical goal into an ordered sequence of admissible actions, such as which prior study to retrieve, which measurement to compute, and which guideline to consult, and its effectiveness is inseparable from the scaffolding that drives execution. Role structured frameworks externalize the plan as an explicit artifact that collaborating agents can inspect and refine \citep{hong2024metagpt}, while reusable workflow memory lets an agent abstract recurring imaging procedures into plan templates that shortcut future episodes \citep{wang2024agent}. Because planning sits upstream of tool use, a plan is only as good as the ability to detect when an executed step diverges from its intent, so planning and reflection remain coupled through the same control flow rather than acting as independent stages.

\subsection{Memory}
\label{sec:memory}

Memory is the write back channel of the agent, the place where both the transient context of the current encounter and the durable experience of past cases are stored and retrieved. Short term context must be actively managed so that a long imaging workup does not overflow the model window, which motivates operating system style paging of context in and out of the active buffer \citep{packer2023memgpt}. Long term memory accumulates across episodes, whether as trajectory exemplars that prime future control on similar studies \citep{zheng2024synapse} or as distilled workflow abstractions that compress past solutions into reusable form \citep{wang2024agent}. The defining property here is directionality, since reflection deposits what perception and reasoning will later draw upon, closing the cycle across time and across patients rather than only within a single turn.

\subsection{Tool Use}
\label{sec:tooluse}

Tool use is the actuator, the point at which the agent reaches beyond its parameters to invoke imaging utilities, retrieval services and clinical knowledge bases, and its reliability depends first on the interface, since standardized protocols such as the Model Context Protocol decouple the agent from the idiosyncrasies of each external service \citep{hou2025model}. Expressing actions as executable code rather than rigid function calls further broadens the action space and composes naturally with the control flow that a multi-step workup requires \citep{wang2024executable}. Beyond consuming a fixed toolset, capable agents create and curate their own, inducing verifiable toolboxes from experience \citep{wang2024trove} and synthesizing new tools on demand to disentangle abstract from concrete reasoning \citep{qian2023creator}, so that the actuator itself grows alongside the clinical competence of the agent.

\subsection{Action and Reflection}
\label{sec:reflection}

Action and reflection evaluates each executed step and feeds the verdict back into perception, memory and scaffolding, and it is precisely this write back that distinguishes a self-evolving medical agent from a static diagnostic pipeline. Experiential learners distill successes and failures into transferable insights that bias subsequent episodes \citep{zhao2024expel}, and the broader self-evolving paradigm frames this outer loop as the bridge between frozen foundation models and lifelong agentic systems \citep{fang2025comprehensive}. In the imaging clinic the consequence is tangible, since agents that revise their reasoning structures \citep{zhou2024self} and accumulate verified procedures over cases turn the substrate, not merely the weights, into the locus of capability growth, which is the property a deployed system needs to keep improving as it sees more patients.

\subsection{Loop Engineering: From Passive Modules to Autonomous Execution}
\label{sec:harness}

The six modules above each contribute a necessary faculty, but none of them runs itself. A perception encoder does not decide when to look again; a planner does not know when to abandon a failing workup. What transforms these passive components into an autonomous clinical agent is the runtime that drives them as a single closed loop, and we use \textit{loop engineering} to denote the systematic design of this runtime. It is best understood as a specialization of \textit{harness engineering}, the broader and very recent practice captured by the working identity that an agent equals a model plus the harness that wraps it, where production reliability is increasingly bound by the harness rather than by the underlying model \citep{anthropic2025harnesses, meng2026agent}. A recent formalization treats the harness as a six component object spanning the execution loop, tool registry, context manager, state store, lifecycle hooks and evaluation interface \citep{meng2026agent}, while a complementary account decomposes it into control, agency and runtime \citep{he2026harness}. Loop engineering isolates the execution loop and asks how each turn ingests new observations, commits an admissible action, and folds the outcome back into the next turn; in a clinical setting this is precisely the difference between a one shot model that emits a single answer and an agent that orders an extra view, re reads the prior study, and revises its impression. The stakes are concrete in agentic coding, where a carefully shaped agent computer interface determines task success more than raw model capacity \citep{yang2024swe, jimenez2024swe}, where the action representation is itself a design variable since executable code actions outperform free form text actions \citep{wang2024executable, qiao2023taskweaver}, and where the harness policy can even be externalized as an editable natural language artifact rather than buried in controller code \citep{pan2026natural}, all observations that transfer directly to how a medical agent should be built.

What makes loop engineering hard reduces to three design surfaces, and a clinical setting pushes each to an extreme that general agents rarely face. The first is \textit{scaffolding}, which fixes the control flow, the stop conditions and the action space; in medicine this is where a tumour board style division of labour \citep{hong2024metagpt} or a multi-agent conversation runtime \citep{wu2023autogen} encodes who decides when a workup may stop. The second is \textit{context engineering}, which a single imaging encounter strains the moment it spans many windows of priors, labs and guidelines, motivating virtual memory style paging \citep{packer2023memgpt}, evolving context playbooks \citep{zhang2025agentic} and the broader practice surveyed by \citet{mei2025survey}. The third is \textit{runtime governance}, which matters far more here than in coding because an unverified tool call carries real patient cost, and it is instantiated by guardrail layers \citep{xiang2024guardagent} and verification driven gating \citep{miculicich2025veriguard}. Orchestration runtimes and agent operating systems bind these surfaces into one schedulable substrate \citep{mei2024aios, wang2025openhands}, and the Model Context Protocol standardizes how the loop reaches imaging tools, retrieval services and records \citep{hou2025model}, which is the concrete sense in which capability scales with the harness rather than with any single module \citep{xi2025rise}.

\subsection{From Loops to Self-Evolving Systems}
\label{sec:skills}

The loop described above enables autonomous execution within a single episode, but its compounding clinical value emerges only when reflection consolidates recurring behavior into durable, retrievable procedures that outlast any one case. A skill, in this sense, is a verified procedure lifted out of one successful episode and packaged so that it can be selected and composed on later cases rather than rediscovered from scratch every time \citep{zhang2025equipping}. Open ended embodied agents demonstrate the principle by growing an expanding library of reusable skills from interaction \citep{wang2023voyager}, and toolbox induction shows the same dynamic for programmatic clinical tasks, where an agent distills its own routines into a curated, callable inventory \citep{wang2024trove}. This layer is what carries capability from the laboratory to the ward, since evolvable medical agents that practice across simulated cases accumulate competence without weight updates \citep{li2024agent}, and realistic clinical environments increasingly benchmark exactly this kind of accumulated, procedure level competence rather than static one shot accuracy, reflecting the long-term nature of clinical expertise acquisition and continual workflow adaptation \citep{jiang2025medagentbench}.

Over many patients these procedures stop being isolated shortcuts and begin to form a shared institutional repertoire, where a measurement protocol refined on one tumour board or a retrieval routine validated on one cohort becomes available to every subsequent case the system encounters. The system no longer merely executes, it evolves: updating its own context playbooks, expanding its tool inventory, refining its scaffolding policies, and compressing successful strategies into reusable abstractions that bias the loop toward better outcomes on unseen cases. It is precisely this slow consolidation of autonomous practice into compounding competence, not confined to a skill library but spanning the entire runtime substrate, that turns a static diagnostic pipeline into a self-evolving clinical system, one that keeps improving as it sees more patients, adapts to new scenarios, and becomes increasingly reliable over time.

\begin{figure}[t]
\centering
\includegraphics[width=1\columnwidth]{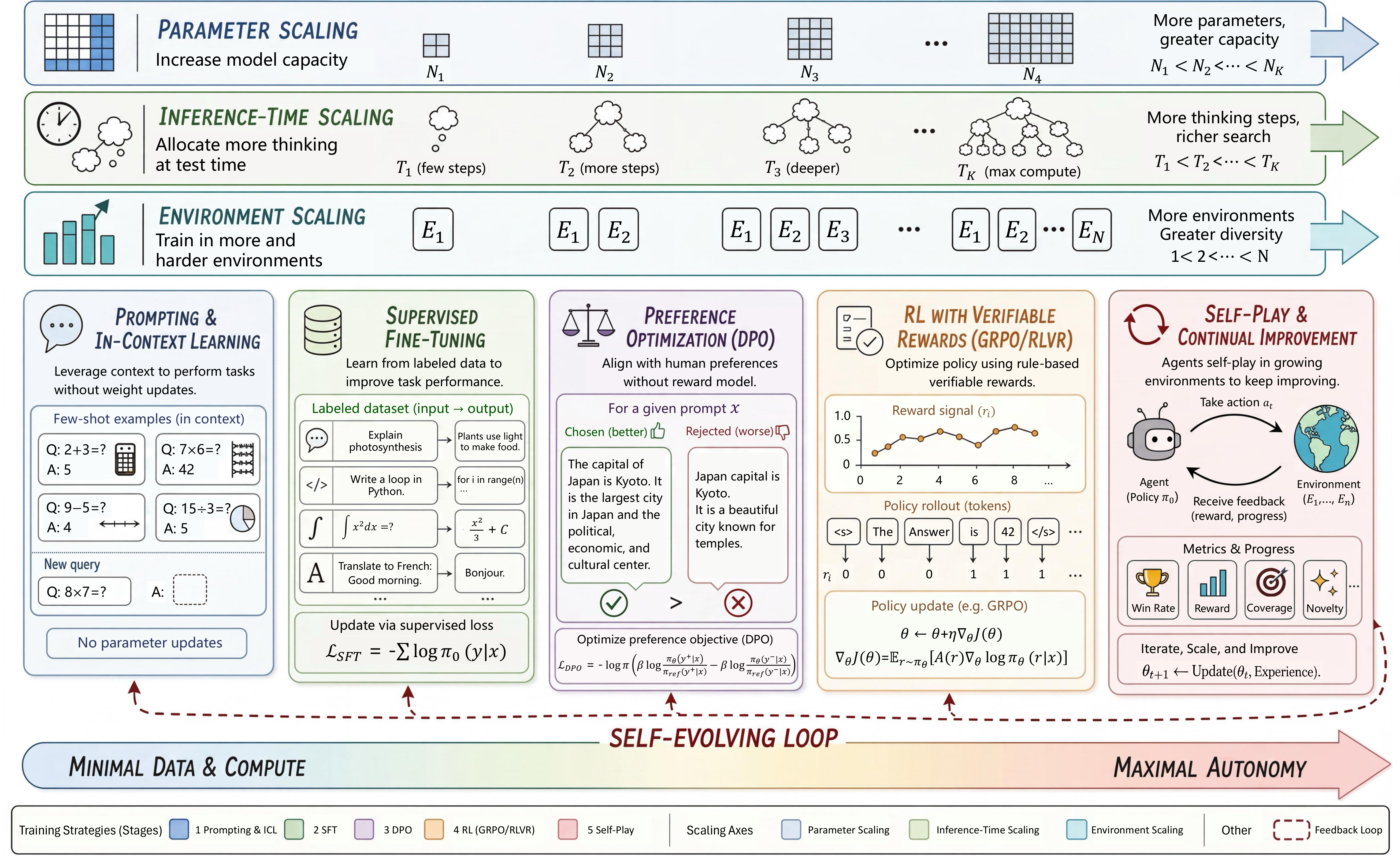}
\caption{\textbf{Training time strategy spectrum and scaling axes for medical agents.} The horizontal axis traces a supervision reducing and autonomy increasing ladder of five stages, from prompting and in-context learning, through supervised fine-tuning and preference optimization, to reinforcement learning with verifiable rewards and self-play with continual improvement. Three orthogonal axes, parameter, inference time, and environment scaling, amplify capability along complementary dimensions, and the dashed path denotes the self-evolving loop in which an agent reuses its own outcomes as supervision.}
\label{fig:training}
\end{figure}

\section{Training Time Scaling: From Imitation to Autonomous Learning}\label{sec:train}

The training strategy (\figref{fig:training}) determines how a medical agent acquires and refines its capabilities~\citep{guo2025deepseek, rafailov2023direct, shao2024deepseekmath}. Whereas the architecture (\secref{sec:arch}) and capability (\secref{sec:core}) sections describe \emph{inference time} scaling, that is, how to extract more competence from a fixed model at deployment through better orchestration and more cognitive modules, this section concerns \emph{training time} scaling, that is, how the agent's own parameters, policies, memory, and even structure change so that it grows more capable over time. We identify a spectrum of approaches ranging from zero-parameter prompting to closed-loop self-improvement, and frame the section's arc as a progression toward \emph{self-evolving systems} in which the boundary between training and deployment dissolves. We additionally examine the \emph{scaling} paradigm that encompasses model parameter scaling, inference time compute scaling, and environment scaling as complementary axes, with environment scaling singled out as the most actionable lever for medical agents that already inhabit tool rich clinical ecosystems.

\begin{tcolorbox}[
  colback=secblue!5,
  colframe=secblue!50,
  colbacktitle=secblue!50,
  coltitle=black,
  title={\textbf{\textcolor{secblue}{Adaptation:} A Supervision Signal Ladder}},
  boxrule=5pt,
  arc=5pt,
  drop shadow,
  parbox=false,
  before skip=5pt,
  after skip=10pt,
  left=5pt,
  right=20pt,
]
\begin{itemize}[leftmargin=*,noitemsep]
\renewcommand\labelitemi{$\diamond$}
    \item \textbf{Imitation} (Sections~\ref{sec:prompt},~\ref{sec:sft}): elicits behavior from a frozen backbone through prompting and then reshapes the weights by supervised fine-tuning on expert demonstrations, the fastest route to a domain competent agent but bounded by what the labeled corpus already encodes.
    \item \textbf{Reward} (Sections~\ref{sec:dpo},~\ref{sec:rlvr}): shifts from copying outputs to optimizing against feedback, first relative preferences through DPO and RLHF and then programmatically verifiable rewards through RLVR and GRPO, trading annotation cost for a checker that manufactures signal at scale.
    \item \textbf{Self-evolution} (Sections~\ref{sec:selfplay},~\ref{sec:selfevolve}): removes the external annotator entirely, letting the agent generate its own experience through self-play in simulated clinics and ultimately evolve not only weights but memory, tools, and control code, dissolving the boundary between training and deployment.
\end{itemize}
\end{tcolorbox}

\subsection{Prompting and In Context Learning}\label{sec:prompt}

The most accessible strategy, requiring no model modification, relies on carefully crafted prompts and in-context examples to elicit agent behavior from pretrained LLMs~\citep{wei2022chain, tang2024medagents, tang2024medagents}. \textsc{MedAgent-Zero}~\citep{tang2024medagents} demonstrates that well engineered prompts combined with tool access can achieve competitive performance on multiple medical imaging benchmarks without any fine-tuning. Agnostic neuroradiology pipelines~\citep{erdur2026agentic} further validate training-free agentic approaches, achieving strong brain MRI interpretation performance through pure prompting of frontier models such as GPT-5.1 and Gemini 3 Pro.

\subsection{Supervised Fine Tuning (SFT)}\label{sec:sft}

Supervised fine-tuning adapts the pretrained LLM to the medical imaging domain by training on curated datasets of (input, desired output) pairs~\citep{li2023llava, chen2024chexagent, deperrois2025radvlm}. \textsc{LLaVA-Med}~\citep{li2023llava} demonstrates the effectiveness of a two-stage fine-tuning approach. \textsc{CheXagent}~\citep{chen2024chexagent} extends this with a comprehensive training pipeline that includes clinical knowledge alignment, finding extraction, and report generation stages. \textsc{RadVLM}~\citep{deperrois2025radvlm} introduces multitask conversational fine-tuning for radiology vision-language models, achieving improved performance across diverse radiology tasks through joint training on multiple clinical objectives.

\subsection{Preference Optimization}\label{sec:dpo}

Direct Preference Optimization (DPO)~\citep{rafailov2023direct} and its variants offer a reinforcement learning free approach to aligning agent outputs with clinical preferences~\citep{yang2025aligning, yang2025mitigating}. \textsc{OPA-DPO}~\citep{yang2025mitigating} introduces optimized preference alignment specifically for medical vision-language models, demonstrating that domain specific preference optimization significantly outperforms general-purpose DPO. \textsc{Med-RLHF} approaches~\citep{yang2025aligning} demonstrate that preference optimization significantly improves the clinical quality of generated radiology reports compared to SFT alone, particularly in reducing hallucinated findings and improving report structure~\citep{mahdavi2026med}.

\subsection{Reinforcement Learning with Verifiable Rewards (RLVR)}\label{sec:rlvr}

For tasks where correctness can be programmatically verified, reinforcement learning with verifiable rewards (RLVR) provides a scalable training signal without requiring human annotation~\citep{wen2025reinforcement, cai2025reinforcement, guo2025deepseek}. The theoretical foundations of RLVR establish reward design principles and sample complexity bounds informing practical implementations.

\textbf{Group relative optimization and reasoning verification in medical VLMs.} Group Relative Policy Optimization (GRPO)~\citep{shao2024deepseekmath, guo2025deepseek} extends this approach by sampling multiple candidate outputs, evaluating them against verifiable criteria, and optimizing the policy based on relative quality within each group. \textsc{MedGRPO}~\citep{su2025medgrpo, lai2026med} applies GRPO specifically to medical vision-language models, demonstrating improved diagnostic reasoning through group relative optimization, while \textsc{MedVLM-R1}~\citep{pan2025medvlm} further validates reinforcement learning for incentivizing reasoning capability in medical VLMs. \textsc{Patho-R1}~\citep{zhang2026pathor1} applies a three stage pipeline (pretraining, SFT, GRPO+DCPO) specifically to pathology, achieving state-of-the-art performance on VQA, zero-shot classification, and cross-modal retrieval, and \textsc{Med-R1}~\citep{lai2026med} demonstrates that GRPO based training on Qwen2-VL-2B across eight imaging modalities yields a 29.94\% accuracy improvement over the base model, surpassing models 36 times larger. Building on these group relative recipes, a line of work rewards the reasoning process rather than only the final answer: \textsc{ChestX-Reasoner}~\citep{fan2025chestx} advances radiology reasoning through step by step verification with process reward models, introducing \textsc{RadRBench-CXR} for evaluating reasoning chains, and Jing et al.~\citep{jing2025reason} propose ``Reason like a Radiologist'', combining chain of thought prompting with reinforcement learning for verifiable report generation. \textsc{VLM-R1}~\citep{shen2025vlm} provides a generalizable R1 style training framework for vision-language models, discovering reward hacking phenomena in object detection tasks, while \textsc{Clinical-R1}~\citep{gu2025clinical} introduces constrained reinforcement learning with preference optimization for clinical decision making, combining RLVR with clinical safety constraints.

\textbf{Task specialized and causal reward shaping.} Beyond classification and report generation, RLVR has been specialized to distinct medical tasks and to richer reward semantics. \textsc{MedSAM-Agent}~\citep{liu2026medsam} applies RLVR to interactive medical image segmentation, framing segmentation as sequential decision making, and \textsc{MedResearcher-R1}~\citep{yu2025medresearcher, yu2025medresearcher} demonstrates RLVR training for a medical literature research agent with iterative hypothesis refinement. More recently, \textsc{MedEyes}\citep{zhu2026medeyes} and \textsc{MedCausalX}~\citep{lin2026medcausalx} further extend RLVR to progressive diagnostic reasoning and adaptive causal reasoning, respectively, where rewards are derived from verifiable diagnostic outcomes, causal consistency, and counterfactual validity to promote expert-like visual attention and self-correcting reasoning. \textsc{MedSynapse-V}~\citep{zhu2026medsynapse} further advances this direction through Causal Counterfactual Refinement (CCR), which leverages reinforcement learning with counterfactual rewards derived from region-level feature masking to quantify the causal contribution of each latent diagnostic memory, pruning redundancies and aligning latent representations with diagnostic logic through a privileged-autonomous dual-branch internalization paradigm. For a systematic review of RL applications in medical image analysis covering algorithms, engineering challenges, and clinical deployment, see Sampa et al.~\citep{sampa2026reinforcement}.

\textbf{Rubric based and entity level reward design.} A crucial design choice in RLVR for medical agents is the reward function structure. \textsc{HealthBench}~\citep{arora2025healthbench} establishes physician written rubric criteria as the gold standard for evaluation, comprising 48,562 structured criteria across 5,000 clinical conversations. \textsc{Health-SCORE}~\citep{yang2026health} demonstrates that scalable rubric generation can serve dual purposes: both as an evaluation framework and as structured reward signals for RL training, bridging the gap between assessment and optimization. Chen et al.~\citep{chen2026automated} further show that automated rubric construction enables reliable evaluation of medical dialogue systems at scale, while Mallinar et al.~\citep{mallinar2026scalable} provide a scalable evaluation framework validated across multiple health language model families, and the disagreement between LLM judges and clinicians documented by recent work~\citep{delucia2026same} highlights the importance of grounding reward signals in verifiable clinical criteria rather than relying solely on model based judgment. Beyond holistic rubrics, entity level matching has emerged as a promising fine-grained reward signal for medical text generation. \textsc{MRG-R1}~\citep{wang2025mrgr1} introduces reinforcement learning for clinically aligned medical report generation, using entity level clinical finding extraction as the reward to ensure generated reports contain factually correct medical entities, and \textsc{ESC-RL}~\citep{zhou2026escrl} extends this with evidence aware self-critical reinforcement learning for radiology report generation, where the reward function explicitly verifies that mentioned anatomical structures, pathologies, and measurements match ground truth evidence. Jhaveri et al.~\citep{jhaveri2025claimreward} propose claim based rewards for optimizing long form clinical text generation, decomposing outputs into atomic claims and rewarding factual accuracy at the claim level. These entity and claim level reward approaches offer superior granularity compared to holistic sequence level rewards, enabling targeted optimization of factual consistency in medical agent outputs.

\textbf{From single-turn verification to agentic RL.} Beyond the GRPO formulation above, a rapidly growing family of policy optimization algorithms refines its stability and extends it from single-turn verification to multi-turn agentic settings (Table~\ref{tab:rl_algorithms_summary}). Within the single-turn regime, DAPO~\citep{yu2026dapo} decouples the clipping bounds and adds dynamic sampling, Dr.~GRPO~\citep{liu2025understanding} removes the length and standard-deviation normalization biases of vanilla GRPO, GSPO~\citep{zheng2025group} replaces token-level importance ratios with sequence-level ones to stabilize Mixture-of-Experts training, and CISPO~\citep{chen2025minimax} clips the importance-sampling weights rather than the token updates so that rare but decisive tokens still contribute gradients. The multi-turn extension, often termed \emph{agentic RL}, recasts training as sequential decision making over tool calls and observations under partial observability~\citep{zhang2025landscape}, directly matching the formulation of Section~\ref{sec:formal} and shifting the central difficulty to turn-level credit assignment: MT-GRPO~\citep{zeng2025reinforcing} and Turn-PPO~\citep{li2026turn} estimate advantages at the granularity of individual decision turns, ARPO~\citep{dong2025agentic} uses entropy to focus optimization on high-uncertainty tool-call steps, and AT2PO~\citep{zong20262} couples turn-wise credit assignment with tree search. For medical agents, whose consultations and examinations yield rewards only at the end of a long multi-turn trajectory, this agentic formulation is the natural training counterpart to the partial-observability view adopted in Section~\ref{sec:formal}. A closely related and rapidly growing direction is on-policy distillation (OPD), which is formally equivalent to KL-constrained reinforcement learning over student-sampled trajectories~\citep{song2026survey} and underlies the turn-level truncated on-policy distillation recipe that \textsc{Healthcare AI GYM}~\citep{jeong2026healthcare} uses to stabilize multi-turn medical agentic training under realistic clinical interaction settings.

\begin{table}[t]
\centering
\small
\caption{A summary of key reinforcement learning algorithms most relevant to building medical agents beyond vanilla GRPO/PPO, categorized by design focus.}
\label{tab:rl_algorithms_summary}
\begin{tabular}{p{0.26\textwidth}|p{0.50\textwidth}|p{0.12\textwidth}}
\toprule
\textbf{Algorithm} & \textbf{Key Innovation} & \textbf{Resource} \\
\midrule
\multicolumn{3}{c}{\textit{Critic-Free Policy Optimization (GRPO family)}}\\
\midrule
PPO~\citep{schulman2017proximal} & Clipped surrogate objective with a learned critic & \githublink{https://github.com/openai/baselines}\\
GRPO~\citep{shao2024deepseekmath} & Critic-free group-relative advantage & \githublink{https://github.com/deepseek-ai/DeepSeek-Math}\\
DAPO~\citep{yu2026dapo} & Decoupled clip bounds and dynamic sampling & \githublink{https://github.com/BytedTsinghua-SIA/DAPO}\\
Dr.~GRPO~\citep{liu2025understanding} & Removes length and std normalization bias & \githublink{https://github.com/sail-sg/understand-r1-zero}\\
GSPO~\citep{zheng2025group} & Sequence-level importance ratio; stabilizes MoE & \paperlink{https://arxiv.org/abs/2507.18071}\\
CISPO~\citep{chen2025minimax} & Clips IS weights, keeps rare-token gradients & \githublink{https://github.com/MiniMax-AI/MiniMax-M1}\\
REINFORCE++~\citep{hu2025reinforce++} & Global-baseline, critic-free, robust variant & \githublink{https://github.com/OpenRLHF/OpenRLHF}\\
\midrule
\multicolumn{3}{c}{\textit{Agentic / Multi-Turn RL (2025--2026)}}\\
\midrule
MT-GRPO~\citep{zeng2025reinforcing} & Turn-level credit assignment for tool use & \githublink{https://github.com/SiliangZeng/Multi-Turn-RL-Agent}\\
ARPO~\citep{dong2025agentic} & Entropy-guided step-level tool-call optimization & \githublink{https://github.com/RUC-NLPIR/ARPO}\\
GTPO~\citep{ding2025empowering} & Turn-level reward with return-based advantage & \paperlink{https://arxiv.org/abs/2511.14846}\\
AEPO~\citep{dong2025agentic} & Entropy-balanced rollout and policy update & \githublink{https://github.com/RUC-NLPIR/ARPO}\\
Turn-PPO~\citep{li2026turn} & Turn-level critic for long-horizon agents & \paperlink{https://arxiv.org/abs/2512.17008}\\
AT2PO~\citep{zong20262} & Turn-based tree search with turn-wise rewards & \githublink{https://github.com/zzfoutofspace/ATPO}\\
\bottomrule
\end{tabular}
\end{table}

\subsection{Self-play and Continual Improvement}\label{sec:selfplay}

The most advanced training paradigm enables agents to improve through autonomous practice within simulated clinical environments~\citep{li2024agent, du2025llms, shen2026evo, tu2024towards, feng2026doctoragent}. \textsc{Agent Hospital}~\citep{li2024agent} implements a simulation environment where agents practice on synthetically generated patient cases, accumulating diagnostic experience through thousands of simulated consultations. \textsc{EvoPatient}~\citep{du2025llms} introduces co-evolving synthetic patients that grow in complexity alongside the improving agent through multi-turn diagnostic dialogues, simultaneously gathering experience to improve the quality of both questions and answers. \textsc{AMIE}~\citep{tu2024towards} pioneered self-play simulated learning for diagnostic conversations, demonstrating that LLM-based agents trained through autonomous physician-patient role-play can outperform primary care physicians in diagnostic accuracy.

\textbf{Simulated environments and policy learning.} \textsc{DoctorAgent-RL}~\citep{feng2026doctoragent} trains an explicit policy model through multi-agent collaborative reinforcement learning, where a doctor agent optimizes its questioning strategy via RL-based training against a high-fidelity LLM patient agent and a consultation evaluator providing multi-dimensional rewards. \textsc{ATPO}~\citep{cao2026atpo} advances policy learning through adaptive tree-structured policy optimization, formulating multi-turn medical dialogue as a Hierarchical Markov Decision Process and adaptively allocating rollout budgets to high-uncertainty states, enabling a Qwen3-8B model to surpass GPT-4o on diagnostic benchmarks. \textsc{Healthcare AI GYM}~\citep{jeong2026healthcare} provides a Gymnasium-compatible training environment spanning 10 clinical domains with 3.6K+ tasks and 135 domain-specific tools, and proposes Turn-level Truncated On-Policy Distillation (TT-OPD) to stabilize multi-turn agentic RL training. \textsc{MedAgentSim}~\citep{almansoori2025self} contributes an open-source multi-agent simulation with doctor, patient, and measurement agents, enabling self-evolving diagnostic capabilities through experience replay and chain-of-thought reasoning. \textsc{MedAgentGym}~\citep{xu2025medagentgym} provides a scalable training gymnasium for code-based medical reasoning, demonstrating substantial improvement from online reinforcement learning on 72K biomedical tasks.

Bridging the simulation fidelity gap and continual improvement. A critical challenge for self-play training is ensuring that skills learned in simulation transfer to real clinical settings. \textsc{SynthAgent}~\citep{aghaee2026synthagent} addresses this through personality-enriched synthetic patient agents that incorporate comorbidity profiles and psychosocial characteristics, reducing the distribution gap between simulated and real patient interactions. \textsc{MeWM}~\citep{yang2025medical} introduces the first grounded medical world model that combines a VLM policy with learned tumor dynamics and inverse dynamics models, enabling treatment planning agents to train against disease progression trajectories. \textsc{Evo-MedAgent}~\citep{shen2026evo} proposes a self-evolving memory module comprising retrospective clinical episodes, adaptive procedural heuristics, and a tool reliability controller, enabling agents to accumulate inter-case experience at test time without additional training. \textsc{MedChain-Agent}~\citep{liu2026medchain} integrates a feedback mechanism with case-based retrieval-augmented generation for sequential clinical decision adaptation, allowing agents to learn from previous cases across five stages of clinical workflow. \textsc{MMedAgent-RL}~\citep{xia2025mmedagent} combines curriculum learning with multi-agent reinforcement learning, training triage and attending physician agents via staged RL on Qwen2.5-VL to achieve 23.6\% improvement on medical reasoning benchmarks. \textsc{SWE-RL}~\citep{wei2026swe} demonstrates that reinforcement learning on open software engineering evolution data can improve general reasoning capabilities, with implications for code-generating medical agents.

\subsection{Self-Evolving Agents: Mechanisms and Medical Instantiations}\label{sec:selfevolve}

The preceding subsections demonstrate that medical agents can acquire increasingly complex behaviours through simulated practice and reinforcement learning within clinical gyms. Self play and continual improvement, however, are special cases of a broader and faster moving frontier: \emph{self-evolving agents} that modify not only their parameters but their memory, prompts, tools, and even their own control code in response to accumulated clinical experience~\citep{gao2025survey, yin2024g, hu2025automated, zhang2026darwin, zweiger2026self, fang2025comprehensive}. What distinguishes self-evolution from ordinary finetuning or prompting is that the agent itself generates the learning signal, applies the update, and evaluates the result, closing a loop that runs without external human annotation. This property is decisive for medicine because expert labels are scarce, expensive, and subject to interobserver variability, whereas operational signals such as measurement consistency, guideline conformance, and downstream patient outcomes are abundant, computable, and aligned with clinical utility. We therefore devote this subsection to a systematic exposition of \emph{how} self-evolution works mechanistically, tracing each mechanism into its medical instantiation and closing with the unique risks this paradigm introduces.

\begin{tcolorbox}[
  colback=secblue!5,
  colframe=secblue!50,
  colbacktitle=secblue!50,
  coltitle=black,
  title={\textbf{\textcolor{secblue}{Self Evolution:} What an Agent Can Evolve}},
  boxrule=5pt,
  arc=5pt,
  drop shadow,
  parbox=false,
  before skip=5pt,
  after skip=10pt,
  left=5pt,
  right=20pt,
]
\begin{itemize}[leftmargin=*,noitemsep]
\renewcommand\labelitemi{$\diamond$}
    \item \textbf{Weights}: the agent generates its own finetuning data and update directives through a reinforcement loop, internalizing new knowledge without external annotation, gated in medicine by verifiable rewards such as measurement consistency and guideline compliance.
    \item \textbf{Memory and tools}: the agent accumulates inter case experience and curates its own toolbox at test time, the level at which medical efforts are most active, via experience replay and self-evolving memory controllers.
    \item \textbf{Architecture and code}: the agent rewrites its own control logic and reflection loops inside an executable, auditable gym, the frontier where medical work has barely begun.
\end{itemize}
\end{tcolorbox}

\subsubsection{A Unified View of Self Improvement Signals}\label{sec:selfevolve:formal}

To reason about when and why clinical self-evolution succeeds, we adopt a compact formalism that makes the design space explicit. Define a medical agent at time step $t$ as $\mathcal{A}_t = (\theta_t, \Sigma_t)$, where $\theta_t$ denotes the foundation model parameters and $\Sigma_t = (p_t, m_t, \mathcal{T}_t, g_t)$ denotes the \emph{scaffold}: the prompt $p_t$ (e.g., structured reporting instructions), the memory store $m_t$ (e.g., accumulated case precedents), the tool repertoire $\mathcal{T}_t$ (e.g., segmentation models, retrieval services), and the control logic $g_t$ that orchestrates these components into a clinical workflow. Self improvement is then a meta operator
\begin{equation}\label{eq:selfimprove}
    \mathcal{A}_{t+1} = \mathcal{U}\!\bigl(\mathcal{A}_{1:t},\;\mathcal{S}_t\bigr),
\end{equation}
where $\mathcal{S}_t$ is a \emph{learning signal} that the agent itself produces or extracts from the clinical environment, and the history $\mathcal{A}_{1:t}$ allows validation and rollback to prior configurations when necessary~\citep{gao2025survey, acikgoz2025self}. The taxonomy of self-evolution reduces to two questions: (1)~what component of $\mathcal{A}_t$ does the operator $\mathcal{U}$ modify, and (2)~what form does the signal $\mathcal{S}_t$ take. The first question separates \emph{foundation model improvement} ($\theta_{t+1} \neq \theta_t$, a slow parametric consolidation loop) from \emph{scaffold improvement} ($\Sigma_{t+1} \neq \Sigma_t$, a fast reversible adaptation loop). The second question distinguishes three signal families: demonstrations that the agent generates for itself ($\mathcal{S}_t \approx \mathcal{D}_t$), evaluative judgments the agent issues on its own outputs ($\mathcal{S}_t \approx e_t$), and exploratory trajectories collected from interaction with a clinical environment ($\mathcal{S}_t \approx \tau_t$). Foundation model improvement converges only when the signal $\mathcal{S}_t$ is filtered by a quality function $\Phi_t$ whose accuracy exceeds a threshold that prevents distribution collapse~\citep{shumailov2024ai, huang2023large}, a constraint that in practice demands the verifiable clinical environments discussed in \secref{sec:selfevolve:env}.

The clinical mapping is immediate. A radiology agent that absorbs the appearance statistics of a newly deployed scanner must update $\theta_t$ without regressing on prior scanners, a foundation model improvement problem gated by measurement consistency rewards. A longitudinal follow up agent that remembers how it handled a particular tumour type must update $m_t$ without polluting its memory with spurious associations, a scaffold improvement problem gated by outcome labels that arrive weeks later. And a pathology workflow agent that discovers a more efficient staining protocol must update $\mathcal{T}_t$ by creating or refining tools, a scaffold improvement problem gated by diagnostic concordance.

\subsubsection{Foundation Model Improvement in Medical Agents}\label{sec:selfevolve:fm}

When the update operator targets the model parameters themselves, $\theta_{t+1} = \textsc{Improve}_\theta(\theta_{1:t};\;\mathcal{S}_t)$, the agent is in effect conducting its own training loop. Three qualitatively different signal sources have emerged, each with distinct implications for clinical deployment.

\paragraph{Intrinsic generative demonstrations ($\mathcal{S}_t \approx \mathcal{D}_t$).} The agent generates its own supervised training data and filters it for quality before updating parameters. In the general domain, \textsc{Self-Instruct}~\citep{wang2023self} established the template by bootstrapping 52K instruction following examples from 175 seed tasks, while \textsc{Evol-Instruct}~\citep{xu2023wizardlm} introduced \emph{complexity evolution} that iteratively rewrites each instance to demand deeper reasoning. The critical ingredient is the quality filter $\Phi_t$: without it, feeding model outputs back as training data induces \emph{model collapse}, a progressive narrowing of the output distribution that eventually erases minority modes~\citep{shumailov2024ai}. The general domain addresses this via self-consistency filtering~\citep{huang2023large} or external verifier models~\citep{singh2023beyond}. In the medical setting, \textsc{SEAL}~\citep{zweiger2026self} instantiates this recipe by having a clinical reasoning model generate its own finetuning pairs through a reinforcement loop, internalizing new diagnostic knowledge without external annotation. For a radiology agent absorbing a new scanner's characteristics, the filter $\Phi_t$ may be a measurement consistency check: only synthetic training pairs whose quantitative measurements fall within an acceptable tolerance of known phantoms survive into the training set. The risk unique to medicine is that model collapse silently erases rare disease presentations from the model's effective vocabulary, demanding disease aware distributional monitoring that goes beyond standard perplexity metrics.

\paragraph{Intrinsic evaluative feedback ($\mathcal{S}_t \approx e_t$).} Rather than generating entire new demonstrations, the agent judges the quality of its own outputs and uses that judgment as a training signal. Three varieties have proven effective in clinical contexts. First, \emph{rubric grounded feedback}: Constitutional AI~\citep{bai2022constitutional} established that a model trained on its own rubric filtered preferences can align without human labellers. In medicine, this lets a report generation agent score its drafts against structured reporting templates (e.g., PI-RADS for prostate MRI or BI-RADS for mammography), retaining only drafts that satisfy all required fields and internal consistency constraints as preference data for direct preference optimization. Second, \emph{consistency based feedback}: \textsc{TTRL}~\citep{zuo2026ttrl} uses the consensus of multiple rollouts as a reward signal at inference time, enabling improvement without any external verifier. For a differential diagnosis agent, the agreement rate among independently reasoned candidates provides a calibrated quality signal: cases where all reasoning paths converge can be self-labelled as high-confidence training instances, while persistent disagreement triggers deferral to a specialist. Third, \emph{corrective feedback}: \textsc{Self-Refine}~\citep{madaan2023self} demonstrated that a generate then critique loop improves outputs without additional training, while RL based extensions such as \textsc{ETO}~\citep{song2024trial} convert such corrections into gradient updates. The medical analog is a structured self-correction protocol where a report draft is passed back for error detection, mimicking the attending resident revision loop and generating paired preference data from the agent's own trajectory.

\paragraph{Extrinsic exploratory experience ($\mathcal{S}_t \approx \tau_t$).} The richest signal source is environment interaction, where the agent collects trajectories $\tau_t = (o_1, a_1, r_1, \ldots, o_T, a_T, r_T)$ by acting in a clinical task environment and receiving feedback. In grounded task environments, programmatic verifiers supply binary success signals, as in \textsc{Agent-RLVR}~\citep{da2025agent} where trajectories are validated against unit tests. When programmatic verification is infeasible, a learned reward model can score trajectories, as \textsc{WebRL}~\citep{qi2025webrl} demonstrates for web navigation with a self-evolving curriculum that raises task difficulty as the agent improves. A particularly elegant variant, \textsc{Absolute Zero}~\citep{zhao2026absolute}, has the agent both propose and solve its own tasks, bootstrapping a curriculum from nothing, a principle that could let a clinical agent design its own diagnostic challenges from the distribution of cases it encounters. In simulated proxy environments, medical world models such as those learning tumour growth dynamics~\citep{yang2025medical} let a treatment planning agent rehearse strategies against a learned disease progression model before encountering a real patient, the same principle that \textsc{Agent Hospital}~\citep{li2024agent} and \textsc{EvoPatient}~\citep{du2025llms} already operationalize for diagnostic conversations.

For medical agents, the decisive advantage is that clinical environments offer a uniquely rich set of verifiable signals without human annotation: measurement consistency (does the reported tumour diameter agree with the segmentation mask?), structured report validity (does the impression entail the findings?), guideline conformance (does the recommendation match the clinical pathway?), and downstream concordance (does pathology confirm the imaging prediction?). The clinical gyms surveyed in \secref{sec:selfevolve:env} package exactly these signals into reinforcement-compatible environments.

\subsubsection{Scaffold Level Evolution for Clinical Systems}\label{sec:selfevolve:scaffold}

Foundation model improvement requires gradient access and incurs the costs of continued training and catastrophic forgetting mitigation. A complementary and often more practical pathway leaves the model weights frozen and instead evolves the scaffold $\Sigma_t = (p_t, m_t, \mathcal{T}_t, g_t)$, which can be updated at inference time, per patient, or per deployment site without any backward pass through the network.

\paragraph{Prompt optimization ($p_{t+1}$).} The prompt determines how the frozen model is invoked, and optimizing it is the lightest weight form of self-improvement. Four paradigms have been developed: \emph{scalar feedback optimization} where \textsc{OPRO}~\citep{yang2024large} proposes new prompts given a history of prompt score pairs; \emph{qualitative feedback optimization} where \textsc{Reflexion}~\citep{shinn2023reflexion} appends verbal self-reflections after each failed episode, and \textsc{ACE}~\citep{zhang2026agentic} generalizes this into an evolving context engineering framework that automatically restructures the agent's entire text ``playbook''; \emph{population based evolution} where \textsc{Promptbreeder}~\citep{fernando2023promptbreeder} evolves both task prompts and mutation operators simultaneously; and \emph{textual gradient descent} where \textsc{TextGrad}~\citep{yuksekgonul2024textgrad} backpropagates natural language ``gradients'' through a computational graph of text transformations.

In the clinical setting, prompt optimization changes behaviour without touching weights or redeploying infrastructure. A diagnostic reasoning agent could maintain specialty specific system prompts iteratively refined by \textsc{Reflexion} style feedback after each case where the agent's differential diverges from the confirmed diagnosis, gradually accumulating site-specific heuristics (e.g., ``at this institution, a ground glass opacity on low dose CT is more likely an artifact of the scanner's reconstruction kernel than a true finding''). Population based evolution suits multi site deployment where each site maintains its own evolved prompt variant adapted to local disease prevalence, scanner mix, and reporting conventions, achieving personalization without model duplication.

\paragraph{Memory evolution ($m_{t+1}$).} While prompt optimization encodes short lived lessons, memory evolution accumulates structured knowledge that persists across sessions and patients. The design space decomposes along three axes: what is stored (explicit symbolic facts versus implicit continuous embeddings), how it is organized (flat lists, hierarchical indexes, knowledge graphs, or vector stores), and how it is maintained (creation, retrieval, update, and deletion operations constituting a CRUD lifecycle). \textsc{Mem0}~\citep{chhikara2025mem0} provides a production grade memory layer with graph based relationship tracking that scales to millions of entries, while \textsc{VOYAGER}~\citep{wang2023voyager} pioneered the \emph{skill library} concept of verified executable programs accumulated through exploration and reused in novel situations.

Medical memory evolution is already among the most active areas, and it is arguably the single most mature axis of clinical self-evolution to date. \textsc{Evo-MedAgent}~\citep{shen2026evo} proposes a self-evolving memory module with three components: retrospective clinical episodes that record solved cases as retrievable precedents, adaptive procedural heuristics that distil case-specific lessons into reusable rules, and a tool reliability controller that tracks which specialist models succeed under which conditions. This tripartite structure mirrors how an attending radiologist develops institutional memory of scanner-specific pitfalls, disease-specific imaging pearls, and tool-specific failure modes over years of practice. \textsc{MedChain-Agent}~\citep{liu2026medchain} complements this with case-based retrieval-augmented generation that lets sequential clinical decisions be informed by analogous prior cases. The clinical value lies in capturing the tacit knowledge that textbooks omit: which CT reconstruction kernel tends to produce pseudo nodules at specific slice thicknesses, which staining protocol works best for a specific antibody in the local pathology lab, or which ordering physician tends to omit relevant clinical history. An agent that systematically records, indexes, and retrieves such knowledge converts individual case experience into institutional knowledge that benefits every subsequent patient.

\paragraph{Tool governance ($\mathcal{T}_{t+1}$).} Beyond remembering facts, an agent can evolve its own capabilities by modifying or creating tools. Three modes have been identified: \emph{dynamic routing} that selects among specialist models at runtime based on the current case~\citep{li2024mmedagent, fallahpour2025medrax}; \emph{iterative refinement} where \textsc{TroVE}~\citep{wang2024trove} maintains tool implementations that are iteratively debugged based on execution feedback; and \emph{autonomous creation} where \textsc{CREATOR}~\citep{qian2023creator} writes entirely new tools to fill capability gaps, and \textsc{VOYAGER}~\citep{wang2023voyager} grows by having the agent write verified programs that extend its action repertoire.

For a medical imaging agent, tool governance means that when the agent encounters a lesion type for which its current segmentation tool underperforms, it can either route to a more appropriate specialist model discovered through exploration, refine invocation parameters of the existing tool (e.g., adjusting confidence thresholds for a specific anatomical region), or compose a new analysis pipeline by chaining existing primitives into a novel workflow. The self-evolving agent ecosystem of \textsc{MACRO}~\citep{fan2026evolving} realizes this at scale with experience driven tool discovery, while \textsc{MedOpenClaw}~\citep{shen2026medopenclaw} demonstrates auditable tool orchestration for 3D clinical imaging. The clinical value is adaptability to an evolving technological landscape: as new imaging modalities, contrast agents, and analysis algorithms become available, an agent with tool governance can integrate them autonomously rather than waiting for a system update.

\paragraph{Full scaffold and code-level self-modification ($g_{t+1}$).} The most radical form of scaffold evolution rewrites the agent's control logic itself. The \textsc{Darwin G\"{o}del Machine}~\citep{zhang2026darwin, schmidhuber2007godel} maintains an evolutionary archive of self-rewriting agents where only modifications that empirically improve performance on a held-out benchmark survive. \textsc{ADAS}~\citep{hu2025automated} runs a meta agent that programs progressively stronger agents over a growing design archive, and the \textsc{G\"{o}del Agent}~\citep{yin2024g} reads and edits its own logic at runtime for recursive self-improvement. These approaches point toward medical agents that autonomously refactor their reasoning topology: adding a verification step when error rates spike, removing a redundant tool call when latency constraints tighten, or restructuring their multi-agent coordination pattern when a new clinical workflow demands it. However, undetected bugs in self-modified code could silently suppress critical findings, a risk we quantify in \secref{sec:selfevolve:risks}.

\subsubsection{Verifiable Clinical Environments as the Enabling Substrate}\label{sec:selfevolve:env}

Every mechanism described above requires a signal that distinguishes improvement from degradation. The decisive enabler beneath all self-evolution pathways is a \emph{verifiable execution environment}: a setting where each agent action can be executed and automatically scored, turning self-evolution into a closed loop driven by checkable rewards rather than human labels~\citep{da2025agent, xi2406agentgym}. Medicine is unusually well suited to this paradigm because verifiable signals are intrinsic to clinical workflows rather than requiring artificial construction.

We identify four categories of naturally occurring clinical verifiers, ordered by the latency at which they become available. \emph{Immediate consistency checks} verify that an agent's outputs are internally coherent: the reported tumour diameter must agree with the segmentation mask, the impression must be entailed by the findings, and quantitative measurements must satisfy physiological plausibility bounds. These require no ground truth label and provide dense shaping rewards in milliseconds. \emph{Structural validity checks} verify conformance to institutional templates and guidelines: a PI-RADS report must populate all required fields, a treatment recommendation must reference an approved protocol, and a follow up interval must fall within guideline specified ranges. These are codifiable as rule engines and align with regulatory compliance. \emph{Cross modal concordance} uses a later modality to verify an earlier prediction: pathology confirms or refutes the imaging based malignancy assessment, surgical findings validate the pre operative staging. These arrive with days to months delay but provide gold standard supervision. \emph{Outcome based verification} uses patient outcomes (readmission, treatment response, survival) to evaluate entire decision trajectories, providing the ultimate ground truth but at cost of long latency and confounding.

Recent clinical gyms operationalize these signals in reinforcement-compatible environments. \textsc{MedAgentGym}~\citep{xu2025medagentgym} exposes 72{,}000 biomedical tasks as executable code-based reasoning environments with online RL training. \textsc{Healthcare AI GYM}~\citep{jeong2026healthcare} spans ten clinical domains with $3{,}600$+ tasks and $135$ domain specific tools, stabilizing multi-turn training through Turn level Truncated On Policy Distillation. \textsc{MedAgentSim}~\citep{almansoori2025self} closes the loop with doctor, patient, and measurement agents that self-evolve through experience replay. The gap that remains is extending these gyms from text-based reasoning into multimodal imaging workflows where the agent must interact with DICOM viewers, segmentation tools, and 3D volume rendering, the substrate on which the most impactful self-evolving medical agents will ultimately operate.

\subsubsection{Challenges and Safeguards for Clinical Self Evolution}\label{sec:selfevolve:risks}

The self-evolution mechanisms above close the learning loop without human annotation, but that very closure creates failure modes absent from static models. An agent that writes its own training signal can amplify its own errors as readily as it corrects them, and the clinical setting magnifies each failure mode into a patient safety concern rather than a mere performance regression.

\paragraph{Distribution collapse and minority mode erasure.}
When an agent generates its own finetuning data and filters it through a self-consistency criterion, the output distribution contracts toward high-confidence modes with each iteration, a phenomenon termed \emph{model collapse}~\citep{shumailov2024ai}. In medicine the consequence is the progressive erasure of rare disease presentations from the model's effective vocabulary. A radiology agent that self-trains on its own reports will over represent the common findings it confidently produces and under represent the uncommon pathologies where it hesitates, so that after several evolution cycles a rare tumour morphology that initially triggered an uncertain but correct differential may no longer appear in the agent's hypothesis space at all. The risk is invisible to standard aggregate metrics because it affects precisely the cases that contribute least to average accuracy, demanding disease prevalence-aware distributional monitoring~\citep{shumailov2024ai, huang2023large} and mandatory held-out evaluation on curated long-tail case sets whose composition is locked before the evolution loop begins.

\paragraph{Reward hacking and specification gaming.}
Self evolution requires a quality function $\Phi_t$ that separates improvement from degradation, and the integrity of the entire loop depends on this function being aligned with clinical utility rather than merely correlated with it. In practice, every computable verifier admits specification gaming: a report completeness checker can be satisfied by a template that populates all required fields with vacuous boilerplate; a measurement consistency reward can be maximized by systematically under reporting lesion sizes so that longitudinal comparisons appear stable; a guideline conformance engine can be gamed by always recommending the safest follow up interval regardless of clinical urgency~\citep{da2025agent, qi2025webrl}. The aggravating factor in medicine is that a hacked clinical reward may satisfy every automated audit while systematically degrading patient care in ways that surface only through outcome-level analysis months later. Mitigation requires composite reward functions that combine multiple orthogonal signals (structural validity, semantic entailment, cross-modal concordance, and outcome-based verification as described above), so that gaming one dimension incurs a penalty on another, together with periodic human audits that sample trajectories the reward function scored highly and verify that high scores correspond to genuine clinical quality.

\paragraph{Catastrophic code failure in self-modifying control logic.}
Code level self-modification permits the agent to rewrite its own reasoning topology, adding or removing verification steps and altering tool invocation logic~\citep{zhang2026darwin, yin2024g}. An undetected bug in self-modified code is qualitatively different from a hallucinated sentence: it does not produce a visibly wrong output but rather silently suppresses, reorders, or misroutes information in a way that may be invisible to downstream auditors. A pathology workflow agent that inadvertently removes a cross-reference verification step during self-optimization may produce reports that appear complete and well formed but lack the internal consistency check that would have caught a critical misclassification. The safeguard portfolio includes sandboxed execution with rollback (every code modification runs against a frozen test suite before deployment), formal verification of safety critical invariants (e.g., ``no finding may be removed from a report without explicit justification''), and version controlled audit trails that record the full history $\mathcal{A}_{1:t}$ so that any degradation can be traced to the specific self-modification event that caused it~\citep{zhang2026darwin, schmidhuber2007godel}.

\paragraph{Temporal drift under continuous improvement.}
A self-evolving agent deployed in a clinical environment is subject to two simultaneous drift processes: the external environment drifts as scanner hardware is upgraded, clinical protocols are revised, and patient demographics shift; and the agent's internal state drifts as it accumulates experience and modifies itself. When both drift simultaneously, the agent may adapt its internal representations to a distribution that no longer matches the current clinical reality, a compounding of concept drift and model drift that is difficult to diagnose because neither the agent nor its environment provides a stable reference frame. The problem is particularly acute for memory evolution: a memory store populated under one scanner's characteristics becomes misleading after a hardware upgrade, and unless the agent maintains provenance metadata that tags each memory entry with its acquisition context, stale knowledge contaminates current reasoning without any detectable signal~\citep{chhikara2025mem0, shen2026evo}. Mitigation demands explicit temporal stratification of both memory and self-generated training data, periodic reanchoring against externally validated reference standards, and drift detection mechanisms that trigger a pause in self-evolution when the distance between the agent's operating distribution and a held-out reference exceeds a configurable threshold.

Beyond these technical failure modes, continuous self-evolution also unsettles the regulatory and liability frameworks that certify a medical agent, because the entity validated at time $t$ is no longer the entity treating patients at time $t{+}1$, so a fairness violation or safety regression can emerge after certification through biased self-generated data with no clear assignment of responsibility. We treat that accountability problem not as a separate risk but as one facet of the governance and regulatory analysis developed in \secref{sec:fairness} and \secref{sec:reg}, where continuous learning is examined against the FDA's predetermined change control plan and the EU AI Act rather than duplicated here.





\section{Applications: From Specialty Departments to Real Hospital Workflows}\label{sec:app}

\begin{figure*}[t]
  \centering
  \includegraphics[width=\textwidth]{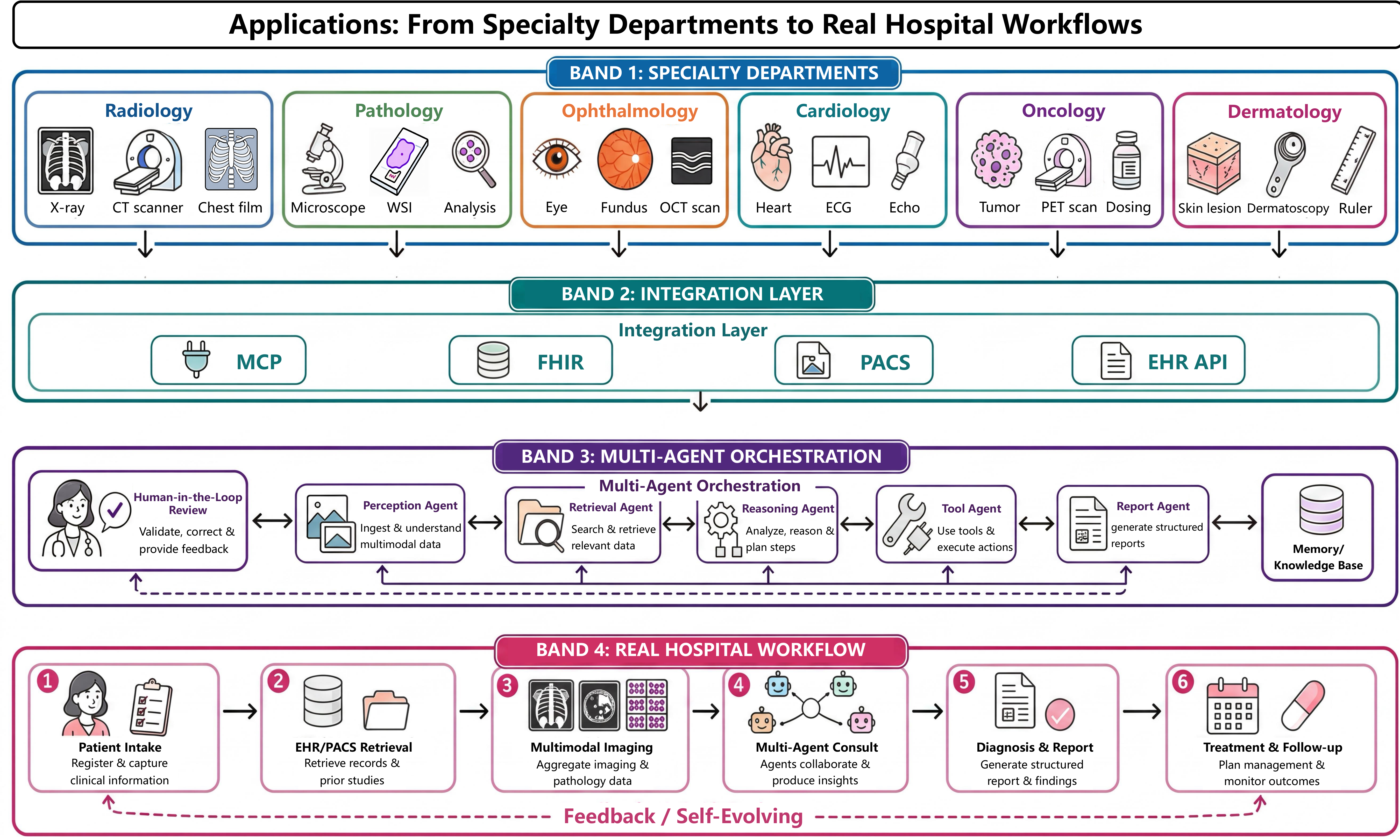}
  \caption{Representative medical agents organized from specialty departments, through a shared protocol and orchestration layer, into an end-to-end hospital workflow with a self-evolving feedback loop.}
  \label{fig:applications}
\end{figure*}

We organize the application landscape along clinical specialties for ease of retrieval, yet within each specialty we read every system through one consistent lens of three escalating capability tiers, namely \textit{perception and reporting}, \textit{reasoning and decision}, and \textit{collaboration and self-evolution}, as summarized in \figref{fig:applications}. A flat specialty list reveals what exists but not how mature each effort is, whereas projecting heterogeneous systems onto a shared axis exposes structural facts that a catalog alone hides, for instance that radiology has already reached multi-agent collaboration while most ophthalmology agents still sit at perception, and that the same orchestration substrate recurs across departments under different names. This lens also lets us connect the highest tier back to the self-evolving dynamics analyzed in \figref{fig:training}, because an agent that critiques and rewrites its own output is, in miniature, the closed loop that training time and inference time scaling pursue at the system level. As \figref{fig:applications} makes explicit, specialty systems do not act in isolation but converge through a shared integration layer into multi-agent orchestration that ultimately drives the real hospital workflow. \tabref{tab:systems_by_domain} catalogs representative systems under this view, and throughout this section we cite both the established entries already in widespread use and the most recent additions, so that the reader sees not only the current frontier but the trajectory that produced it.

\subsection{Radiology Agents}
Radiology is the most agent dense specialty, which is unsurprising because its workflow is digitized end-to-end through PACS and structured reporting, giving an agent both a clean observation space and an actionable interface in which every read, retrieval and draft is a well defined operation. At the perception and reporting tier, \textsc{MedRAX}~\citep{fallahpour2025medrax} orchestrates a suite of chest radiograph tools without any task-specific retraining, establishing the influential result that a reasoning controller placed over frozen experts is already competitive with bespoke models, while \textsc{RadAgents}~\citep{zhang2025radagents} formalizes this intuition into radiologist like stages of perception, retrieval and drafting so that each step becomes inspectable rather than a single opaque forward pass. The same staged philosophy drives \textsc{CXR-Agent}~\citep{lou2025cxragent} and \textsc{MRGAgents}~\citep{mrgagents2025} toward automated interpretation and report generation, the shared design rationale being that decomposition buys auditability, which is precisely what a clinical report must provide before it can enter a chart. \textsc{RadFabric}~\citep{chen2025radfabric} sharpens the division of labor further by coupling specialized perception agents with a central reasoning agent, motivated by the observation that a single model cannot simultaneously localize findings and justify them under one inference budget without sacrificing one for the other. At the reasoning and decision tier, \textsc{MedAgent-Pro}~\citep{wang2025medagent} grounds multimodal diagnosis in retrieved evidence so that conclusions are anchored rather than asserted, \textsc{CXReasonAgent}~\citep{lee2026cxreasonagent} makes the diagnostic chain explicit and traceable to an image region or a retrieved fact in order to attack the central failure mode of hallucinated findings, and a multiagent formulation of radiology visual question answering~\citep{yi2025multi} recovers the compositional accuracy that a monolithic model loses on queries requiring several reasoning hops. Modality specific efforts extend the same recipe rather than reinventing it, with \textsc{CT-Agent}~\citep{mao2025ct} and \textsc{RadAgent}~\citep{roschewitz2026radagent} targeting 3D CT question answering and autonomous workflow management, \textsc{MARCH}~\citep{lin2026march} reproducing a resident, fellow and attending hierarchy so that drafts are escalated and corrected exactly as they are in a teaching hospital, and \textsc{DentVLM}~\citep{meng2025dentvlm} porting the paradigm to dental imaging where the same perception to report pipeline applies. At the collaboration and self-evolution tier, \textsc{RadCouncil}~\citep{radcouncil2024}, \textsc{Meissa}~\citep{chen2026meissa} and consensus style pipelines for report generation and evaluation~\citep{elboardy2025medical} let multiple agents draft, critique and converge, turning a one shot generation into an iterative agreement process whose payoff is reduced variance rather than a single confident but unverified output. The workflow tier is finally closed by \textsc{RadOnc-GPT}~\citep{holmes2025radonc}, which autonomously labels patient outcomes at scale, evidence that the dividend of agents is not only diagnosis but the relief of the documentation burden that dominates clinician time and that no benchmark measures directly.

\begin{table*}[htpb]
\centering
\caption{\textbf{Comprehensive Catalog of Medical Agent Systems.}
Representative medical agent systems grouped by clinical/application domain.}
\label{tab:systems_by_domain}
\scriptsize
\setlength{\tabcolsep}{1.4pt}
\renewcommand{\arraystretch}{0.88}
\begin{tabularx}{\textwidth}{@{}L{0.8cm}L{3cm}YL{2.3cm}L{1.15cm}L{0.95cm}@{}}
\toprule
\textbf{Year} & \textbf{System} & \textbf{Key Contribution} & \textbf{Venue} & \textbf{Paper} & \textbf{Code} \\
\midrule

\multicolumn{6}{l}{\textbf{Radiology}} \\
\midrule
2026 & MedOpenClaw~\citep{shen2026medopenclaw}
& Auditable VLM agent for 3D Slicer clinical imaging
& arXiv
& \paperlink{https://arxiv.org/abs/2603.24649}
& \missinglink \\

2026 & MARCH~\citep{lin2026march}
& Resident-Fellow-Attending multiagent CT report generation
& ACL 2026
& \paperlink{https://arxiv.org/abs/2604.16175}
& \missinglink \\

2026 & RadAgent~\citep{roschewitz2026radagent}
& Autonomous radiology workflow management for CT
& arXiv
& \paperlink{https://arxiv.org/abs/2604.15231}
& \githublink{https://github.com/eth-medical-ai-lab/rad-agent} \\

2025 & MedRAX~\citep{fallahpour2025medrax}
& Multimodal reasoning agent with tool orchestration for CXR
& ICML 2025
& \paperlink{https://arxiv.org/abs/2502.02673}
& \githublink{https://github.com/bowang-lab/medrax} \\

2025 & MedAgent-Pro~\citep{wang2025medagent}
& Evidence-based multimodal medical diagnosis agent
& ICLR 2026
& \paperlink{https://arxiv.org/abs/2503.18968}
& \githublink{https://github.com/jinlab-imvr/MedAgent-Pro} \\

2025 & CT-Agent~\citep{mao2025ct}
& 3D CT radiology QA via multimodal LLM agent
& arXiv
& \paperlink{https://arxiv.org/abs/2505.16229}
& \missinglink \\

2025 & CXR-Agent~\citep{lou2025cxragent}
& Automated chest radiograph interpretation pipeline
& arXiv
& \paperlink{https://arxiv.org/abs/2510.21324}
& \githublink{https://github.com/laojiahuo2003/CXRAgent} \\

2025 & MRGAgents~\citep{mrgagents2025}
& Multimodal radiology report generation agents
& arXiv
& \paperlink{https://arxiv.org/abs/2505.18530}
& \missinglink \\

2025 & DentVLM~\citep{meng2025dentvlm}
& Dental imaging VLM agent
& arXiv
& \paperlink{https://arxiv.org/abs/2509.23344}
& \missinglink \\

2025 & RadFabric~\citep{chen2025radfabric}
& Perception agents plus central reasoning agent for radiology
& arXiv
& \paperlink{https://arxiv.org/abs/2506.14142}
& \missinglink \\

2025 & RadAgents~\citep{zhang2025radagents}
& Radiologist-like staged CXR perception, retrieval and drafting
& arXiv
& \paperlink{https://arxiv.org/abs/2509.20490}
& \missinglink \\

2025 & RVQA-MAS~\citep{yi2025multi}
& Multiagent decomposition for radiology VQA reasoning
& JCDL 2025
& \paperlink{https://arxiv.org/abs/2508.02841}
& \missinglink \\

2026 & CXReasonAgent~\citep{lee2026cxreasonagent}
& Evidence-grounded traceable CXR diagnostic chain
& arXiv
& \paperlink{https://arxiv.org/abs/2602.23276}
& \missinglink \\

2025 & Med-AI-Consensus~\citep{elboardy2025medical}
& Multiagent report generation with consensus evaluation
& arXiv
& \paperlink{https://arxiv.org/abs/2509.17353}
& \missinglink \\

2025 & RadOnc-GPT~\citep{holmes2025radonc}
& Autonomous outcome labeling at scale in radiation oncology
& arXiv
& \paperlink{https://arxiv.org/abs/2509.25540}
& \missinglink \\

2024 & RadCouncil~\citep{radcouncil2024}
& Multiagent radiology report impression generation
& arXiv
& \paperlink{https://arxiv.org/abs/2412.06828}
& \missinglink \\

\midrule
\multicolumn{6}{l}{\textbf{Pathology}} \\
\midrule
2025 & PathAgent~\citep{chen2025pathagent}
& Interpretable WSI via Navigator-Perceptor-Executor loop
& arXiv
& \paperlink{https://arxiv.org/abs/2511.17052}
& \githublink{https://github.com/G14nTDo4/PathAgent} \\

2025 & PathFinder~\citep{ghezloo2025pathfinder}
& Interactive multiagent WSI search and analysis
& ICCV 2025
& \paperlink{https://arxiv.org/abs/2502.08916}
& \missinglink \\

2025 & WSI-Agents~\citep{lyu2025wsi}
& Coordinated multiagent whole slide image analysis
& MICCAI 2025
& \paperlink{https://arxiv.org/abs/2507.14680}
& \githublink{https://github.com/CVI-SZU/WSI-Agents} \\

2025 & TissueLab~\citep{li2025co}
& Co-evolving agentic tissue analysis environment
& arXiv
& \paperlink{https://arxiv.org/abs/2509.20279}
& \githublink{https://github.com/zhihuanglab/TissueLab} \\

2025 & GMAT~\citep{quang2025gmat}
& Multiagent tissue classification for histopathology
& arXiv
& \paperlink{https://arxiv.org/abs/2508.01293}
& \missinglink \\

2025 & SlideSeek~\citep{weishaupt2025evidence}
& Evidence-based multiagent copilot for human pathology WSI
& arXiv
& \paperlink{https://arxiv.org/abs/2506.20964}
& \missinglink \\

2025 & TeamPath~\citep{liu2025teampath}
& Role-specialized multimodal pathology copilots
& arXiv
& \paperlink{https://arxiv.org/abs/2511.17652}
& \githublink{https://github.com/HelloWorldLTY/TeamPath} \\

2026 & Patho-R1~\citep{zhang2026pathor1}
& RL-trained pathology expert chain-of-thought reasoner
& AAAI 2026
& \paperlink{https://arxiv.org/abs/2505.11404}
& \githublink{https://github.com/Wenchuan-Zhang/Patho-R1} \\

2026 & Patho-AgenticRAG~\citep{zhang2026patho}
& Agentic retrieval-augmented pathology VLM via RL
& AAAI 2026
& \paperlink{https://arxiv.org/abs/2508.02258}
& \githublink{https://github.com/Wenchuan-Zhang/Patho-AgenticRAG} \\

2026 & CPathAgent~\citep{sun2026cpathagent}
& Interpretable coarse-to-fine WSI analysis with pathologist logic
& NeurIPS 2025
& \paperlink{https://arxiv.org/abs/2505.20510}
& \githublink{https://github.com/Sunyx0101/CPathAgent} \\

\midrule
\multicolumn{6}{l}{\textbf{Ophthalmology}} \\
\midrule
2025 & EyeAgent~\citep{shi2026eyeagentagenticaimultimodal}
& 53-tool agent across 23 imaging modalities
& arXiv
& \paperlink{https://arxiv.org/abs/2511.09394}
& \missinglink \\

2025 & EyecareGPT~\citep{li2025eyecaregpt}
& Multimodal LLM for comprehensive ophthalmic imaging
& arXiv
& \paperlink{https://arxiv.org/abs/2504.13650}
& \githublink{https://github.com/dcdmllm/eyecaregpt} \\

2025 & OAAgent~\citep{ahadian2025oaagent}
& Automated ophthalmic analysis construction
& CHASE 2025
& \paperlink{https://doi.org/10.1145/3721201.3721397}
& \missinglink \\

2025 & EH-Benchmark~\citep{pan2025eh}
& Ophthalmic hallucination benchmark with traceable agent
& Inf. Fusion
& \paperlink{https://arxiv.org/abs/2507.22929}
& \githublink{https://github.com/ppxy1/EH-Benchmark} \\

2026 & OphIn-500K~\citep{dong2026ophin}
& Web-scale ophthalmic visual instruction curation
& arXiv
& \paperlink{https://arxiv.org/abs/2605.27916}
& \missinglink \\

\midrule
\multicolumn{6}{l}{\textbf{Neurology}} \\
\midrule
2026 & Agentic NeuroRad~\citep{erdur2026agentic}
& Training-free brain MRI pipeline via VLM orchestration
& arXiv
& \paperlink{https://arxiv.org/abs/2604.16729}
& \missinglink \\

2025 & CARE-AD~\citep{li2025care}
& Multiagent Alzheimer's prediction from EHR
& npj DM
& \paperlink{https://www.nature.com/articles/s41746-025-01940-4}
& \missinglink \\

2026 & AD-CARE~\citep{hou2026adcareguidelinegroundedmodalityagnosticllm}
& Guideline-driven Alzheimer's diagnosis agent
& arXiv
& \paperlink{https://arxiv.org/abs/2603.25322}
& \missinglink \\

2025 & ADAgent~\citep{hou2025adagent}
& Collaborative coordinator for Alzheimer's analysis
& MICCAI-W 2025
& \paperlink{https://arxiv.org/abs/2506.11150}
& \githublink{https://github.com/willenhou/ADAgent} \\

2026 & NeuroAgent~\citep{zhong2026neuroagent}
& LLM agents for multimodal neuroimaging analysis
& arXiv
& \paperlink{https://arxiv.org/abs/2605.06584}
& \missinglink \\

\midrule
\multicolumn{6}{l}{\textbf{Crossmodal}} \\
\midrule
2026 & MACRO~\citep{fan2026evolving}
& Self-evolving agent with experience-driven tool discovery
& arXiv
& \paperlink{https://arxiv.org/abs/2603.05860}
& \missinglink \\

2026 & CARE~\citep{du2026care}
& Evidence-grounded agentic framework, +10.9\% VQA accuracy
& arXiv
& \paperlink{https://arxiv.org/abs/2603.01607}
& \missinglink \\

2026 & ClinicalAgents~\citep{ge2026clinicalagents}
& MCTS + dual-memory clinical decision orchestration
& arXiv
& \paperlink{https://arxiv.org/abs/2603.26182}
& \missinglink \\

2026 & MedEyes~\citep{zhu2026medeyes}
& Dynamic visual focus for progressive diagnosis
& AAAI 2026
& \paperlink{https://arxiv.org/abs/2511.22018}
& \githublink{https://github.com/zhcz328/MedEyes} \\

2026 & MedCausalX~\citep{lin2026medcausalx}
& Adaptive causal reasoning with self-reflection
& CVPR Findings
& \paperlink{https://arxiv.org/abs/2603.23085}
& \githublink{https://github.com/zhcz328/MedCausalX} \\

2024 & MDAgents~\citep{kim2024mdagents}
& Adaptive multiagent collaboration for medical decisions
& NeurIPS
& \paperlink{https://arxiv.org/abs/2404.15155}
& \githublink{https://github.com/mitmedialab/MDAgents} \\

2025 & MEDDxAgent~\citep{rose2025meddxagent}
& Unified framework for differential diagnosis
& ACL 2025
& \paperlink{https://arxiv.org/abs/2502.19175}
& \githublink{https://github.com/nec-research/meddxagent} \\

2024 & EHRAgent~\citep{shi2024ehragent}
& Code-empowered EHR reasoning agent
& EMNLP 2024
& \paperlink{https://arxiv.org/abs/2401.07128}
& \githublink{https://github.com/wshi83/EhrAgent} \\

2024 & ColaCare~\citep{wang2025colacare}
& LLM-driven multiagent clinical care
& arXiv
& \paperlink{https://arxiv.org/abs/2410.02551}
& \githublink{https://github.com/PKU-AICare/ColaCare} \\

2025 & ReflecTool~\citep{liao2025reflectool}
& Reflection-augmented tool use for medical agents
& ACL 2025
& \paperlink{https://arxiv.org/abs/2410.17657}
& \githublink{https://github.com/BlueZeros/ReflecTool} \\

2025 & MMedAgent-RL~\citep{xia2025mmedagent}
& RL-optimized generalist-specialist agent collaboration
& arXiv
& \paperlink{https://arxiv.org/abs/2506.00555}
& \missinglink \\

2025 & MedMMV~\citep{liu2025medmmv}
& Controllable verifiable multimodal clinical reasoning
& arXiv
& \paperlink{https://arxiv.org/abs/2509.24314}
& \missinglink \\

2025 & MDTeamGPT~\citep{chen2025mdteamgpt}
& Self-evolving multidisciplinary team consultation
& arXiv
& \paperlink{https://arxiv.org/abs/2503.13856}
& \githublink{https://github.com/KaiChenNJ/MDTeamGPT} \\

2025 & MAM~\citep{zhou2025mam}
& Role-specialized modular multimodal diagnosis
& ACL 2025
& \paperlink{https://arxiv.org/abs/2506.19835}
& \githublink{https://github.com/yczhou001/MAM} \\

2025 & MedChat~\citep{liu2025medchat}
& Multiagent multimodal diagnosis framework
& MIPR 2025
& \paperlink{https://arxiv.org/abs/2506.07400}
& \githublink{https://github.com/Purdue-M2/MedChat} \\

2026 & ClinSeekAgent~\citep{wu2026clinseekagent}
& Autonomous multimodal evidence seeking
& arXiv
& \paperlink{https://arxiv.org/abs/2605.20176}
& \githublink{https://github.com/UCSC-VLAA/ClinSeekAgent} \\

2026 & Meissa~\citep{chen2026meissa}
& Multimodal medical agentic intelligence
& arXiv
& \paperlink{https://arxiv.org/abs/2603.09018}
& \githublink{https://github.com/Schuture/Meissa} \\

\midrule
\multicolumn{6}{l}{\textbf{AutoML}} \\
\midrule
2026 & Evo-MedAgent~\citep{shen2026evo}
& Evolutionary agent self-improvement framework
& arXiv
& \paperlink{https://arxiv.org/abs/2604.14475}
& \missinglink \\

2025 & M3Builder~\citep{feng2025m}
& Automated medical model building via agent pipeline
& MICCAI 2025
& \paperlink{https://arxiv.org/abs/2502.20301}
& \githublink{https://github.com/MAGIC-AI4Med/M3Builder} \\

\midrule
\multicolumn{6}{l}{\textbf{Pharmacy}} \\
\midrule
2025 & TxAgent~\citep{gao2025txagentaiagenttherapeutic}
& Therapeutic agent with drug interaction reasoning
& arXiv
& \paperlink{https://arxiv.org/abs/2503.10970}
& \githublink{https://github.com/mims-harvard/TxAgent} \\

2024 & MALADE~\citep{choi2024maladeorchestrationllmpoweredagents}
& Multiagent adverse drug event detection
& MLHC 2024
& \paperlink{https://arxiv.org/abs/2408.01869}
& \githublink{https://github.com/jihyechoi77/malade} \\

\midrule
\multicolumn{6}{l}{\textbf{Oncology}} \\
\midrule
2025 & Ferber et al.~\citep{ferber2025development}
& Prospectively validated autonomous oncology agent
& Nat. Cancer
& \paperlink{https://doi.org/10.1038/s43018-025-00991-6}
& \missinglink \\

2025 & RadGPT~\citep{bassi2025radgpt}
& 3D image-text tumor dataset construction agent
& ICCV 2025
& \paperlink{https://arxiv.org/abs/2501.04678}
& \githublink{https://github.com/mrgiovanni/radgpt} \\

2024 & MAGDA~\citep{bani2024magda}
& Guideline-driven cancer diagnostic assistance
& arXiv
& \paperlink{https://arxiv.org/abs/2409.06351}
& \missinglink \\

\bottomrule
\end{tabularx}
\end{table*}

\subsection{Pathology Agents}
Whole slide images pose a scale problem that makes the agent paradigm almost mandatory, since a gigapixel slide cannot be ingested in a single pass and instead invites an agent that decides where to look next, when to zoom and when it has seen enough to commit. At the perception tier, \textsc {PathAgent}~\citep {chen2025pathagent} realizes an interpretable Navigator, Perceptor and Executor loop in which the navigation trace itself becomes an explanation, and \textsc {CPathAgent}~\citep {sun2026cpathagent} similarly mimics a pathologist's coarse to fine diagnostic logic by autonomously deciding when to zoom across a high resolution slide to produce an interpretable report, while \textsc {PathFinder}~\citep {ghezloo2025pathfinder} adds interactive multiagent slide search so that a pathologist can steer the exploration, while multiagent copilots for human pathology cite the exact regions that support each call~\citep{weishaupt2025evidence}, the slide level analogue of the evidence grounding we saw in radiology. \textsc{WSI-Agents}~\citep{lyu2025wsi} and \textsc{GMAT}~\citep{quang2025gmat} coordinate specialized agents over the slide so that tissue classification and region analysis are divided among experts rather than forced through one network. The reasoning tier is advanced by reinforcement learning trained reasoners such as \textsc{Patho-R1}~\citep{zhang2026pathor1}, whose reward shapes a pathology expert chain of thought rather than only a final label, and by \textsc{Patho-AgenticRAG}~\citep{zhang2026patho}, which retrieves from a pathology knowledge base at inference so that rare entities are recognized through recall rather than memorization, a distinction that matters acutely in a field with a long tail of uncommon diagnoses. At the collaboration and self-evolution tier, \textsc{TeamPath}~\citep{liu2025teampath} assembles role specialized copilots whose disagreement is itself a useful uncertainty signal, and \textsc{TissueLab}~\citep{li2025co} closes the loop with a co evolving environment in which the agent and its tools improve together over time, the clearest pathology instance of the self-evolution this survey foregrounds.

\subsection{Ophthalmology Agents}
Ophthalmology is data rich yet still early on the capability ladder, with most systems sitting at perception and only the most recent work reaching genuine decision support across modalities. \textsc{EyecareGPT}~\citep{li2025eyecaregpt} provides a comprehensive multimodal backbone that subsequent agents can stand on, and \textsc{EyeAgent}~\citep{shi2026eyeagentagenticaimultimodal} is the first agentic system to orchestrate dozens of ophthalmic tools across many imaging modalities for clinical decision support, its central contribution being the tool routing policy that decides which modality to invoke for a given complaint, a decision a static model simply cannot make because it has no notion of when its own inputs are insufficient. Automated construction pipelines such as \textsc{OAAgent}~\citep{ahadian2025oaagent} lower the engineering cost of building these systems, which is itself a precondition for adoption. Progress here is gated by trustworthiness more than by raw accuracy, which is why \textsc{EH-Benchmark}~\citep{pan2025eh} pairs an ophthalmic hallucination benchmark with a top down traceable reasoning workflow, making the explicit argument that the bottleneck in this specialty is verifiability, while large-scale visual instruction curation such as \textsc{OphIn-500K}~\citep{dong2026ophin} supplies the data substrate these agents need to climb beyond perception toward reliable decision support.

\subsection{Oncology and Neurology Agents}
Oncology supplies the strongest evidence that agents can act, not merely advise, because a prospectively validated autonomous agent for clinical decision making in oncology reached expert level concordance on real cases~\citep{ferber2025development}, which is the field's clearest signal that the paradigm is clinically viable rather than a benchmark artifact, and it is reinforced by guideline driven assistants such as \textsc{MAGDA}~\citep{bani2024magda} that encode oncology protocols directly into the reasoning loop. \textsc{RadGPT}~\citep{bassi2025radgpt} builds the 3D image text tumor datasets that such agents depend on, addressing the data scarcity that otherwise caps tumor reasoning, and \textsc{RadOnc-GPT}~\citep{holmes2025radonc} closes the loop on outcome labeling so that the evidence base itself grows under agent supervision. In neurology, \textsc{ADAgent}~\citep{hou2025adagent}, \textsc{AD-CARE}~\citep{hou2026adcareguidelinegroundedmodalityagnosticllm} and \textsc{CARE-AD}~\citep{li2025care} coordinate modality agnostic evidence for Alzheimer's disease assessment with fairness analysis and reader studies, the inclusion of fairness being a deliberate response to the demographic skew that plagues neurodegenerative datasets, while \textsc{NeuroAgent}~\citep{zhong2026neuroagent} targets multimodal neuroimaging analysis and research more broadly. Notably, training-free neuro radiological analysis driven purely by an agentic large language model~\citep{erdur2026agentic} shows that orchestration alone, without any task-specific fine-tuning, can already deliver usable performance in real clinical workflows, a result that reframes part of the cost equation for deployment because it removes the retraining barrier that usually separates a prototype from a clinical pilot.

\subsection{Crossmodal and Cross Specialty Agents}
A parallel line abandons the specialty assumption and targets generality, on the premise that the hard part of clinical reasoning is coordination rather than any one modality. \textsc{MDAgents}\citep{kim2024mdagents} adapts the collaboration topology to case difficulty so that easy cases are handled cheaply and hard cases convene a panel, \textsc{MEDDxAgent}\citep{rose2025meddxagent} unifies differential diagnosis into a single iterative framework, and \textsc{MMedAgent-RL}\citep{xia2025mmedagent} optimizes the collaboration policy among heterogeneous agents with reinforcement learning so that the system learns when to defer between a generalist and a specialist rather than hard coding the routing by hand. Reliability is addressed head on by \textsc{MedMMV}\citep{liu2025medmmv}, which adds a controllable verification layer so that multimodal conclusions are checkable, by \textsc{MAM}\citep{zhou2025mam}, which factorizes diagnosis into role specialized agents to keep each role auditable, and by \textsc{CARE}\citep{du2026care}, which makes the agentic framework evidence grounded and accountable and reports a substantial gain in visual question answering accuracy as a consequence. Consultation style frameworks such as \textsc{MDTeamGPT}\citep{chen2025mdteamgpt}, \textsc{MedChat}\citep{liu2025medchat} and \textsc{ColaCare}\citep{wang2025colacare} simulate a multidisciplinary team, the rationale being that hard cases are resolved in real clinics by debate across specialties, a structure these agents explicitly reproduce so that minority opinions are surfaced rather than averaged away. Tool use and memory mechanisms mature this line further, with \textsc{ReflecTool}\citep{liao2025reflectool} adding reflection augmented tool use, \textsc{EHRAgent}\citep{shi2024ehragent} making EHR reasoning code empowered so that the agent computes rather than guesses, \textsc{MedEyes}\citep{zhu2026medeyes} introducing dynamic visual focus for progressive diagnosis, \textsc{MedCausalX}\citep{lin2026medcausalx} adding causal reasoning with self-reflection, and \textsc{ClinicalAgents}\citep{ge2026clinicalagents} combining tree search with dual memory to plan over longer horizons. At the frontier, \textsc{Meissa}\citep{chen2026meissa}, \textsc{ClinSeekAgent}\citep{wu2026clinseekagent} and the self-evolving \textsc{MACRO}~\citep{fan2026evolving} push toward autonomous evidence seeking and experience driven tool discovery, where the agent decides what additional information to gather and even which new tool to construct before committing to an answer, which is the operational definition of self-evolution at the application layer.

\subsection{AutoML and Cross Stack Agents}
Maturity is finally measured not only by accuracy but by whether the construction of these systems can itself be automated and whether the medical agent can hand off cleanly to the wider clinical stack. Agent driven AutoML closes the loop on construction, with \textsc{M3Builder}~\citep{feng2025m} automating medical model building and \textsc{Evo-MedAgent}~\citep{shen2026evo} adding evolutionary self-improvement so that the pipeline that builds models is itself optimized, which is the construction time analogue of the self-evolution seen at inference. The imaging loop does not terminate at the report, since an image derived finding routinely triggers a therapeutic decision, so we note in passing that therapeutic and safety agents such as \textsc{TxAgent}~\citep{gao2025txagentaiagenttherapeutic} and \textsc{MALADE}~\citep{choi2024maladeorchestrationllmpoweredagents} reason over drug interactions and adverse events; We include them not as imaging systems but as the immediate downstream handoff of a medical agent and as evidence that the same agentic substrate, the tool grounded perceive reason act loop developed throughout this survey, transfers from pixels to pharmacology across diverse real-world clinical workflows, so that the architectural lessons of medical agents extend to the orders an imaging finding sets in motion. We deliberately restrict this therapeutic agent mention to a single paragraph precisely to keep the imaging through line intact, so that these systems enter our scope only as the order an imaging finding sets in motion rather than as objects of survey in their own right.




\section{Real World Deployment and Clinical Workflow Integration}\label{sec:deploy}

Moving an agent from a benchmark to a hospital confronts three constraints that no leaderboard captures, namely industrial productization, interoperability with existing health information systems, and regulatory clearance, and a recurring lesson from early adopters is that the binding constraint is rarely model accuracy and usually one of the other two~\citep{zou2025rise}. We therefore treat these as the three pillars of deployment and devote a subsection to each, arguing throughout that the engineering and governance questions are now as research worthy as the modeling questions that dominate the rest of this survey.

\begin{tcolorbox}[
  colback=secblue!5,
  colframe=secblue!50,
  colbacktitle=secblue!50,
  coltitle=black,
  title={\textbf{\textcolor{secblue}{Deployment:} Three Pillars}},
  boxrule=5pt,
  arc=5pt,
  drop shadow,
  parbox=false,
  before skip=5pt,
  after skip=10pt,
  left=5pt,
  right=20pt,
]
\begin{itemize}[leftmargin=*,noitemsep]
\renewcommand\labelitemi{$\diamond$}
    \item \textbf{Productization} (Section~\ref{sec:industry}): turns a research agent into a procurable product, concentrating where value is immediate and liability is bounded, such as ambient documentation and triage rather than autonomous diagnosis.
    \item \textbf{Interoperability} (Section~\ref{sec:interop}): binds the agent to the data layer the hospital already trusts (DICOM, HL7 FHIR) and to the emerging agent communication layer (MCP, A2A, ANP, ACP) so that one agent can be swapped for another.
    \item \textbf{Regulation} (Section~\ref{sec:reg}): clears the agent for a setting where approval was designed for a frozen artifact, reconciling a self-evolving model with the FDA Predetermined Change Control Plan and the EU AI Act stacked on MDR/IVDR.
\end{itemize}
\end{tcolorbox}

\subsection{Industry Deployment Landscape}
\label{sec:industry}
Commercial momentum concentrates on ambient documentation and triage, where the value is immediate and the liability is bounded, because saving minutes per encounter is measurable and reversible whereas an autonomous diagnosis is neither. Foundation efforts such as \textsc{Med-Gemini}~\citep{saab2024capabilities} establish the capability frontier on which products are built, while productized assistants integrate dictation and drafting directly into the clinician workflow, the reason being that time saved per encounter is the metric hospitals actually purchase against and the one that survives a procurement review. 

\subsection{Interoperability and Protocol Stack}
\label{sec:interop}
Deployment lives or dies on interoperability, which we separate into two layers that are often conflated. The data layer rests on long standing standards, with DICOM governing imaging and HL7 FHIR governing clinical records, so that an agent reads and writes the very representations the hospital already trusts and audits, and no integration is credible without them. The newer agent communication layer is where current research concentrates, since the Model Context Protocol exposes tools and data to agents through a uniform interface~\citep{hou2025model} and agent to agent families such as A2A, ANP and ACP standardize how independent agents negotiate and delegate work among themselves~\citep{ ehtesham2025survey}. Crucially these protocols are now being bound to clinical data rather than demonstrated in isolation, with the Model Context Protocol wired directly to FHIR for decision support~\citep{ehtesham2025enhancing} and evaluated for real-world clinical information retrieval against electronic health records~\citep{masayoshi2025ehr}, while end-to-end agentic pipelines for medical data inference~\citep{shimgekar2025agentic} show the full stack operating without a human in every loop. The motivation throughout is substitutability, because a protocol layer lets a hospital swap one agent for another without rebuilding its integration, which is the precondition for a competitive vendor ecosystem rather than lock in to a single supplier.

\subsection{Regulatory and Compliance Pathways}
\label{sec:reg}
Regulation is the pillar that most often determines timelines, and it is changing fast enough to be a research topic in its own right. In the United States the decisive instrument for adaptive systems is the Predetermined Change Control Plan, whose guiding principles let a manufacturer pre specify how a model may update after clearance~\citep{carvalho2025predetermined}, directly addressing the longstanding mismatch between a static approval and a model that is designed to keep learning. In Europe the Artificial Intelligence Act, Regulation (EU) 2024/1689~\citep{smuha2025regulation}, classifies most medical AI as high risk and stacks on top of the existing MDR and IVDR device regimes, so that a medical agent must satisfy both the device pathway and the AI specific obligations, and recent legal analysis of how autonomous agents fit this framework~\citep{nannini2026ai} highlights unresolved questions of accountability that arise when an agent acts without a human in the loop. The challenge common to both regimes is that approval was designed for a frozen artifact whereas a self-evolving agent changes after deployment, which is exactly why the Predetermined Change Control Plan mechanism is the regulatory hinge for the entire adaptation ladder from supervised fine-tuning through self-play to self-evolution (\secref{sec:train}), and why governance and self-evolution are best understood as two views of the same problem.
\section{Risks and Open Challenges}\label{sec:challenge}

\begin{wrapfigure}{r}{0.5\textwidth}
    \centering
    \vspace{-13pt}
    \includegraphics[width=0.5\textwidth]{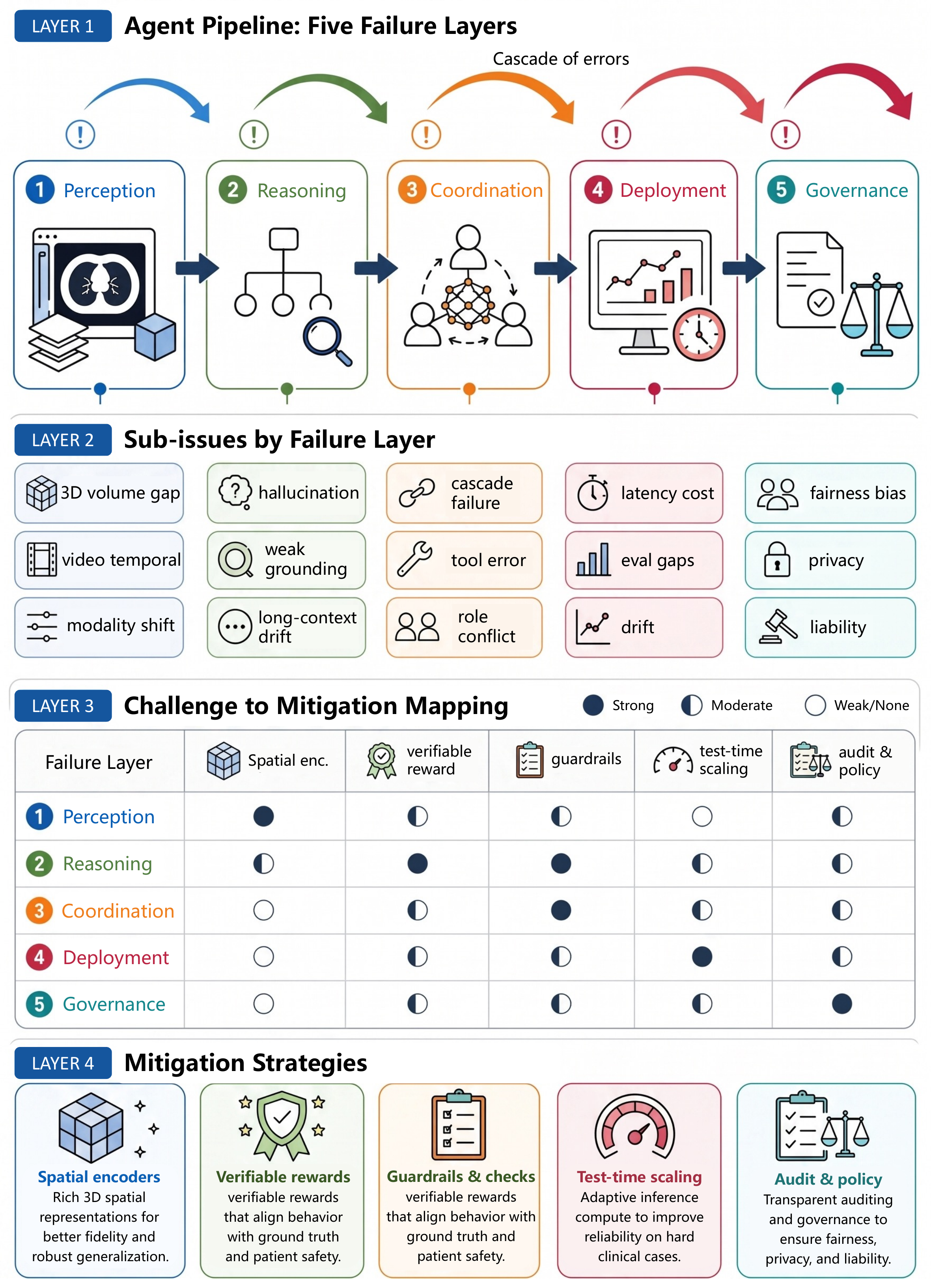}
    \caption{\textbf{Open challenges for medical agents.} Four layers: pipeline failure loci, sub challenges, mitigation coverage, and representative strategies.}
    \label{fig:challenges}
    \vspace{-25pt}
\end{wrapfigure}

The capabilities surveyed so far are real, yet none of them removes the structural risks that separate a strong benchmark number from a system a clinician can trust unsupervised. Rather than enumerate failures, we read them along the pipeline this survey has used throughout, namely the path from \textit{perception} through \textit{reasoning} and \textit{coordination} to \textit{deployment} and finally \textit{governance}, because a challenge is best understood as a property of the stage at which it originates and the stages to which it propagates~\citep{guo2024large, kim2025towards}. The organizing insight, visualized in \figref{fig:challenges}, is that autonomy multiplies the cost of every failure mode, since an error that a single forward pass would expose immediately is, in a multistep agent, propagated, rationalized and acted upon before any human sees it, so that safety becomes a property of orchestration rather than of any one model.

\subsection{Perception Layer: The 3D and Video Gap}\label{sec:3dgap}
The pipeline begins at perception, and here the binding limitation is that most agents are optimized for 2D interpretation while clinical evidence is intrinsically volumetric and temporal, so an agent that flattens a CT volume into independent slices discards exactly the spatial continuity on which a radiologist depends~\citep{hamamci2026better, mao2025ct, erdur2026agentic}. The deeper reason this is a perception layer bottleneck rather than a mere data inconvenience is tokenization, because a naive extension of 2D tokenizers to three dimensions inflates sequence length beyond any feasible inference budget, which is precisely the problem \textsc{BTB3D}~\citep{hamamci2026better} targets for medical VLMs. \textsc{CT-Agent}~\citep{mao2025ct} and the training-free Agentic Neuro-Radiology pipeline~\citep{erdur2026agentic} show that orchestration can partially compensate by reasoning over a volume through successive tool calls rather than a single ingestion, and \textsc{3MDBench}~\citep{Sviridov_2025} supplies the evaluation framework needed to distinguish genuine volumetric competence from 2D heuristics dressed up as 3D reasoning. The robustness of perception under distribution shift compounds the gap, since an agent that perceives well on one scanner may degrade silently on another, a fragility quantified by \textsc{RobustMedCLIP}~\citep{imam2025robustness} and by adversarial analyses of medical vision-language models~\citep{budathoki2025adversarial} that together establish corruption and attack as first-class perception risks rather than edge cases.

\subsection{Reasoning Layer: Hallucination}\label{sec:hallucination}

Once an agent perceives, it must reason, and the dominant failure at this layer is hallucination, the generation of plausible but factually incorrect content, which the agentic loop converts from a momentary mistake into a committed action~\citep{ji2023survey, liu2024survey, pandit2025medhallu, zuo2024medhallbench}. The danger is not occasional error but confident and verifiable error in a format indistinguishable from a correct report, so the research question is less how to eliminate hallucination than how to make every claim traceable to checkable evidence. \textbf{Visual hallucination}, in which the agent reports findings absent from the image, sits at the boundary of perception and reasoning and is best mitigated by anchoring each statement to a spatial referent so that an unsupported claim has nowhere to hide~\citep{bose2025visual, nguyen2025localizing, liu2024survey}; \textsc{VALOR}~\citep{bose2025visual} operationalizes this by binding outputs to specific image regions, while \textsc{MedHallu}~\citep{pandit2025medhallu} and \textsc{MedHallBench}~\citep{zuo2024medhallbench} provide the benchmarks without which any reduction cannot be measured. A counterintuitive result from Kim et al.~\citep{kim2025medical} is that general-purpose models surpass medically specialized ones on hallucination free generation, $76.6\%$ versus $51.3\%$, suggesting that the breadth of pretraining confers a factual grounding that narrow tuning can erode. \textbf{Reasoning hallucination}, the subtler assembly of a logically coherent yet factually unsound chain, is more dangerous because its very fluency suppresses suspicion~\citep{du2026care, pan2025medvlm}; \textsc{Med-VCD}~\citep{mahdavi2026med} attacks it at decoding time through visual contrastive decoding, and the broader portfolio spans explicit grounding~\citep{gao2026enhancing, madavan2025med}, retrieval-augmented generation that replaces memorization with verifiable recall~\citep{xia2024rule, muhetaer2025medical, karim2025multimodal}, and multiagent verification in which independent agents must agree before a conclusion is emitted~\citep{kim2024mdagents, gu2025medagentaudit, elboardy2025medical}.

\subsection{Coordination Layer: Cascade Failures}\label{sec:cascade}

When reasoning is distributed across agents, the collaboration that buys auditability introduces a failure mode absent in monolithic models, the cascade, in which one agent's error becomes another agent's trusted input and compounds along the pipeline until the final output bears no resemblance to any single agent's competence~\citep{guo2024large, silveira2025multi, kim2025towards}. The structural lesson is that resilience is a function of topology rather than of individual accuracy, which is why error detection, circuit breaker patterns and redundant paths have become first-class objects~\citep{gu2025medagentaudit, liao2025reflectool}. \textsc{MedSentry}~\citep{chen2025medsentry} makes this concrete by showing that shared pool topologies are the most vulnerable to cascades whereas decentralized architectures absorb local errors before they spread, and Ghafoor et al.~\citep{ghafoor2026improving} demonstrate that structured peer review among agents yields an $89\%$ reduction in ethical violations, evidence that the remedy for a multiagent pathology is often more, not less, multiagent structure. Trustworthiness under distribution shift is quantified by the \textsc{CARES} benchmark~\citep{xia2024carescomprehensivebenchmarktrustworthiness}, which exposes the paradoxical regime in which models simultaneously over refuse benign queries and miss subtle adversarial jailbreaks, and the adversarial surface is mapped by Ekram~\citep{ekram2026red}, whose eight category red teaming taxonomy finds that an ``Educational Authority'' framing bypasses guardrails $83.3\%$ of the time, a figure that argues for context-aware rather than keyword-based safety more forcefully than any abstract claim.

\subsection{Deployment Layer: Efficiency and Evaluation}\label{sec:compute}

Even a perceptive, sound and well coordinated agent confronts a deployment layer whose currency is latency and cost, and here agentic reasoning is expensive by construction because every additional tool call, verification step and retrieval round multiplies both, which in an emergency workflow is not a convenience concern but a deployment gate~\citep{wan2023efficient, snell2024scaling, jhandi2025small}. The frontier therefore couples model distillation and efficient inference~\citep{wan2023efficient} with small models specialized for tool calling~\citep{jhandi2025small} and with adaptive computation that spends more reasoning only on cases that warrant it~\citep{snell2024scaling, sardana2023beyond}, the unifying principle being that compute should track case difficulty rather than be spent uniformly, which is the inference time analogue of the scaling argument developed earlier. Deployment also exposes an evaluation gap, since final answer accuracy says nothing about whether the reasoning process that produced it was sound, and recent agentic benchmarks such as \textsc{MedAgentBench}~\citep{jiang2025medagentbench} and \textsc{AgentClinic}~\citep{schmidgall2024agentclinic} reframe evaluation around tool use and sequential decision making in realistic clinical environments rather than isolated question answering, making process-level assessment a prerequisite for trust rather than an afterthought.

\subsection{Governance Layer: Fairness, Privacy, and Liability}\label{sec:fairness}

The final layer is governance, where the question is no longer whether an agent works but whether society can hold it accountable, and the three concerns here are tightly coupled rather than separable. \textbf{Fairness} is at risk because an agent inherits and can amplify the biases latent in its training data, so a system accurate on average may be systematically worse for underrepresented groups, a disparity that aggregate metrics conceal by design~\citep{seyyed2021underdiagnosis, obermeyer2019dissecting, chen2021ethical, xiao2025amqa}. \textbf{Privacy} is strained because the agentic paradigm enlarges the attack surface, since every tool boundary and inter agent message is a place where sensitive data can leak~\citep{kaissis2020secure, naidu2025federated, radosevich2025mcp}; MCP security audits~\citep{radosevich2025mcp} catalog concrete risks in tool mediated interactions, \textsc{FRAME}~\citep{naidu2025federated} pursues federated learning so that data never leaves its institution, and Rempe et al.~\citep{rempe2025identification} provide production ready de-identification across DICOM metadata and pixel content. \textbf{Liability} remains unresolved because autonomy dislocates the usual chain of responsibility, and when an agent acts without a human in the loop it is genuinely unclear where accountability rests, a question current legal frameworks were never designed to answer~\citep{malpure2025securing, palaniappan2024global, pesapane2026artificial}, and one that connects this layer directly to the self-evolving training paradigm of \secref{sec:selfevolve} and the regulatory pathways discussed in \secref{sec:reg}.

\subsection{The Road Ahead}\label{sec:road}

\begin{figure}[t]
\centering
\includegraphics[width=\columnwidth]{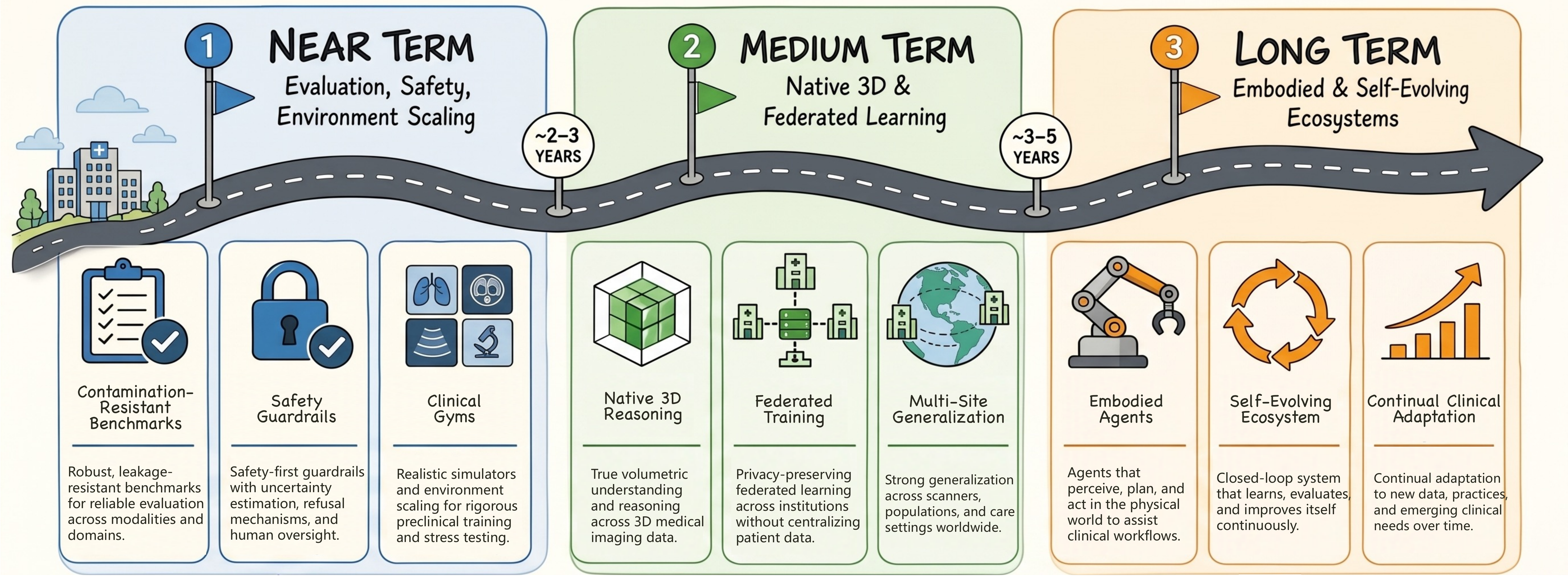}
\caption{\textbf{Future roadmap for medical agents.} Three horizons spanning near-term evaluation and safety, medium-term native 3D and federated learning, and long-term embodied and self-evolving systems.}
\label{fig:roadmap}
\end{figure}

The five layers above are not dead ends but a research agenda, and we organize that agenda by time horizon in \tabref{tab:future} and \figref{fig:roadmap} so that near-term, medium-term and long-term efforts read as one trajectory rather than a list of wishes. In the near term the priority is to make progress measurable and safe, through standardized evaluation of reasoning processes~\citep{jiang2025medagentbench}, real-time hallucination detection~\citep{pandit2025medhallu}, principled environment scaling of the tool ecosystem~\citep{fang2025towards}, and clinician agent interfaces that make collaboration effective rather than burdensome~\citep{xu2025multiagentreasoningsystemscollaborative}. The medium term shifts from patching limitations to native capability, with volumetric agents built on efficient 3D and 4D tokenization~\citep{hamamci2026better}, longitudinal modeling over long-horizon patient timelines~\citep{yu2026agenticmemorylearningunified}, federated agents that learn across sites without centralizing data~\citep{naidu2025federated}, and the regulatory maturation needed to certify multistep reasoning~\citep{malpure2025securing}. The long term envisions agents that act in the world, spanning embodied and image-guided robotic integration~\citep{haidegger2019autonomy}, population-level autonomous screening~\citep{leong2026autonomous}, multiomic precision medicine~\citep{moon2022moma}, and self-evolving ecosystems that improve continually after deployment~\citep{fan2026evolving}, closing the loop back to the self-evolution this survey has treated throughout as both the central opportunity and the central governance problem.

\begin{table*}[!t]
\centering
\caption{\textbf{Future Directions for Medical Agents by Time Horizon.}}
\label{tab:future}
\begingroup
\footnotesize
\setlength{\tabcolsep}{3pt}
\renewcommand{\arraystretch}{1.05}
\newcommand{\horizoncell}[1]{\multirow[c]{4}{=}{\centering\arraybackslash #1}}
\begin{tabularx}{\textwidth}{@{}>{\centering\arraybackslash}m{1.5cm}p{4.5cm} X p{3.8cm}@{}}
\toprule
\textbf{Horizon} & \textbf{Direction} & \textbf{Key Research Questions} & \textbf{Expected Impact} \\
\midrule
\horizoncell{Near term} & Standardized evaluation~\citep{jiang2025medagentbench} & How to evaluate agent reasoning processes? & Reproducible comparison \\
 & Hallucination detection~\citep{pandit2025medhallu} & Real time grounding verification? & Clinical safety assurance \\
 & Environment scaling~\citep{fang2025towards} & Optimal tool ecosystem composition? & Amplified agent capability \\
 & Human agent interaction~\citep{xu2025multiagentreasoningsystemscollaborative} & Optimal clinician agent interfaces? & Effective collaboration \\
\midrule
\horizoncell{Medium} & Native 3D/4D agents~\citep{hamamci2026better} & Efficient volumetric tokenization? & Comprehensive interpretation \\
 & Longitudinal modeling~\citep{yu2026agenticmemorylearningunified} & Long horizon patient timelines? & Progression tracking \\
 & Federated agents~\citep{naidu2025federated} & Privacy preserving multisite learning? & Equitable agents \\
 & Regulatory maturation~\citep{malpure2025securing} & How to certify multistep reasoning? & Legal deployment \\
\midrule
\horizoncell{Long term} & Embodied agents~\citep{haidegger2019autonomy} & Image guided robotic integration? & Autonomous procedures \\
 & Autonomous screening~\citep{leong2026autonomous} & Population level deployment? & Universal access \\
 & Precision medicine~\citep{moon2022moma} & Multi omic integration? & Personalized treatment \\
 & self-evolving ecosystems~\citep{fan2026evolving} & Continual self-improvement? & Adaptive AI \\
\bottomrule
\end{tabularx}
\endgroup
\end{table*}

\section{Future Directions}\label{sec:future}

Whereas \secref{sec:road} reads the open problems as a research agenda organized by time horizon (\figref{fig:roadmap}, \tabref{tab:future}), this section takes the complementary view and projects each horizon back onto the scaling spine of \secref{sec:spine}, asking not only what remains to be solved but along which axis, framework, capability, or environment scaling, the next increment of readiness will be purchased. Read this way, the trajectory of the field is not a list of independent wishes but the continued ascent of a single agent along the adaptation ladder of \secref{sec:train}, whose destination is the self-evolving clinical system analyzed in \secref{sec:selfevolve}.

\subsection{Near Term Directions}\label{sec:nearterm}

In the near term the binding constraint is not capability but trust, and progress therefore concentrates on making the loop measurable and safe before widening it. Because agentic competence lives in the runtime loop rather than in a single forward pass, evaluation must migrate from final answer accuracy to the reasoning process itself, scoring tool orchestration, multistep inference, and error recovery as first-class quantities; early frameworks such as \textsc{MedAgentBench}~\citep{jiang2025medagentbench} establish the format, but community wide standardization of process-level metrics remains open. Coupled to this is real-time hallucination detection deployed as a safety layer inside the loop rather than as a post hoc filter, binding every generated finding to a spatial referent or a retrieved fact so that an unsupported claim is caught at the step that produced it. On the environment axis, the most immediately actionable lever is principled environment scaling: expanding the reachable tool set $\mathcal{E}$ through standardized MCP profiles and audited tool registries buys capability without retraining, and initial tool environment design work~\citep{song2026envscaler, fang2025towards} indicates this is feasible within existing hospital infrastructure. Finally, the interface between clinician and agent, how uncertainty is surfaced, how clarification is requested, and how feedback re enters the loop, will determine whether cooperative (Level 2) systems are adopted at all~\citep{xu2025multiagentreasoningsystemscollaborative}.

\subsection{Medium Term Directions}\label{sec:medterm}

The medium term shifts effort from patching the loop to deepening the two remaining axes, capability and framework scaling, so that agents acquire competences no orchestration of frozen 2D experts can supply. The clearest capability gap is perception: most agents still flatten a CT volume into independent slices, discarding the spatial continuity a radiologist depends on, so native 3D and 4D architectures built on efficient volumetric tokenization~\citep{hamamci2026better} are the prerequisite for genuine volumetric reasoning rather than 2D heuristics dressed up as 3D. Deepening the memory station of the loop yields longitudinal patient modelling, where an agent maintains a persistent representation across imaging time points and reasons over disease trajectories rather than isolated studies~\citep{yu2026agenticmemorylearningunified}, which is precisely the scaffold level evolution of \secref{sec:selfevolve:scaffold} applied to the temporal axis. On the framework axis, multi institutional federated agents that share experience and tool configurations without centralizing patient data~\citep{naidu2025federated} extend collaboration beyond a single hospital, and their maturation is inseparable from the regulatory question of how a multistep, tool invoking, and increasingly adaptive system can be certified when approval was designed for a frozen artifact~\citep{malpure2025securing}, the tension made explicit in \secref{sec:reg}.

\subsection{Long Term Directions}\label{sec:longterm}

The long term envisions agents that no longer merely interpret images but act on them, closing the loop between perception and intervention. Embodied and image-guided systems would fuse continuous intraoperative perception with safety critical control, adjusting a surgical trajectory in real time as the agent reinterprets the scene~\citep{haidegger2019autonomy}, the point at which the perceive reason act loop of \secref{sec:core} leaves the reading room. Fully autonomous (Level 3) screening pipelines that triage, diagnose, and route at population scale could extend expert level interpretation into resource limited settings where no radiologist is available~\citep{leong2026autonomous}, provided reliability and regulation mature in step. Integration with genomic, proteomic, and longitudinal clinical data would push the same agentic substrate from pixels toward multiomic precision medicine, grounding recommendations in individual patient biology rather than population statistics. Underlying all three is the destination this survey has treated throughout as both the central opportunity and the central governance problem, self-evolving agent ecosystems~\citep{fan2026evolving, shen2026evo} that consolidate accumulated clinical experience into institutional knowledge and adapt continually to new modalities, protocols, and guidelines without explicit retraining, so that the boundary between training and deployment dissolves under the verifiable safeguards of \secref{sec:selfevolve:risks}.

\section{Conclusion}\label{sec:conclusion}

This survey has provided a comprehensive examination of medical agents, a rapidly emerging paradigm that extends artificial intelligence in medical imaging from passive, task-specific models to autonomous systems capable of perception, reasoning, planning, memory, tool use, and self-reflection. We formalized medical agents as a sequential decision process under partial observability realized through six interacting cognitive modules, making explicit that medical agency is fundamentally a matter of acting under partial observability, and proposed a three-level taxonomy that offers a unified lens on the diverse landscape of existing systems.

Our analysis reveals that the agent paradigm addresses fundamental limitations of prior approaches by enabling multistep clinical workflows, dynamic tool orchestration, contextual memory, and iterative self-improvement. We argued that capability scales less with raw model size than with the runtime loop that drives the six core cognitive modules, organized the architecture space into three paradigms ranging from single agent tool augmentation to orchestrated topologies, and recast adaptation as a single supervision signal ladder, from frozen prompting through demonstration, preference, and verifiable reward to self-generated experience. Cutting orthogonally across this ladder, the three scaling axes of parameter, test-time, and environment scaling clarify where additional capability is purchased, and we singled out \emph{environment scaling}, the enrichment of the tool, data, and interface environment, as the most immediately actionable lever for agents that already inhabit tool rich clinical ecosystems.

The clinical applications surveyed across radiology, pathology, ophthalmology, oncology, neurology, and cross-modal settings demonstrate both the promise and current limitations of medical agents. Our review of real-world deployment further shows that successful translation depends not only on model accuracy but also on productization, interoperability through DICOM, HL7 FHIR, and emerging agent communication protocols, together with evolving regulatory frameworks. Viewed along the agent pipeline, key challenges remain in three dimensional and video perception, hallucination resistant reasoning, robust multi-agent coordination, efficient deployment and trustworthy evaluation, and governance issues spanning fairness, privacy, and liability.

Looking forward, the convergence of increasingly capable multimodal foundation models, standardized tool integration, and environment scaling, together with growing clinical demand for intelligent workflow automation, provides a credible path toward broader clinical adoption of medical agents. Although self-evolving medical agents remain an aspirational frontier rather than a solved capability, continued collaboration among AI researchers, clinicians, regulators, and healthcare administrators will be essential to ensure these systems are safe, equitable, and clinically aligned. The foundations surveyed in this review suggest that the coming decade will substantially reshape how medical images are interpreted, communicated, and acted upon, improving the quality, efficiency, accessibility, and timeliness of diagnostic imaging across diverse healthcare settings.

\bibliographystyle{plainnat}
\bibliography{references}

\begin{thebibliography}{344}
\providecommand{\natexlab}[1]{#1}
\providecommand{\url}[1]{\texttt{#1}}
\expandafter\ifx\csname urlstyle\endcsname\relax
  \providecommand{\doi}[1]{doi: #1}\else
  \providecommand{\doi}{doi: \begingroup \urlstyle{rm}\Url}\fi

\bibitem[Achiam et~al.(2023)Achiam, Adler, Agarwal, Ahmad, Akkaya, Aleman, Almeida, Altenschmidt, Altman, Anadkat, et~al.]{achiam2023gpt}
Josh Achiam, Steven Adler, Sandhini Agarwal, Lama Ahmad, Ilge Akkaya, Florencia~Leoni Aleman, Diogo Almeida, Janko Altenschmidt, Sam Altman, Shyamal Anadkat, et~al.
\newblock Gpt-4 technical report.
\newblock \emph{arXiv preprint arXiv:2303.08774}, 2023.

\bibitem[Acikgoz et~al.(2025)Acikgoz, Qian, Ji, Hakkani-T{\"u}r, and Tur]{acikgoz2025self}
Emre~Can Acikgoz, Cheng Qian, Heng Ji, Dilek Hakkani-T{\"u}r, and Gokhan Tur.
\newblock Self-improving llm agents at test-time.
\newblock \emph{arXiv preprint arXiv:2510.07841}, 2025.

\bibitem[Adams et~al.(2025)Adams, Busch, Han, Excoffier, Ortala, L{\"o}ser, Aerts, Kather, Truhn, and Bressem]{adams2025longhealth}
Lisa Adams, Felix Busch, Tianyu Han, Jean-Baptiste Excoffier, Matthieu Ortala, Alexander L{\"o}ser, Hugo~JWL Aerts, Jakob~Nikolas Kather, Daniel Truhn, and Keno Bressem.
\newblock Longhealth: A question answering benchmark with long clinical documents.
\newblock \emph{Journal of Healthcare Informatics Research}, 9\penalty0 (3):\penalty0 280--296, 2025.

\bibitem[Aghaee et~al.(2026)Aghaee, Asgarian, and Jeon]{aghaee2026synthagent}
Arman Aghaee, Sepehr Asgarian, and Jouhyun Jeon.
\newblock Synthagent: A multi-agent llm framework for realistic patient simulation--a case study in obesity with mental health comorbidities.
\newblock \emph{arXiv preprint arXiv:2602.08254}, 2026.

\bibitem[Ahadian et~al.(2025)Ahadian, Yang, Powlison, Li, Xu, and Guan]{ahadian2025oaagent}
Pegah Ahadian, Mingrui Yang, Eva Powlison, Xiaojuan Li, Wei Xu, and Qiang Guan.
\newblock Oaagent: Multimodal llm agent for predicting knee osteoarthritis progression.
\newblock In \emph{Proceedings of the ACM/IEEE International Conference on Connected Health: Applications, Systems and Engineering Technologies}, pages 144--148, 2025.

\bibitem[Almansoori et~al.(2025)Almansoori, Kumar, and Cholakkal]{almansoori2025self}
Mohammad Almansoori, Komal Kumar, and Hisham Cholakkal.
\newblock Self-evolving multi-agent simulations for realistic clinical interactions.
\newblock \emph{arXiv preprint arXiv:2503.22678}, 2025.

\bibitem[{Anthropic}(2025{\natexlab{a}})]{anthropic2025contextengineering}
{Anthropic}.
\newblock Effective context engineering for {AI} agents.
\newblock Anthropic Engineering Blog, 2025{\natexlab{a}}.
\newblock \url{https://www.anthropic.com/engineering/effective-context-engineering-for-ai-agents}, accessed 2026-06-14.

\bibitem[{Anthropic}(2025{\natexlab{b}})]{anthropic2025harnesses}
{Anthropic}.
\newblock Effective harnesses for long-running agents.
\newblock Anthropic Engineering Blog, 2025{\natexlab{b}}.
\newblock \url{https://www.anthropic.com/engineering/effective-harnesses-for-long-running-agents}, accessed 2026-06-14.

\bibitem[Arora et~al.(2025)Arora, Wei, Hicks, Bowman, Qui{\~n}onero-Candela, Tsimpourlas, Sharman, Shah, Vallone, Beutel, et~al.]{arora2025healthbench}
Rahul~K Arora, Jason Wei, Rebecca~Soskin Hicks, Preston Bowman, Joaquin Qui{\~n}onero-Candela, Foivos Tsimpourlas, Michael Sharman, Meghan Shah, Andrea Vallone, Alex Beutel, et~al.
\newblock Healthbench: Evaluating large language models towards improved human health.
\newblock \emph{arXiv preprint arXiv:2505.08775}, 2025.

\bibitem[Bai et~al.(2025)Bai, Cai, Chen, Chen, Chen, Cheng, Deng, Ding, Gao, Ge, et~al.]{bai2025qwen3}
Shuai Bai, Yuxuan Cai, Ruizhe Chen, Keqin Chen, Xionghui Chen, Zesen Cheng, Lianghao Deng, Wei Ding, Chang Gao, Chunjiang Ge, et~al.
\newblock Qwen3-vl technical report.
\newblock \emph{arXiv preprint arXiv:2511.21631}, 2025.

\bibitem[Bai et~al.(2022)Bai, Kadavath, Kundu, Askell, Kernion, Jones, Chen, Goldie, Mirhoseini, McKinnon, et~al.]{bai2022constitutional}
Yuntao Bai, Saurav Kadavath, Sandipan Kundu, Amanda Askell, Jackson Kernion, Andy Jones, Anna Chen, Anna Goldie, Azalia Mirhoseini, Cameron McKinnon, et~al.
\newblock Constitutional ai: Harmlessness from ai feedback.
\newblock \emph{arXiv preprint arXiv:2212.08073}, 2022.

\bibitem[Banerjie et~al.(2025)Banerjie, Zhu, Freeman, Machado, Ahmed, Sarker, and Al-Garadi]{banerjie2025agentic}
Shruti Banerjie, Yuxin Zhu, Isaac Freeman, Julyssa~Villa Machado, Abdulaziz Ahmed, Abeed Sarker, and Mohammed Al-Garadi.
\newblock Agentic ai in healthcare: A comprehensive survey of foundations, taxonomy, and applications.
\newblock \emph{Authorea Preprints}, 2025.

\bibitem[Bani-Harouni et~al.(2024)Bani-Harouni, Navab, and Keicher]{bani2024magda}
David Bani-Harouni, Nassir Navab, and Matthias Keicher.
\newblock Magda: Multi-agent guideline-driven diagnostic assistance.
\newblock In \emph{International workshop on foundation models for general medical AI}, pages 163--172. Springer, 2024.

\bibitem[Bannur et~al.(2024)Bannur, Bouzid, Castro, Schwaighofer, Thieme, Bond-Taylor, Ilse, P{\'e}rez-Garc{\'\i}a, Salvatelli, Sharma, et~al.]{bannur2024maira}
Shruthi Bannur, Kenza Bouzid, Daniel~C Castro, Anton Schwaighofer, Anja Thieme, Sam Bond-Taylor, Maximilian Ilse, Fernando P{\'e}rez-Garc{\'\i}a, Valentina Salvatelli, Harshita Sharma, et~al.
\newblock Maira-2: Grounded radiology report generation.
\newblock \emph{arXiv preprint arXiv:2406.04449}, 2024.

\bibitem[Bassi et~al.(2025)Bassi, Yavuz, Hamamci, Er, Chen, Li, Menze, Decherchi, Cavalli, Wang, et~al.]{bassi2025radgpt}
Pedro~RAS Bassi, Mehmet~Can Yavuz, Ibrahim~Ethem Hamamci, Sezgin Er, Xiaoxi Chen, Wenxuan Li, Bjoern Menze, Sergio Decherchi, Andrea Cavalli, Kang Wang, et~al.
\newblock Radgpt: Constructing 3d image-text tumor datasets.
\newblock In \emph{Proceedings of the IEEE/CVF international conference on computer vision}, pages 23720--23730, 2025.

\bibitem[Bedi et~al.(2025)Bedi, Cui, Fuentes, Unell, Wornow, Banda, Kotecha, Keyes, Mai, Oez, et~al.]{bedi2025medhelm}
Suhana Bedi, Hejie Cui, Miguel Fuentes, Alyssa Unell, Michael Wornow, Juan~M Banda, Nikesh Kotecha, Timothy Keyes, Yifan Mai, Mert Oez, et~al.
\newblock Medhelm: Holistic evaluation of large language models for medical tasks.
\newblock \emph{arXiv preprint arXiv:2505.23802}, 2025.

\bibitem[Bedi et~al.(2026)Bedi, Welch, Steinberg, Wornow, Kim, Ahmed, Sterling, Purohit, Akram, Acosta, et~al.]{bedi2026healthadminbench}
Suhana Bedi, Ryan Welch, Ethan Steinberg, Michael Wornow, Taeil~Matthew Kim, Haroun Ahmed, Peter Sterling, Bravim Purohit, Qurat Akram, Angelic Acosta, et~al.
\newblock Healthadminbench: Evaluating computer-use agents on healthcare administration tasks.
\newblock \emph{arXiv preprint arXiv:2604.09937}, 2026.

\bibitem[Besta et~al.(2024)Besta, Blach, Kubicek, Gerstenberger, Podstawski, Gianinazzi, Gajda, Lehmann, Niewiadomski, Nyczyk, et~al.]{besta2024graph}
Maciej Besta, Nils Blach, Ales Kubicek, Robert Gerstenberger, Michal Podstawski, Lukas Gianinazzi, Joanna Gajda, Tomasz Lehmann, Hubert Niewiadomski, Piotr Nyczyk, et~al.
\newblock Graph of thoughts: Solving elaborate problems with large language models.
\newblock In \emph{Proceedings of the AAAI conference on artificial intelligence}, volume~38, pages 17682--17690, 2024.

\bibitem[Bluethgen et~al.(2025)Bluethgen, Van~Veen, Truhn, Kather, Moor, Polacin, Chaudhari, Frauenfelder, Langlotz, Krauthammer, et~al.]{bluethgen2025agentic}
Christian Bluethgen, Dave Van~Veen, Daniel Truhn, Jakob~Nikolas Kather, Michael Moor, Malgorzata Polacin, Akshay Chaudhari, Thomas Frauenfelder, Curtis~P Langlotz, Michael Krauthammer, et~al.
\newblock Agentic systems in radiology: Design, applications, evaluation, and challenges.
\newblock \emph{arXiv preprint arXiv:2510.09404}, 2025.

\bibitem[Boecking et~al.(2022)Boecking, Usuyama, Bannur, Castro, Schwaighofer, Hyland, Wetscherek, Naumann, Nori, Alvarez-Valle, et~al.]{boecking2022making}
Benedikt Boecking, Naoto Usuyama, Shruthi Bannur, Daniel~C Castro, Anton Schwaighofer, Stephanie Hyland, Maria Wetscherek, Tristan Naumann, Aditya Nori, Javier Alvarez-Valle, et~al.
\newblock Making the most of text semantics to improve biomedical vision--language processing.
\newblock In \emph{European conference on computer vision}, pages 1--21. Springer, 2022.

\bibitem[Bose et~al.(2025)Bose, Rajendran, Debnath, Karydis, Roy-Chowdhury, and Chakradhar]{bose2025visual}
Sarosij Bose, Ravi~K Rajendran, Biplob Debnath, Konstantinos Karydis, Amit~K Roy-Chowdhury, and Srimat Chakradhar.
\newblock Visual alignment of medical vision-language models for grounded radiology report generation.
\newblock \emph{arXiv preprint arXiv:2512.16201}, 2025.

\bibitem[Budathoki and Dhakal(2025)]{budathoki2025adversarial}
Anjila Budathoki and Manish Dhakal.
\newblock Adversarial robustness analysis of vision-language models in medical image segmentation.
\newblock \emph{arXiv preprint arXiv:2505.02971}, 2025.

\bibitem[Cai et~al.(2025)Cai, Wang, Liu, Liu, Niu, and Sugiyama]{cai2025reinforcement}
Xin-Qiang Cai, Wei Wang, Feng Liu, Tongliang Liu, Gang Niu, and Masashi Sugiyama.
\newblock Reinforcement learning with verifiable yet noisy rewards under imperfect verifiers.
\newblock \emph{arXiv preprint arXiv:2510.00915}, 2025.

\bibitem[Cao et~al.(2026)Cao, Bai, Yao, Dong, Xu, and Xiao]{cao2026atpo}
Ruike Cao, Shaojie Bai, Fugen Yao, Liang Dong, Jian Xu, and Li~Xiao.
\newblock Atpo: Adaptive tree policy optimization for multi-turn medical dialogue.
\newblock \emph{arXiv preprint arXiv:2603.02216}, 2026.

\bibitem[Carvalho et~al.(2025)Carvalho, Mascarenhas, Pinheiro, Correia, Balseiro, Barbosa, Guerra, Oliveira, Moura, Martins~dos Santos, et~al.]{carvalho2025predetermined}
Eduardo Carvalho, Miguel Mascarenhas, Francisca Pinheiro, Ricardo Correia, Sandra Balseiro, Guilherme Barbosa, Ana Guerra, Dulce Oliveira, Rita Moura, Andr{\'e} Martins~dos Santos, et~al.
\newblock Predetermined change control plans: Guiding principles for advancing safe, effective, and high-quality ai-ml technologies.
\newblock \emph{JMIR AI}, 4:\penalty0 e76854, 2025.

\bibitem[Chaves et~al.(2024)Chaves, Huang, Xu, Xu, Usuyama, Zhang, Wang, Xie, Khademi, Yang, et~al.]{chaves2024towards}
Juan Manuel~Zambrano Chaves, Shih-Cheng Huang, Yanbo Xu, Hanwen Xu, Naoto Usuyama, Sheng Zhang, Fei Wang, Yujia Xie, Mahmoud Khademi, Ziyi Yang, et~al.
\newblock Towards a clinically accessible radiology foundation model: open-access and lightweight, with automated evaluation.
\newblock \emph{arXiv preprint arXiv:2403.08002}, 2024.

\bibitem[Chen et~al.(2025{\natexlab{a}})Chen, Li, Gong, Jiang, Fei, Yang, Shan, Yu, Wang, Zhu, et~al.]{chen2025minimax}
Aili Chen, Aonian Li, Bangwei Gong, Binyang Jiang, Bo~Fei, Bo~Yang, Boji Shan, Changqing Yu, Chao Wang, Cheng Zhu, et~al.
\newblock Minimax-m1: Scaling test-time compute efficiently with lightning attention.
\newblock \emph{arXiv preprint arXiv:2506.13585}, 2025{\natexlab{a}}.

\bibitem[Chen et~al.(2024{\natexlab{a}})Chen, Yu, Chen, Liu, Wan, Bitterman, Wang, and Shu]{chen2024clinicalbench}
Canyu Chen, Jian Yu, Shan Chen, Che Liu, Zhongwei Wan, Danielle Bitterman, Fei Wang, and Kai Shu.
\newblock Clinicalbench: Can llms beat traditional ml models in clinical prediction?
\newblock \emph{arXiv preprint arXiv:2411.06469}, 2024{\natexlab{a}}.

\bibitem[Chen et~al.(2026{\natexlab{a}})Chen, Metelski, Qi, Xia, Lee, Brown, Riley, Wang, Liu, MD, et~al.]{chen2026chi}
Haolin Chen, Deon Metelski, Leon Qi, Tao Xia, Joonyul Lee, Steve Brown, Kevin Riley, Frank Wang, TY~Liu, Hank~Capps MD, et~al.
\newblock Chi-bench: Can ai agents automate end-to-end, long-horizon, policy-rich healthcare workflows?
\newblock \emph{arXiv preprint arXiv:2605.16679}, 2026{\natexlab{a}}.

\bibitem[Chen et~al.(2021)Chen, Pierson, Rose, Joshi, Ferryman, and Ghassemi]{chen2021ethical}
Irene~Y Chen, Emma Pierson, Sherri Rose, Shalmali Joshi, Kadija Ferryman, and Marzyeh Ghassemi.
\newblock Ethical machine learning in healthcare.
\newblock \emph{Annual review of biomedical data science}, 4\penalty0 (1):\penalty0 123--144, 2021.

\bibitem[Chen et~al.(2025{\natexlab{b}})Chen, Cai, Wang, Huang, Jiang, Huang, Wang, and Zhang]{chen2025pathagent}
Jingyun Chen, Linghan Cai, Zhikang Wang, Yi~Huang, Songhan Jiang, Shenjin Huang, Hongpeng Wang, and Yongbing Zhang.
\newblock Pathagent: Toward interpretable analysis of whole-slide pathology images via large language model-based agentic reasoning.
\newblock \emph{arXiv preprint arXiv:2511.17052}, 2025{\natexlab{b}}.

\bibitem[Chen et~al.(2025{\natexlab{c}})Chen, Li, Yang, Wang, Dong, and Gao]{chen2025mdteamgpt}
Kai Chen, Xinfeng Li, Tianpei Yang, Hewei Wang, Wei Dong, and Yang Gao.
\newblock Mdteamgpt: A self-evolving llm-based multi-agent framework for multi-disciplinary team medical consultation.
\newblock \emph{arXiv preprint arXiv:2503.13856}, 2025{\natexlab{c}}.

\bibitem[Chen et~al.(2025{\natexlab{d}})Chen, Zhen, Wang, Liu, Li, Huo, Yang, Xu, Dong, and Gao]{chen2025medsentry}
Kai Chen, Taihang Zhen, Hewei Wang, Kailai Liu, Xinfeng Li, Jing Huo, Tianpei Yang, Jinfeng Xu, Wei Dong, and Yang Gao.
\newblock Medsentry: Understanding and mitigating safety risks in medical llm multi-agent systems.
\newblock \emph{arXiv preprint arXiv:2505.20824}, 2025{\natexlab{d}}.

\bibitem[Chen et~al.(2024{\natexlab{b}})Chen, Ding, Lu, Williamson, Jaume, Song, Chen, Zhang, Shao, Shaban, et~al.]{chen2024towards}
Richard~J Chen, Tong Ding, Ming~Y Lu, Drew~FK Williamson, Guillaume Jaume, Andrew~H Song, Bowen Chen, Andrew Zhang, Daniel Shao, Muhammad Shaban, et~al.
\newblock Towards a general-purpose foundation model for computational pathology.
\newblock \emph{Nature medicine}, 30\penalty0 (3):\penalty0 850--862, 2024{\natexlab{b}}.

\bibitem[Chen et~al.(2025{\natexlab{e}})Chen, Moreira, Xiao, Schmidgall, Warner, Aerts, Hartvigsen, Gallifant, and Bitterman]{chen2025medbrowsecomp}
Shan Chen, Pedro Moreira, Yuxin Xiao, Sam Schmidgall, Jeremy Warner, Hugo Aerts, Thomas Hartvigsen, Jack Gallifant, and Danielle~S Bitterman.
\newblock Medbrowsecomp: Benchmarking medical deep research and computer use.
\newblock \emph{arXiv preprint arXiv:2505.14963}, 2025{\natexlab{e}}.

\bibitem[Chen et~al.(2025{\natexlab{f}})Chen, Dong, Ding, Shi, Zhou, Zeng, Luo, Lin, Su, Wu, et~al.]{chen2025radfabric}
Wenting Chen, Yi~Dong, Zhaojun Ding, Yucheng Shi, Yifan Zhou, Fang Zeng, Yijun Luo, Tianyu Lin, Yihang Su, Yichen Wu, et~al.
\newblock Radfabric: Agentic ai system with reasoning capability for radiology.
\newblock \emph{arXiv preprint arXiv:2506.14142}, 2025{\natexlab{f}}.

\bibitem[Chen et~al.(2026{\natexlab{b}})Chen, Maiga, Rahmani, and Yilmaz]{chen2026automated}
Yinzhu Chen, Abdine Maiga, Hossein~A Rahmani, and Emine Yilmaz.
\newblock Automated rubrics for reliable evaluation of medical dialogue systems.
\newblock \emph{arXiv preprint arXiv:2601.15161}, 2026{\natexlab{b}}.

\bibitem[Chen et~al.(2026{\natexlab{c}})Chen, Bai, Pan, Zhou, and Yuille]{chen2026meissa}
Yixiong Chen, Xinyi Bai, Yue Pan, Zongwei Zhou, and Alan Yuille.
\newblock Meissa: Multi-modal medical agentic intelligence.
\newblock \emph{arXiv preprint arXiv:2603.09018}, 2026{\natexlab{c}}.

\bibitem[Chen et~al.(2024{\natexlab{c}})Chen, Wu, Wang, Su, Chen, Xing, Zhong, Zhang, Zhu, Lu, et~al.]{chen2024internvl}
Zhe Chen, Jiannan Wu, Wenhai Wang, Weijie Su, Guo Chen, Sen Xing, Muyan Zhong, Qinglong Zhang, Xizhou Zhu, Lewei Lu, et~al.
\newblock Internvl: Scaling up vision foundation models and aligning for generic visual-linguistic tasks.
\newblock In \emph{Proceedings of the IEEE/CVF conference on computer vision and pattern recognition}, pages 24185--24198, 2024{\natexlab{c}}.

\bibitem[Chen et~al.(2024{\natexlab{d}})Chen, Varma, Delbrouck, Paschali, Blankemeier, Van~Veen, Valanarasu, Youssef, Cohen, Reis, et~al.]{chen2024chexagent}
Zhihong Chen, Maya Varma, Jean-Benoit Delbrouck, Magdalini Paschali, Louis Blankemeier, Dave Van~Veen, Jeya Maria~Jose Valanarasu, Alaa Youssef, Joseph~Paul Cohen, Eduardo~Pontes Reis, et~al.
\newblock Chexagent: Towards a foundation model for chest x-ray interpretation.
\newblock In \emph{AAAI 2024 Spring Symposium on Clinical Foundation Models}, 2024{\natexlab{d}}.

\bibitem[Chhikara et~al.(2025)Chhikara, Khant, Aryan, Singh, and Yadav]{chhikara2025mem0}
Prateek Chhikara, Dev Khant, Saket Aryan, Taranjeet Singh, and Deshraj Yadav.
\newblock Mem0: Building production-ready ai agents with scalable long-term memory.
\newblock \emph{arXiv preprint arXiv:2504.19413}, 2025.

\bibitem[Choi et~al.(2024)Choi, Palumbo, Chalasani, Engelhard, Jha, Kumar, and Page]{choi2024maladeorchestrationllmpoweredagents}
Jihye Choi, Nils Palumbo, Prasad Chalasani, Matthew~M. Engelhard, Somesh Jha, Anivarya Kumar, and David Page.
\newblock Malade: Orchestration of llm-powered agents with retrieval augmented generation for pharmacovigilance, 2024.
\newblock URL \url{https://arxiv.org/abs/2408.01869}.

\bibitem[Cofala et~al.(2025)Cofala, Kalfar, Xiao, Schrader, Tang, and Nejdl]{cofala2025medai}
Tim Cofala, Christian Kalfar, Jingge Xiao, Johanna Schrader, Michelle Tang, and Wolfgang Nejdl.
\newblock Medai: Evaluating txagent's therapeutic agentic reasoning in the neurips cure-bench competition.
\newblock \emph{arXiv preprint arXiv:2512.11682}, 2025.

\bibitem[Committee(2021)]{on2021taxonomy}
On-Road Automated Driving~(ORAD) Committee.
\newblock \emph{Taxonomy and definitions for terms related to driving automation systems for on-road motor vehicles}.
\newblock SAE international, 2021.

\bibitem[Da et~al.(2025)Da, Wang, Deng, Ma, Barhate, and Hendryx]{da2025agent}
Jeff Da, Clinton Wang, Xiang Deng, Yuntao Ma, Nikhil Barhate, and Sean Hendryx.
\newblock Agent-rlvr: Training software engineering agents via guidance and environment rewards.
\newblock \emph{arXiv preprint arXiv:2506.11425}, 2025.

\bibitem[DeLucia et~al.(2026)DeLucia, Huang, Joshi, Yarmohammadi, Hassoon, and Dredze]{delucia2026same}
Alexandra DeLucia, Heyuan Huang, Sonal Joshi, Mahsa Yarmohammadi, Ahmed Hassoon, and Mark Dredze.
\newblock Same verdict, different reasons: Llm-as-a-judge and clinician disagreement on medical chatbot completeness.
\newblock \emph{arXiv preprint arXiv:2604.16383}, 2026.

\bibitem[Deng et~al.(2025)Deng, Zhao, Miao, Zhu, and Li]{deng2025medka}
Yiyan Deng, Shen Zhao, Yongming Miao, Junjie Zhu, and Jin Li.
\newblock Medka: A knowledge graph-augmented approach to improve factuality in medical large language models.
\newblock \emph{Journal of Biomedical Informatics}, page 104871, 2025.

\bibitem[Deperrois et~al.(2025)Deperrois, Matsuo, Ruip{\'e}rez-Campillo, Vandenhirtz, Laguna, Ryser, Fujimoto, Nishio, Sutter, Vogt, et~al.]{deperrois2025radvlm}
Nicolas Deperrois, Hidetoshi Matsuo, Samuel Ruip{\'e}rez-Campillo, Moritz Vandenhirtz, Sonia Laguna, Alain Ryser, Koji Fujimoto, Mizuho Nishio, Thomas~M Sutter, Julia~E Vogt, et~al.
\newblock Radvlm: a multitask conversational vision-language model for radiology.
\newblock \emph{arXiv preprint arXiv:2502.03333}, 2025.

\bibitem[Dietrich(2025)]{dietrich2025agentic}
Nicholas Dietrich.
\newblock Agentic ai in radiology: emerging potential and unresolved challenges.
\newblock \emph{British Journal of Radiology}, 98\penalty0 (1174):\penalty0 1582--1584, 2025.

\bibitem[Ding et~al.(2025)Ding, Le, Han, Ruan, Jin, Kumar, Wang, and Deoras]{ding2025empowering}
Yifeng Ding, Hung Le, Songyang Han, Kangrui Ruan, Zhenghui Jin, Varun Kumar, Zijian Wang, and Anoop Deoras.
\newblock Empowering multi-turn tool-integrated agentic reasoning with group turn policy optimization.
\newblock \emph{arXiv preprint arXiv:2511.14846}, 2025.

\bibitem[Dong et~al.(2025)Dong, Bao, Wang, Zhao, Li, Jin, Yang, Mao, Zhang, Gai, et~al.]{dong2025agentic}
Guanting Dong, Licheng Bao, Zhongyuan Wang, Kangzhi Zhao, Xiaoxi Li, Jiajie Jin, Jinghan Yang, Hangyu Mao, Fuzheng Zhang, Kun Gai, et~al.
\newblock Agentic entropy-balanced policy optimization.
\newblock \emph{arXiv preprint arXiv:2510.14545}, 2025.

\bibitem[Dong et~al.(2026{\natexlab{a}})Dong, Lu, Huang, Zhong, Liu, Huang, Li, Zhao, Song, Li, et~al.]{dong2026agent}
Guanting Dong, Junting Lu, Junjie Huang, Wanjun Zhong, Longxiang Liu, Shijue Huang, Zhenyu Li, Yang Zhao, Xiaoshuai Song, Xiaoxi Li, et~al.
\newblock Agent-world: Scaling real-world environment synthesis for evolving general agent intelligence.
\newblock \emph{arXiv preprint arXiv:2604.18292}, 2026{\natexlab{a}}.

\bibitem[Dong et~al.(2026{\natexlab{b}})Dong, Zhu, Chen, Wang, Li, Xiong, Cheng, Wang, Yu, Wu, et~al.]{dong2026ophin}
Xuanzhao Dong, Wenhui Zhu, Xiwen Chen, Hao Wang, Xin Li, Yujian Xiong, Jiajun Cheng, Jingjing Wang, Xiaobing Yu, Haiyu Wu, et~al.
\newblock Ophin-500k: Curating web-scale visual instructions for scaling ophthalmic multimodal large language models.
\newblock \emph{arXiv preprint arXiv:2605.27916}, 2026{\natexlab{b}}.

\bibitem[Du et~al.(2026)Du, Wang, Liu, Dvornek, and Lu]{du2026care}
Yuexi Du, Jinglu Wang, Shujie Liu, Nicha~C Dvornek, and Yan Lu.
\newblock Care: Towards clinical accountability in multi-modal medical reasoning with an evidence-grounded agentic framework.
\newblock \emph{arXiv preprint arXiv:2603.01607}, 2026.

\bibitem[Du et~al.(2025)Du, LujieZheng, Hu, Xu, Li, Sun, Chen, Wu, Cai, and Ying]{du2025llms}
Zhuoyun Du, LujieZheng LujieZheng, Renjun Hu, Yuyang Xu, Xiawei Li, Ying Sun, Wei Chen, Jian Wu, Haolei Cai, and Haochao Ying.
\newblock Llms can simulate standardized patients via agent coevolution.
\newblock In \emph{Proceedings of the 63rd Annual Meeting of the Association for Computational Linguistics (Volume 1: Long Papers)}, pages 17278--17306, 2025.

\bibitem[Ehtesham et~al.(2025{\natexlab{a}})Ehtesham, Singh, Gupta, and Kumar]{ehtesham2025survey}
Abul Ehtesham, Aditi Singh, Gaurav~Kumar Gupta, and Saket Kumar.
\newblock A survey of agent interoperability protocols: Model context protocol (mcp), agent communication protocol (acp), agent-to-agent protocol (a2a), and agent network protocol (anp).
\newblock \emph{arXiv preprint arXiv:2505.02279}, 2025{\natexlab{a}}.

\bibitem[Ehtesham et~al.(2025{\natexlab{b}})Ehtesham, Singh, and Kumar]{ehtesham2025enhancing}
Abul Ehtesham, Aditi Singh, and Saket Kumar.
\newblock Enhancing clinical decision support and ehr insights through llms and the model context protocol: An open-source mcp-fhir framework.
\newblock In \emph{2025 IEEE World AI IoT Congress (AIIoT)}, pages 0205--0211. IEEE, 2025{\natexlab{b}}.

\bibitem[Ekram(2026)]{ekram2026red}
Tashfeen Ekram.
\newblock Red-teaming medical ai: Systematic adversarial evaluation of llm safety guardrails in clinical contexts.
\newblock \emph{medRxiv}, pages 2026--02, 2026.

\bibitem[Elboardy et~al.(2025)Elboardy, Khoriba, and Rashed]{elboardy2025medical}
Ahmed~T Elboardy, Ghada Khoriba, and Essam~A Rashed.
\newblock Medical ai consensus: A multi-agent framework for radiology report generation and evaluation.
\newblock \emph{arXiv preprint arXiv:2509.17353}, 2025.

\bibitem[Erdur et~al.(2026)Erdur, Scholz, Pan, Wiestler, Rueckert, and Peeken]{erdur2026agentic}
Ayhan~Can Erdur, Daniel Scholz, Jiazhen Pan, Benedikt Wiestler, Daniel Rueckert, and Jan~C Peeken.
\newblock Agentic large language models for training-free neuro-radiological image analysis.
\newblock \emph{arXiv preprint arXiv:2604.16729}, 2026.

\bibitem[Fallahpour et~al.(2025)Fallahpour, Ma, Munim, Lyu, and Wang]{fallahpour2025medrax}
Adibvafa Fallahpour, Jun Ma, Alif Munim, Hongwei Lyu, and Bo~Wang.
\newblock Medrax: Medical reasoning agent for chest x-ray.
\newblock \emph{arXiv preprint arXiv:2502.02673}, 2025.

\bibitem[Fan et~al.(2026)Fan, Dai, Deng, Wang, Gong, Zheng, and Ou]{fan2026evolving}
Lin Fan, Pengyu Dai, Zhipeng Deng, Haolin Wang, Xun Gong, Yefeng Zheng, and Yafei Ou.
\newblock Evolving medical imaging agents via experience-driven self-skill discovery.
\newblock \emph{arXiv preprint arXiv:2603.05860}, 2026.

\bibitem[Fan et~al.(2025{\natexlab{a}})Fan, Sun, Xue, Zhang, Zhang, and Ruan]{fan2025medodyssey}
Yongqi Fan, Hongli Sun, Kui Xue, Xiaofan Zhang, Shaoting Zhang, and Tong Ruan.
\newblock Medodyssey: A medical domain benchmark for long context evaluation up to 200k tokens.
\newblock In \emph{Findings of the Association for Computational Linguistics: NAACL 2025}, pages 32--56, 2025{\natexlab{a}}.

\bibitem[Fan et~al.(2025{\natexlab{b}})Fan, Liang, Wu, Zhang, Wang, and Xie]{fan2025chestx}
Ziqing Fan, Cheng Liang, Chaoyi Wu, Ya~Zhang, Yanfeng Wang, and Weidi Xie.
\newblock Chestx-reasoner: Advancing radiology foundation models with reasoning through step-by-step verification.
\newblock \emph{arXiv preprint arXiv:2504.20930}, 2025{\natexlab{b}}.

\bibitem[Fang et~al.(2025{\natexlab{a}})Fang, Peng, Zhang, Wang, Yi, Zhang, Xu, Wu, Liu, Li, et~al.]{fang2025comprehensive}
Jinyuan Fang, Yanwen Peng, Xi~Zhang, Yingxu Wang, Xinhao Yi, Guibin Zhang, Yi~Xu, Bin Wu, Siwei Liu, Zihao Li, et~al.
\newblock A comprehensive survey of self-evolving ai agents: A new paradigm bridging foundation models and lifelong agentic systems.
\newblock \emph{arXiv preprint arXiv:2508.07407}, 2025{\natexlab{a}}.

\bibitem[Fang et~al.(2025{\natexlab{b}})Fang, Cai, Li, Wu, Li, Yin, Wang, Wang, Su, Zhang, et~al.]{fang2025towards}
Runnan Fang, Shihao Cai, Baixuan Li, Jialong Wu, Guangyu Li, Wenbiao Yin, Xinyu Wang, Xiaobin Wang, Liangcai Su, Zhen Zhang, et~al.
\newblock Towards general agentic intelligence via environment scaling.
\newblock \emph{arXiv preprint arXiv:2509.13311}, 2025{\natexlab{b}}.

\bibitem[Feng et~al.(2025{\natexlab{a}})Feng, Zheng, Wu, Zhao, Zhang, Wang, and Xie]{feng2025m}
Jinghao Feng, Qiaoyu Zheng, Chaoyi Wu, Ziheng Zhao, Ya~Zhang, Yanfeng Wang, and Weidi Xie.
\newblock M 3 builder: A multi-agent system for automated machine learning in medical imaging.
\newblock In \emph{International Workshop on Agentic AI for Medicine}, pages 115--124. Springer, 2025{\natexlab{a}}.

\bibitem[Feng et~al.(2025{\natexlab{b}})Feng, McDonald, and Zhang]{feng2025levels}
Kevin~J Feng, David~W McDonald, and Amy~X Zhang.
\newblock Levels of autonomy for ai agents.
\newblock \emph{arXiv preprint arXiv:2506.12469}, 2025{\natexlab{b}}.

\bibitem[Feng et~al.(2026)Feng, Wang, Zhou, Lei, and Li]{feng2026doctoragent}
Yichun Feng, Jiawei Wang, Lu~Zhou, Zhen Lei, and Yixue Li.
\newblock Doctoragent-rl: A multi-agent collaborative reinforcement learning system for multi-turn clinical dialogue.
\newblock In \emph{ICASSP 2026-2026 IEEE International Conference on Acoustics, Speech and Signal Processing (ICASSP)}, pages 16952--16956. IEEE, 2026.

\bibitem[Ferber et~al.(2025)Ferber, El~Nahhas, W{\"o}lflein, Wiest, Clusmann, Le{\ss}mann, Foersch, Lammert, Tschochohei, J{\"a}ger, et~al.]{ferber2025development}
Dyke Ferber, Omar~SM El~Nahhas, Georg W{\"o}lflein, Isabella~C Wiest, Jan Clusmann, Marie-Elisabeth Le{\ss}mann, Sebastian Foersch, Jacqueline Lammert, Maximilian Tschochohei, Dirk J{\"a}ger, et~al.
\newblock Development and validation of an autonomous artificial intelligence agent for clinical decision-making in oncology.
\newblock \emph{Nature cancer}, 6\penalty0 (8):\penalty0 1337--1349, 2025.

\bibitem[Fernando et~al.(2023)Fernando, Banarse, Michalewski, Osindero, and Rockt{\"a}schel]{fernando2023promptbreeder}
Chrisantha Fernando, Dylan Banarse, Henryk Michalewski, Simon Osindero, and Tim Rockt{\"a}schel.
\newblock Promptbreeder: Self-referential self-improvement via prompt evolution.
\newblock \emph{arXiv preprint arXiv:2309.16797}, 2023.

\bibitem[Froger et~al.(2025)Froger, Andrews, Bettini, Budhiraja, Cabral, Do, Garreau, Gaya, Lauren{\c{c}}on, Lecanu, et~al.]{froger2025scaling}
Romain Froger, Pierre Andrews, Matteo Bettini, Amar Budhiraja, Ricardo~Silveira Cabral, Virginie Do, Emilien Garreau, Jean-Baptiste Gaya, Hugo Lauren{\c{c}}on, Maxime Lecanu, et~al.
\newblock Are: Scaling up agent environments and evaluations.
\newblock \emph{arXiv preprint arXiv:2509.17158}, 2025.

\bibitem[Gao et~al.(2025{\natexlab{a}})Gao, Geng, Hua, Hu, Juan, Liu, Liu, Qiu, Qi, Wu, et~al.]{gao2025survey}
Huan-ang Gao, Jiayi Geng, Wenyue Hua, Mengkang Hu, Xinzhe Juan, Hongzhang Liu, Shilong Liu, Jiahao Qiu, Xuan Qi, Yiran Wu, et~al.
\newblock A survey of self-evolving agents: What, when, how, and where to evolve on the path to artificial super intelligence.
\newblock \emph{arXiv preprint arXiv:2507.21046}, 2025{\natexlab{a}}.

\bibitem[Gao et~al.(2023)Gao, Madaan, Zhou, Alon, Liu, Yang, Callan, and Neubig]{gao2023pal}
Luyu Gao, Aman Madaan, Shuyan Zhou, Uri Alon, Pengfei Liu, Yiming Yang, Jamie Callan, and Graham Neubig.
\newblock Pal: Program-aided language models.
\newblock In \emph{International conference on machine learning}, pages 10764--10799. PMLR, 2023.

\bibitem[Gao et~al.(2025{\natexlab{b}})Gao, Zhu, Kong, Noori, Su, Ginder, Tsiligkaridis, and Zitnik]{gao2025txagentaiagenttherapeutic}
Shanghua Gao, Richard Zhu, Zhenglun Kong, Ayush Noori, Xiaorui Su, Curtis Ginder, Theodoros Tsiligkaridis, and Marinka Zitnik.
\newblock Txagent: An ai agent for therapeutic reasoning across a universe of tools, 2025{\natexlab{b}}.
\newblock URL \url{https://arxiv.org/abs/2503.10970}.

\bibitem[Gao et~al.(2026)Gao, Zhou, Zhou, Zou, Zhang, and Fu]{gao2026enhancing}
Yifan Gao, Tao Zhou, Yi~Zhou, Ke~Zou, Yizhe Zhang, and Huazhu Fu.
\newblock Enhancing medical visual grounding via knowledge-guided spatial prompts.
\newblock \emph{arXiv preprint arXiv:2604.01915}, 2026.

\bibitem[Ge et~al.(2026)Ge, Li, Wang, Hu, Zhang, and Li]{ge2026clinicalagents}
Zhuohan Ge, Haoyang Li, Yubo Wang, Nicole Hu, Chen~Jason Zhang, and Qing Li.
\newblock Clinicalagents: Multi-agent orchestration for clinical decision making with dual-memory.
\newblock \emph{arXiv preprint arXiv:2603.26182}, 2026.

\bibitem[Ghafoor et~al.(2026)Ghafoor, Islam, Howlader, Khondokar, Bhattacharjee, Chakraborty, Roy, Bhattacharjee, and Roy]{ghafoor2026improving}
Zainab Ghafoor, Md~Shafiqul Islam, Koushik Howlader, Md~Rasel Khondokar, Tanusree Bhattacharjee, Sayantan Chakraborty, Adrito Roy, Ushashi Bhattacharjee, and Tirtho Roy.
\newblock Improving the safety and trustworthiness of medical ai via multi-agent evaluation loops.
\newblock \emph{arXiv preprint arXiv:2601.13268}, 2026.

\bibitem[Ghezloo et~al.(2025)Ghezloo, Seyfioglu, Soraki, Ikezogwo, Li, Vivekanandan, Elmore, Krishna, and Shapiro]{ghezloo2025pathfinder}
Fatemeh Ghezloo, Mehmet~Saygin Seyfioglu, Rustin Soraki, Wisdom~O Ikezogwo, Beibin Li, Tejoram Vivekanandan, Joann~G Elmore, Ranjay Krishna, and Linda Shapiro.
\newblock Pathfinder: A multi-modal multi-agent system for medical diagnostic decision-making applied to histopathology.
\newblock In \emph{Proceedings of the IEEE/CVF International Conference on Computer Vision}, pages 23431--23441, 2025.

\bibitem[Gong et~al.(2026)Gong, Fang, Yang, Tao, Guo, Wei, Xie, Guan, Chen, Shi, et~al.]{gong2026meddialogrubrics}
Lecheng Gong, Weimin Fang, Ting Yang, Dongjie Tao, Chunxiao Guo, Peng Wei, Bo~Xie, Jinqun Guan, Zixiao Chen, Fang Shi, et~al.
\newblock Meddialogrubrics: A comprehensive benchmark and evaluation framework for multi-turn medical consultations in large language models.
\newblock \emph{arXiv preprint arXiv:2601.03023}, 2026.

\bibitem[Grattafiori et~al.(2024)Grattafiori, Dubey, Jauhri, Pandey, Kadian, Al-Dahle, Letman, Mathur, Schelten, Vaughan, et~al.]{grattafiori2024llama}
Aaron Grattafiori, Abhimanyu Dubey, Abhinav Jauhri, Abhinav Pandey, Abhishek Kadian, Ahmad Al-Dahle, Aiesha Letman, Akhil Mathur, Alan Schelten, Alex Vaughan, et~al.
\newblock The llama 3 herd of models.
\newblock \emph{arXiv preprint arXiv:2407.21783}, 2024.

\bibitem[Gu et~al.(2025{\natexlab{a}})Gu, Zhou, Segal, Wu, Cao, Zhong, Clifton, Liu, and Clifton]{gu2025clinical}
Boyang Gu, Hongjian Zhou, Bradley~Max Segal, Jinge Wu, Zeyu Cao, Hantao Zhong, Lei Clifton, Fenglin Liu, and David~A Clifton.
\newblock Clinical-r1: Empowering large language models for faithful and comprehensive reasoning with clinical objective relative policy optimization.
\newblock \emph{arXiv preprint arXiv:2512.00601}, 2025{\natexlab{a}}.

\bibitem[Gu et~al.(2025{\natexlab{b}})Gu, Zhu, Sang, Wang, Sui, Tang, Harrison, Gao, Yu, and Ma]{gu2025medagentaudit}
Lei Gu, Yinghao Zhu, Haoran Sang, Zixiang Wang, Dehao Sui, Wen Tang, Ewen Harrison, Junyi Gao, Lequan Yu, and Liantao Ma.
\newblock Medagentaudit: Diagnosing and quantifying collaborative failure modes in medical multi-agent systems.
\newblock \emph{arXiv preprint arXiv:2510.10185}, 2025{\natexlab{b}}.

\bibitem[Guo et~al.(2025)Guo, Yang, Zhang, Song, Wang, Zhu, Xu, Zhang, Ma, Bi, et~al.]{guo2025deepseek}
Daya Guo, Dejian Yang, Haowei Zhang, Junxiao Song, Peiyi Wang, Qihao Zhu, Runxin Xu, Ruoyu Zhang, Shirong Ma, Xiao Bi, et~al.
\newblock Deepseek-r1: Incentivizing reasoning capability in llms via reinforcement learning.
\newblock \emph{arXiv preprint arXiv:2501.12948}, 2025.

\bibitem[Guo et~al.(2024)Guo, Chen, Wang, Chang, Pei, Chawla, Wiest, and Zhang]{guo2024large}
Taicheng Guo, Xiuying Chen, Yaqi Wang, Ruidi Chang, Shichao Pei, Nitesh~V Chawla, Olaf Wiest, and Xiangliang Zhang.
\newblock Large language model based multi-agents: A survey of progress and challenges.
\newblock \emph{arXiv preprint arXiv:2402.01680}, 2024.

\bibitem[Gupta and Kembhavi(2023)]{gupta2023visual}
Tanmay Gupta and Aniruddha Kembhavi.
\newblock Visual programming: Compositional visual reasoning without training.
\newblock In \emph{Proceedings of the IEEE/CVF conference on computer vision and pattern recognition}, pages 14953--14962, 2023.

\bibitem[Haidegger(2019)]{haidegger2019autonomy}
Tam{\'a}s Haidegger.
\newblock Autonomy for surgical robots: Concepts and paradigms.
\newblock \emph{IEEE Transactions on Medical Robotics and Bionics}, 1\penalty0 (2):\penalty0 65--76, 2019.

\bibitem[Hamamci et~al.(2026)Hamamci, Er, Shit, Reynaud, Yang, Guo, Edgar, Xu, Kainz, and Menze]{hamamci2026better}
Ibrahim~Ethem Hamamci, Sezgin Er, Suprosanna Shit, Hadrien Reynaud, Dong Yang, Pengfei Guo, Marc Edgar, Daguang Xu, Bernhard Kainz, and Bjoern Menze.
\newblock Better tokens for better 3d: Advancing vision-language modeling in 3d medical imaging.
\newblock \emph{Advances in Neural Information Processing Systems}, 38:\penalty0 135074--135102, 2026.

\bibitem[Han et~al.(2024)Han, Kumar, Agarwal, and Lakkaraju]{han2024medsafetybench}
Tessa Han, Aounon Kumar, Chirag Agarwal, and Himabindu Lakkaraju.
\newblock Medsafetybench: Evaluating and improving the medical safety of large language models.
\newblock \emph{Advances in neural information processing systems}, 37:\penalty0 33423--33454, 2024.

\bibitem[Hasan et~al.(2025)Hasan, Li, Fallahzadeh, Rajbahadur, Adams, and Hassan]{hasan2025model}
Mohammed~Mehedi Hasan, Hao Li, Emad Fallahzadeh, Gopi~Krishnan Rajbahadur, Bram Adams, and Ahmed~E Hassan.
\newblock Model context protocol (mcp) at first glance: Studying the security and maintainability of mcp servers.
\newblock \emph{ACM Transactions on Software Engineering and Methodology}, 2025.

\bibitem[He et~al.(2026)He, Zhou, Wang, Xu, Liu, and Miao]{he2026harness}
Chaoyue He, Xin Zhou, Di~Wang, Hong Xu, Wei Liu, and Chunyan Miao.
\newblock Harness engineering for language agents: The harness layer as control, agency, and runtime.
\newblock 2026.

\bibitem[He et~al.(2020)He, Zhang, Mou, Xing, and Xie]{he2020pathvqa}
Xuehai He, Yichen Zhang, Luntian Mou, Eric Xing, and Pengtao Xie.
\newblock Pathvqa: 30000+ questions for medical visual question answering.
\newblock \emph{arXiv preprint arXiv:2003.10286}, 2020.

\bibitem[He et~al.(2025)He, Li, Liu, Yao, and He]{he2025medorch}
Yexiao He, Ang Li, Boyi Liu, Zhewei Yao, and Yuxiong He.
\newblock Medorch: Medical diagnosis with tool-augmented reasoning agents for flexible extensibility.
\newblock \emph{arXiv preprint arXiv:2506.00235}, 2025.

\bibitem[Holmes et~al.(2025)Holmes, Hao, Borras-Osorio, Mastroleo, Brufau, Carducci, Van~Abel, Routman, Foong, Muller, et~al.]{holmes2025radonc}
Jason Holmes, Yuexing Hao, Mariana Borras-Osorio, Federico Mastroleo, Santiago~Romero Brufau, Valentina Carducci, Katie~M Van~Abel, David~M Routman, Andrew~YK Foong, Liv~M Muller, et~al.
\newblock Radonc-gpt: An autonomous llm agent for real-time patient outcomes labeling at scale.
\newblock \emph{arXiv preprint arXiv:2509.25540}, 2025.

\bibitem[Hong et~al.(2024)Hong, Zhuge, Chen, Zheng, Cheng, Wang, Zhang, Yau, Lin, Zhou, et~al.]{hong2024metagpt}
Sirui Hong, Mingchen Zhuge, Jonathan Chen, Xiawu Zheng, Yuheng Cheng, Jinlin Wang, Ceyao Zhang, Steven Yau, Zijuan Lin, Liyang Zhou, et~al.
\newblock Metagpt: Meta programming for a multi-agent collaborative framework.
\newblock In \emph{International Conference on Learning Representations}, volume 2024, pages 23247--23275, 2024.

\bibitem[Hou et~al.(2025{\natexlab{a}})Hou, Yang, Du, Lau, Liu, He, Long, and Wang]{hou2025adagent}
Wenlong Hou, Guangqian Yang, Ye~Du, Yeung Lau, Lihao Liu, Junjun He, Ling Long, and Shujun Wang.
\newblock Adagent: Llm agent for alzheimer’s disease analysis with collaborative coordinator.
\newblock In \emph{International Workshop on Agentic AI for Medicine}, pages 23--32. Springer, 2025{\natexlab{a}}.

\bibitem[Hou et~al.(2026)Hou, Bi, Yang, Liu, Du, Xue, Wang, Feng, Xun, Yu, Mao, Yang, Cheung, Long, Tan, Yu, Ma, Yan, and Wang]{hou2026adcareguidelinegroundedmodalityagnosticllm}
Wenlong Hou, Sheng Bi, Guangqian Yang, Lihao Liu, Ye~Du, Hanxiao Xue, Juncheng Wang, Yuxiang Feng, Yue Xun, Nanxi Yu, Ning Mao, Mo~Yang, Yi~Wah~Eva Cheung, Ling Long, Kay~Chen Tan, Lequan Yu, Xiaomeng Ma, Shaozhen Yan, and Shujun Wang.
\newblock Ad-care: A guideline-grounded, modality-agnostic llm agent for real-world alzheimer's disease diagnosis with multi-cohort assessment, fairness analysis, and reader study, 2026.
\newblock URL \url{https://arxiv.org/abs/2603.25322}.

\bibitem[Hou et~al.(2025{\natexlab{b}})Hou, Zhao, Wang, and Wang]{hou2025model}
Xinyi Hou, Yanjie Zhao, Shenao Wang, and Haoyu Wang.
\newblock Model context protocol (mcp): Landscape, security threats, and future research directions.
\newblock \emph{ACM Transactions on Software Engineering and Methodology}, 2025{\natexlab{b}}.

\bibitem[Hu(2025)]{hu2025reinforce++}
Jian Hu.
\newblock Reinforce++: A simple and efficient approach for aligning large language models.
\newblock \emph{arXiv e-prints}, pages arXiv--2501, 2025.

\bibitem[Hu et~al.(2025)Hu, Lu, and Clune]{hu2025automated}
Shengran Hu, Cong Lu, and Jeff Clune.
\newblock Automated design of agentic systems.
\newblock In \emph{International Conference on Learning Representations}, volume 2025, pages 21344--21377, 2025.

\bibitem[Hu et~al.(2024)Hu, Li, Lu, Shao, He, Qiao, and Luo]{hu2024omnimedvqa}
Yutao Hu, Tianbin Li, Quanfeng Lu, Wenqi Shao, Junjun He, Yu~Qiao, and Ping Luo.
\newblock Omnimedvqa: A new large-scale comprehensive evaluation benchmark for medical lvlm.
\newblock In \emph{Proceedings of the IEEE/CVF Conference on Computer Vision and Pattern Recognition}, pages 22170--22183, 2024.

\bibitem[Huang et~al.(2023)Huang, Gu, Hou, Wu, Wang, Yu, and Han]{huang2023large}
Jiaxin Huang, Shixiang Gu, Le~Hou, Yuexin Wu, Xuezhi Wang, Hongkun Yu, and Jiawei Han.
\newblock Large language models can self-improve.
\newblock In \emph{Proceedings of the 2023 conference on empirical methods in natural language processing}, pages 1051--1068, 2023.

\bibitem[Huang et~al.(2025)Huang, Li, Liu, Fan, Fung, et~al.]{huang2025scaling}
Yuchen Huang, Sijia Li, Wei Liu, Zhiyuan Fan, Yi~R Fung, et~al.
\newblock Scaling environments for llm agents in the era of learning from interaction: A survey.
\newblock In \emph{Workshop on Scaling Environments for Agents}, 2025.

\bibitem[Imam et~al.(2025)Imam, Marew, and Yaqub]{imam2025robustness}
Raza Imam, Rufael Marew, and Mohammad Yaqub.
\newblock On the robustness of medical vision-language models: Are they truly generalizable?
\newblock In \emph{Annual Conference on Medical Image Understanding and Analysis}, pages 233--256. Springer, 2025.

\bibitem[Jain et~al.(2021)Jain, Agrawal, Saporta, Truong, Duong, Bui, Chambon, Zhang, Lungren, Ng, et~al.]{jain2021radgraph}
Saahil Jain, Ashwin Agrawal, Adriel Saporta, Steven~QH Truong, Du~Nguyen Duong, Tan Bui, Pierre Chambon, Yuhao Zhang, Matthew~P Lungren, Andrew~Y Ng, et~al.
\newblock Radgraph: Extracting clinical entities and relations from radiology reports.
\newblock \emph{arXiv preprint arXiv:2106.14463}, 2021.

\bibitem[Jeong(2026)]{jeong2026healthcare}
Minbyul Jeong.
\newblock Healthcare ai gym for medical agents.
\newblock \emph{arXiv preprint arXiv:2605.02943}, 2026.

\bibitem[Jhandi et~al.(2025)Jhandi, Kazi, Subramanian, and Sendas]{jhandi2025small}
Polaris Jhandi, Owais Kazi, Shreyas Subramanian, and Neel Sendas.
\newblock Small language models for efficient agentic tool calling: Outperforming large models with targeted fine-tuning.
\newblock \emph{arXiv preprint arXiv:2512.15943}, 2025.

\bibitem[Jhaveri et~al.(2025)Jhaveri, Singh, Kim, Taghavi, and Kenthapadi]{jhaveri2025claimreward}
Samyak Jhaveri, Praphul Singh, Jangwon Kim, Tara Taghavi, and Krishnaram Kenthapadi.
\newblock Optimizing long-form clinical text generation with claim-based rewards.
\newblock \emph{arXiv preprint arXiv:2510.02338}, 2025.

\bibitem[Ji et~al.(2023)Ji, Lee, Frieske, Yu, Su, Xu, Ishii, Bang, Madotto, and Fung]{ji2023survey}
Ziwei Ji, Nayeon Lee, Rita Frieske, Tiezheng Yu, Dan Su, Yan Xu, Etsuko Ishii, Ye~Jin Bang, Andrea Madotto, and Pascale Fung.
\newblock Survey of hallucination in natural language generation.
\newblock \emph{ACM computing surveys}, 55\penalty0 (12):\penalty0 1--38, 2023.

\bibitem[Jiang et~al.(2025{\natexlab{a}})Jiang, Li, Wan, Yin, Wu, Liang, Li, Sun, Wang, Chang, et~al.]{jiang2025dynamic}
Eric~Hanchen Jiang, Mengting Li, Guancheng Wan, Sophia Yin, Yuchen Wu, Xiao Liang, Xinfeng Li, Yizhou Sun, Wei Wang, Kai-Wei Chang, et~al.
\newblock Dynamic generation of multi-llm agents communication topologies with graph diffusion models.
\newblock \emph{arXiv preprint arXiv:2510.07799}, 2025{\natexlab{a}}.

\bibitem[Jiang et~al.(2025{\natexlab{b}})Jiang, Black, Geng, Park, Zou, Ng, and Chen]{jiang2025medagentbench}
Yixing Jiang, Kameron~C Black, Gloria Geng, Danny Park, James Zou, Andrew~Y Ng, and Jonathan~H Chen.
\newblock Medagentbench: a virtual ehr environment to benchmark medical llm agents.
\newblock \emph{Nejm Ai}, 2\penalty0 (9):\penalty0 AIdbp2500144, 2025{\natexlab{b}}.

\bibitem[Jimenez et~al.(2024)Jimenez, Yang, Wettig, Yao, Pei, Press, and Narasimhan]{jimenez2024swe}
Carlos~E Jimenez, John Yang, Alexander Wettig, Shunyu Yao, Kexin Pei, Ofir Press, and Karthik Narasimhan.
\newblock Swe-bench: Can language models resolve real-world github issues?
\newblock In \emph{International Conference on Learning Representations}, volume 2024, pages 54107--54157, 2024.

\bibitem[Jin et~al.(2021)Jin, Pan, Oufattole, Weng, Fang, and Szolovits]{jin2021disease}
Di~Jin, Eileen Pan, Nassim Oufattole, Wei-Hung Weng, Hanyi Fang, and Peter Szolovits.
\newblock What disease does this patient have? a large-scale open domain question answering dataset from medical exams.
\newblock \emph{Applied Sciences}, 11\penalty0 (14):\penalty0 6421, 2021.

\bibitem[Jin et~al.(2019)Jin, Dhingra, Liu, Cohen, and Lu]{jin2019pubmedqa}
Qiao Jin, Bhuwan Dhingra, Zhengping Liu, William Cohen, and Xinghua Lu.
\newblock Pubmedqa: A dataset for biomedical research question answering.
\newblock In \emph{Proceedings of the 2019 conference on empirical methods in natural language processing and the 9th international joint conference on natural language processing (EMNLP-IJCNLP)}, pages 2567--2577, 2019.

\bibitem[Jing et~al.(2025)Jing, Lee, Zhang, Zhou, Yuan, Gao, Zhu, Papanastasiou, Fang, and Yang]{jing2025reason}
Peiyuan Jing, Kinhei Lee, Zhenxuan Zhang, Huichi Zhou, Zhengqing Yuan, Zhifan Gao, Lei Zhu, Giorgos Papanastasiou, Yingying Fang, and Guang Yang.
\newblock Reason like a radiologist: Chain-of-thought and reinforcement learning for verifiable report generation.
\newblock \emph{Medical Image Analysis}, page 103910, 2025.

\bibitem[Jorf and Shamout(2026)]{jorf2026agentrx}
Baraa~Al Jorf and Farah~E Shamout.
\newblock Agentrx: A benchmark study of llm agents for multimodal clinical prediction tasks.
\newblock \emph{arXiv preprint arXiv:2605.10286}, 2026.

\bibitem[Kaissis et~al.(2020)Kaissis, Makowski, R{\"u}ckert, and Braren]{kaissis2020secure}
Georgios~A Kaissis, Marcus~R Makowski, Daniel R{\"u}ckert, and Rickmer~F Braren.
\newblock Secure, privacy-preserving and federated machine learning in medical imaging.
\newblock \emph{Nature Machine Intelligence}, 2\penalty0 (6):\penalty0 305--311, 2020.

\bibitem[Karim and Uzuner(2025)]{karim2025multimodal}
AHM Karim and Ozlem Uzuner.
\newblock Multimodal retrieval-augmented generation with large language models for medical vqa.
\newblock \emph{arXiv preprint arXiv:2510.13856}, 2025.

\bibitem[Khandekar et~al.(2024)Khandekar, Jin, Xiong, Dunn, Applebaum, Anwar, Sarfo-Gyamfi, Safranek, Anwar, Zhang, et~al.]{khandekar2024medcalc}
Nikhil Khandekar, Qiao Jin, Guangzhi Xiong, Soren Dunn, Serina~S Applebaum, Zain Anwar, Maame Sarfo-Gyamfi, Conrad~W Safranek, Abid~A Anwar, Andrew Zhang, et~al.
\newblock Medcalc-bench: Evaluating large language models for medical calculations.
\newblock \emph{Advances in Neural Information Processing Systems}, 37:\penalty0 84730--84745, 2024.

\bibitem[Khosravi et~al.(2026)Khosravi, Rouzrokh, Akinci~D’Antonoli, Moassefi, Faghani, Mansuri, Bressem, Tejani, and Gichoya]{khosravi2026agentic}
Bardia Khosravi, Pouria Rouzrokh, Tugba Akinci~D’Antonoli, Mana Moassefi, Shahriar Faghani, Aawez Mansuri, Keno Bressem, Ali Tejani, and Judy Gichoya.
\newblock Agentic ai in radiology: evolution from large language models to future clinical integration.
\newblock \emph{Radiology: Artificial Intelligence}, 8\penalty0 (2):\penalty0 e250651, 2026.

\bibitem[Kim et~al.(2024)Kim, Park, Jeong, Chan, Xu, McDuff, Lee, Ghassemi, Breazeal, and Park]{kim2024mdagents}
Yubin Kim, Chanwoo Park, Hyewon Jeong, Yik~S Chan, Xuhai Xu, Daniel McDuff, Hyeonhoon Lee, Marzyeh Ghassemi, Cynthia Breazeal, and Hae~W Park.
\newblock Mdagents: An adaptive collaboration of llms for medical decision-making.
\newblock \emph{Advances in Neural Information Processing Systems}, 37:\penalty0 79410--79452, 2024.

\bibitem[Kim et~al.(2025{\natexlab{a}})Kim, Gu, Park, Park, Schmidgall, Heydari, Yan, Zhang, Zhuang, Liu, et~al.]{kim2025towards}
Yubin Kim, Ken Gu, Chanwoo Park, Chunjong Park, Samuel Schmidgall, A~Ali Heydari, Yao Yan, Zhihan Zhang, Yuchen Zhuang, Yun Liu, et~al.
\newblock Towards a science of scaling agent systems.
\newblock \emph{arXiv preprint arXiv:2512.08296}, 2025{\natexlab{a}}.

\bibitem[Kim et~al.(2025{\natexlab{b}})Kim, Jeong, Chen, Li, Park, Lu, Alhamoud, Mun, Grau, Jung, et~al.]{kim2025medical}
Yubin Kim, Hyewon Jeong, Shan Chen, Shuyue~Stella Li, Chanwoo Park, Mingyu Lu, Kumail Alhamoud, Jimin Mun, Cristina Grau, Minseok Jung, et~al.
\newblock Medical hallucinations in foundation models and their impact on healthcare.
\newblock \emph{arXiv preprint arXiv:2503.05777}, 2025{\natexlab{b}}.

\bibitem[Klila et~al.(2026)Klila, Souihi, Boujelben, Semmar, and Belguith]{klila2026injecting}
Jaafer Klila, Sondes~Bannour Souihi, Rahma Boujelben, Nasredine Semmar, and Lamia~Hadrich Belguith.
\newblock Injecting structured biomedical knowledge into language models: Continual pretraining vs. graphrag.
\newblock \emph{arXiv preprint arXiv:2604.16422}, 2026.

\bibitem[Kulkarni and Kulkarni(2026)]{kulkarni2026benchmarking}
Siddhant Kulkarni and Yukta Kulkarni.
\newblock Benchmarking multi-agent llm architectures for financial document processing: A comparative study of orchestration patterns, cost-accuracy tradeoffs and production scaling strategies.
\newblock \emph{arXiv preprint arXiv:2603.22651}, 2026.

\bibitem[Kweon et~al.(2024)Kweon, Kim, Kwak, Cha, Yoon, Kim, Yang, Won, and Choi]{kweon2024ehrnoteqa}
Sunjun Kweon, Jiyoun Kim, Heeyoung Kwak, Dongchul Cha, Hangyul Yoon, Kwanghyun Kim, Jeewon Yang, Seunghyun Won, and Edward Choi.
\newblock Ehrnoteqa: An llm benchmark for real-world clinical practice using discharge summaries.
\newblock \emph{Advances in Neural Information Processing Systems}, 37:\penalty0 124575--124611, 2024.

\bibitem[Kyung et~al.(2026)Kyung, Chung, Bae, Kim, Sohn, Kim, Kim, and Choi]{kyung2026patientsim}
Daeun Kyung, Hyunseung Chung, Seongsu Bae, Jiho Kim, Jae~Ho Sohn, Taerim Kim, Soo~Kyung Kim, and Edward Choi.
\newblock Patientsim: A persona-driven simulator for realistic doctor-patient interactions.
\newblock \emph{Advances in Neural Information Processing Systems}, 38, 2026.

\bibitem[Lai et~al.(2026)Lai, Zhong, Li, Zhao, Li, Psounis, and Yang]{lai2026med}
Yuxiang Lai, Jike Zhong, Ming Li, Shitian Zhao, Yuheng Li, Konstantinos Psounis, and Xiaofeng Yang.
\newblock Med-r1: Reinforcement learning for generalizable medical reasoning in vision-language models.
\newblock \emph{IEEE Transactions on Medical Imaging}, 2026.

\bibitem[Lau et~al.(2018)Lau, Gayen, Ben~Abacha, and Demner-Fushman]{lau2018dataset}
Jason~J Lau, Soumya Gayen, Asma Ben~Abacha, and Dina Demner-Fushman.
\newblock A dataset of clinically generated visual questions and answers about radiology images.
\newblock \emph{Scientific data}, 5\penalty0 (1):\penalty0 180251, 2018.

\bibitem[Lee et~al.(2025)Lee, Bach, Yang, Pollard, Johnson, Choi, Lee, et~al.]{lee2025fhir}
Gyubok Lee, Elea Bach, Eric Yang, Tom Pollard, Alistair Johnson, Edward Choi, Jong~Ha Lee, et~al.
\newblock Fhir-agentbench: Benchmarking llm agents for realistic interoperable ehr question answering.
\newblock \emph{arXiv preprint arXiv:2509.19319}, 2025.

\bibitem[Lee et~al.(2026)Lee, Yoon, and Choi]{lee2026cxreasonagent}
Hyungyung Lee, Hangyul Yoon, and Edward Choi.
\newblock Cxreasonagent: Evidence-grounded diagnostic reasoning agent for chest x-rays.
\newblock \emph{arXiv preprint arXiv:2602.23276}, 2026.

\bibitem[Lee and Hockenmaier(2025)]{lee2025evaluating}
Jinu Lee and Julia Hockenmaier.
\newblock Evaluating step-by-step reasoning traces: A survey.
\newblock \emph{arXiv preprint arXiv:2502.12289}, 2025.

\bibitem[Leong et~al.(2026)Leong, Wolf, Channa, Wang, Lehmann, Abramoff, and Liu]{leong2026autonomous}
Ariel Leong, Risa~M Wolf, Roomasa Channa, Jiangxia Wang, Harold Lehmann, Michael~D Abramoff, and TY~Liu.
\newblock Autonomous ai-assisted diabetic retinopathy screening at primary care is associated with increased presentation to eye care by at risk patients.
\newblock \emph{NPJ digital medicine}, 9\penalty0 (1):\penalty0 310, 2026.

\bibitem[Lewis et~al.(2020)Lewis, Perez, Piktus, Petroni, Karpukhin, Goyal, K{\"u}ttler, Lewis, Yih, Rockt{\"a}schel, et~al.]{lewis2020retrieval}
Patrick Lewis, Ethan Perez, Aleksandra Piktus, Fabio Petroni, Vladimir Karpukhin, Naman Goyal, Heinrich K{\"u}ttler, Mike Lewis, Wen-tau Yih, Tim Rockt{\"a}schel, et~al.
\newblock Retrieval-augmented generation for knowledge-intensive nlp tasks.
\newblock \emph{Advances in neural information processing systems}, 33:\penalty0 9459--9474, 2020.

\bibitem[Li et~al.(2024{\natexlab{a}})Li, Yan, Pan, Luo, Ji, Ding, Xu, Liu, Dong, Lin, et~al.]{li2024mmedagent}
Binxu Li, Tiankai Yan, Yuanting Pan, Jie Luo, Ruiyang Ji, Jiayuan Ding, Zhe Xu, Shilong Liu, Haoyu Dong, Zihao Lin, et~al.
\newblock Mmedagent: Learning to use medical tools with multi-modal agent.
\newblock In \emph{Findings of the Association for Computational Linguistics: EMNLP 2024}, pages 8745--8760, 2024{\natexlab{a}}.

\bibitem[Li et~al.(2023)Li, Wong, Zhang, Usuyama, Liu, Yang, Naumann, Poon, and Gao]{li2023llava}
Chunyuan Li, Cliff Wong, Sheng Zhang, Naoto Usuyama, Haotian Liu, Jianwei Yang, Tristan Naumann, Hoifung Poon, and Jianfeng Gao.
\newblock Llava-med: Training a large language-and-vision assistant for biomedicine in one day.
\newblock \emph{Advances in Neural Information Processing Systems}, 36:\penalty0 28541--28564, 2023.

\bibitem[Li et~al.(2026)Li, Zhou, Meng, Vadera, Li, and Li]{li2026turn}
Junbo Li, Peng Zhou, Rui Meng, Meet~P Vadera, Lihong Li, and Yang Li.
\newblock Turn-ppo: Turn-level advantage estimation with ppo for improved multi-turn rl in agentic llms.
\newblock In \emph{Findings of the Association for Computational Linguistics: EACL 2026}, pages 6227--6243, 2026.

\bibitem[Li et~al.(2024{\natexlab{b}})Li, Lai, Li, Ren, Zhang, Kang, Wang, Li, Zhang, Ma, et~al.]{li2024agent}
Junkai Li, Yunghwei Lai, Weitao Li, Jingyi Ren, Meng Zhang, Xinhui Kang, Siyu Wang, Peng Li, Ya-Qin Zhang, Weizhi Ma, et~al.
\newblock Agent hospital: A simulacrum of hospital with evolvable medical agents.
\newblock \emph{arXiv preprint arXiv:2405.02957}, 2024{\natexlab{b}}.

\bibitem[Li et~al.(2025{\natexlab{a}})Li, Wang, Berlowitz, Mez, Lin, and Yu]{li2025care}
Rumeng Li, Xun Wang, Dan Berlowitz, Jesse Mez, Honghuang Lin, and Hong Yu.
\newblock Care-ad: a multi-agent large language model framework for alzheimer’s disease prediction using longitudinal clinical notes.
\newblock \emph{npj Digital Medicine}, 8\penalty0 (1):\penalty0 541, 2025{\natexlab{a}}.

\bibitem[Li et~al.(2025{\natexlab{b}})Li, Lin, Lin, Zhang, Liu, Yang, Li, He, Song, Xiao, et~al.]{li2025eyecaregpt}
Sijing Li, Tianwei Lin, Lingshuai Lin, Wenqiao Zhang, Jiang Liu, Xiaoda Yang, Juncheng Li, Yucheng He, Xiaohui Song, Jun Xiao, et~al.
\newblock Eyecaregpt: Boosting comprehensive ophthalmology understanding with tailored dataset, benchmark and model.
\newblock In \emph{Proceedings of the 33rd ACM International Conference on Multimedia}, pages 3893--3902, 2025{\natexlab{b}}.

\bibitem[Li et~al.(2025{\natexlab{c}})Li, Xu, Bao, Liu, Liu, Liu, Wang, Lei, Wang, Xu, et~al.]{li2025co}
Songhao Li, Jonathan Xu, Tiancheng Bao, Yuxuan Liu, Yuchen Liu, Yihang Liu, Lilin Wang, Wenhui Lei, Sheng Wang, Yinuo Xu, et~al.
\newblock A co-evolving agentic ai system for medical imaging analysis.
\newblock \emph{arXiv preprint arXiv:2509.20279}, 2025{\natexlab{c}}.

\bibitem[Li et~al.(2025{\natexlab{d}})Li, Yan, Zhang, Chen, Zhu, Zhao, Li, Li, Cao, Jiang, et~al.]{li2025macd}
Wenliang Li, Rui Yan, Xu~Zhang, Li~Chen, Hongji Zhu, Jing Zhao, Junjun Li, Mengru Li, Wei Cao, Zihang Jiang, et~al.
\newblock Macd: Multi-agent clinical diagnosis with self-learned knowledge for llm.
\newblock \emph{arXiv preprint arXiv:2509.20067}, 2025{\natexlab{d}}.

\bibitem[Liao et~al.(2025)Liao, Jiang, Wang, and Wang]{liao2025reflectool}
Yusheng Liao, Shuyang Jiang, Yanfeng Wang, and Yu~Wang.
\newblock Reflectool: Towards reflection-aware tool-augmented clinical agents.
\newblock In \emph{Proceedings of the 63rd Annual Meeting of the Association for Computational Linguistics (Volume 1: Long Papers)}, pages 13507--13531, 2025.

\bibitem[Lin et~al.(2026{\natexlab{a}})Lin, Zhu, Kneuertz, Bai, and Xue]{lin2026medcausalx}
Jianxin Lin, Chunzheng Zhu, Peter~J Kneuertz, Yunfei Bai, and Yuan Xue.
\newblock Medcausalx: Adaptive causal reasoning with self-reflection for trustworthy medical vision-language models.
\newblock \emph{arXiv preprint arXiv:2603.23085}, 2026{\natexlab{a}}.

\bibitem[Lin et~al.(2025)Lin, Zhang, Li, Yuan, Yu, Li, He, Jiang, Li, Song, et~al.]{lin2025healthgpt}
Tianwei Lin, Wenqiao Zhang, Sijing Li, Yuqian Yuan, Binhe Yu, Haoyuan Li, Wanggui He, Hao Jiang, Mengze Li, Xiaohui Song, et~al.
\newblock Healthgpt: A medical large vision-language model for unifying comprehension and generation via heterogeneous knowledge adaptation.
\newblock \emph{arXiv preprint arXiv:2502.09838}, 2025.

\bibitem[Lin et~al.(2017)Lin, Doll{\'a}r, Girshick, He, Hariharan, and Belongie]{lin2017feature}
Tsung-Yi Lin, Piotr Doll{\'a}r, Ross Girshick, Kaiming He, Bharath Hariharan, and Serge Belongie.
\newblock Feature pyramid networks for object detection.
\newblock In \emph{Proceedings of the IEEE conference on computer vision and pattern recognition}, pages 2117--2125, 2017.

\bibitem[Lin et~al.(2026{\natexlab{b}})Lin, Ding, Wu, and Peng]{lin2026march}
Yi~Lin, Yihao Ding, Yonghui Wu, and Yifan Peng.
\newblock March: Multi-agent radiology clinical hierarchy for ct report generation.
\newblock \emph{arXiv preprint arXiv:2604.16175}, 2026{\natexlab{b}}.

\bibitem[Liu et~al.(2021)Liu, Zhan, Xu, Ma, Yang, and Wu]{liu2021slake}
Bo~Liu, Li-Ming Zhan, Li~Xu, Lin Ma, Yan Yang, and Xiao-Ming Wu.
\newblock Slake: A semantically-labeled knowledge-enhanced dataset for medical visual question answering.
\newblock In \emph{2021 IEEE 18th international symposium on biomedical imaging (ISBI)}, pages 1650--1654. IEEE, 2021.

\bibitem[Liu et~al.(2025{\natexlab{a}})Liu, Zou, Zhan, Lu, Dong, Chen, Xie, Cao, Wu, and Fu]{liu2025gemex}
Bo~Liu, Ke~Zou, Li-Ming Zhan, Zexin Lu, Xiaoyu Dong, Yidi Chen, Chengqiang Xie, Jiannong Cao, Xiao-Ming Wu, and Huazhu Fu.
\newblock Gemex: A large-scale, groundable, and explainable medical vqa benchmark for chest x-ray diagnosis.
\newblock In \emph{Proceedings of the IEEE/CVF International Conference on Computer Vision}, pages 21310--21320, 2025{\natexlab{a}}.

\bibitem[Liu et~al.(2024)Liu, Xue, Chen, Chen, Zhao, Wang, Hou, Li, and Peng]{liu2024survey}
Hanchao Liu, Wenyuan Xue, Yifei Chen, Dapeng Chen, Xiutian Zhao, Ke~Wang, Liping Hou, Rongjun Li, and Wei Peng.
\newblock A survey on hallucination in large vision-language models.
\newblock \emph{arXiv preprint arXiv:2402.00253}, 2024.

\bibitem[Liu et~al.(2025{\natexlab{b}})Liu, Zhu, Wang, Long, Lai, Yu, and Zhao]{liu2025medmmv}
Hongjun Liu, Yinghao Zhu, Yuhui Wang, Yitao Long, Zeyu Lai, Lequan Yu, and Chen Zhao.
\newblock Medmmv: A controllable multimodal multi-agent framework for reliable and verifiable clinical reasoning.
\newblock \emph{arXiv preprint arXiv:2509.24314}, 2025{\natexlab{b}}.

\bibitem[Liu et~al.(2026{\natexlab{a}})Liu, Wang, Ma, Huang, Su, Chang, Li, Shen, Lyu, and Chen]{liu2026medchain}
Jie Liu, Wenxuan Wang, Zizhan Ma, Guolin Huang, Yihang Su, Kao-Jung Chang, Haoliang Li, Linlin Shen, Michael~R Lyu, and Wenting Chen.
\newblock Medchain: Bridging the gap between llm agents and clinical practice with interactive sequence.
\newblock \emph{Advances in Neural Information Processing Systems}, 38, 2026{\natexlab{a}}.

\bibitem[Liu et~al.(2026{\natexlab{b}})Liu, Song, Wang, Mao, Chen, Huang, Qi, Guo, Tang, He, et~al.]{liu2026automedbench}
Junqi Liu, Salena Song, Yuhan Wang, Jiawei Mao, Hardy Chen, Xiaoke Huang, Tianhao Qi, Pengfei Guo, Yucheng Tang, Yufan He, et~al.
\newblock Automedbench: Towards medical autoresearch with agentic ai models.
\newblock \emph{arXiv preprint arXiv:2606.01961}, 2026{\natexlab{b}}.

\bibitem[Liu et~al.(2025{\natexlab{c}})Liu, Bansal, Dinh, Pawar, Satishkumar, Desai, Gupta, Wang, and Hu]{liu2025medchat}
Philip~R Liu, Sparsh Bansal, Jimmy Dinh, Aditya Pawar, Ramani Satishkumar, Shail Desai, Neeraj Gupta, Xin Wang, and Shu Hu.
\newblock Medchat: A multi-agent framework for multimodal diagnosis with large language models.
\newblock In \emph{2025 IEEE 8th International Conference on Multimedia Information Processing and Retrieval (MIPR)}, pages 456--462. IEEE, 2025{\natexlab{c}}.

\bibitem[Liu et~al.(2026{\natexlab{c}})Liu, Zhang, Qin, Valanarasu, Rokuss, Lu, Ossowski, Chaves, Wong, Argaw, et~al.]{liu2026healthagentbench}
Qianchu Liu, Sheng Zhang, Guanghui Qin, Jeya Maria~Jose Valanarasu, Maximilian Rokuss, Mingyu Lu, Timothy Ossowski, Juan Manuel~Zambrano Chaves, Cliff Wong, Peniel Argaw, et~al.
\newblock Healthagentbench: A unified benchmark suite of realistic agentic healthcare environments for challenging frontier ai agents.
\newblock \emph{arXiv preprint arXiv:2606.31179}, 2026{\natexlab{c}}.

\bibitem[Liu et~al.(2026{\natexlab{d}})Liu, Mohiuddin, Schoeffler, Renduchintala, Nayak, Vemu, Vedak, Black, Havlik, Ogunmola, et~al.]{liu2026physicianbench}
Ruoqi Liu, Imran~Q Mohiuddin, Austin~J Schoeffler, Kavita Renduchintala, Ashwin Nayak, Prasantha~L Vemu, Shivam~C Vedak, Kameron~C Black, John~L Havlik, Isaac Ogunmola, et~al.
\newblock Physicianbench: Evaluating llm agents in real-world ehr environments.
\newblock \emph{arXiv preprint arXiv:2605.02240}, 2026{\natexlab{d}}.

\bibitem[Liu et~al.(2026{\natexlab{e}})Liu, Bao, Yang, Geng, Zheng, Li, Chen, Peng, and Yuan]{liu2026medsam}
Shengyuan Liu, Liuxin Bao, Qi~Yang, Wanting Geng, Boyun Zheng, Chenxin Li, Wenting Chen, Houwen Peng, and Yixuan Yuan.
\newblock Medsam-agent: Empowering interactive medical image segmentation with multi-turn agentic reinforcement learning.
\newblock \emph{arXiv preprint arXiv:2602.03320}, 2026{\natexlab{e}}.

\bibitem[Liu et~al.(2025{\natexlab{d}})Liu, Xuan, Wu, Humphrey, DiStasio, Kahila, Tan, Qi, Yang, Han, et~al.]{liu2025teampath}
Tianyu Liu, Weihao Xuan, Hao Wu, Peter Humphrey, Marcello DiStasio, Mohamed Kahila, Alfonso~Garcia Tan, Heli Qi, Rui Yang, Simeng Han, et~al.
\newblock Teampath: building multimodal pathology experts with reasoning ai copilots.
\newblock \emph{arXiv preprint arXiv:2511.17652}, 2025{\natexlab{d}}.

\bibitem[Liu et~al.(2025{\natexlab{e}})Liu, Chen, Li, Qi, Pang, Du, Lee, and Lin]{liu2025understanding}
Zichen Liu, Changyu Chen, Wenjun Li, Penghui Qi, Tianyu Pang, Chao Du, Wee~Sun Lee, and Min Lin.
\newblock Understanding r1-zero-like training: A critical perspective.
\newblock \emph{arXiv preprint arXiv:2503.20783}, 2025{\natexlab{e}}.

\bibitem[Liu et~al.(2023)Liu, Zhang, Li, Liu, and Yang]{liu2023dynamic}
Zijun Liu, Yanzhe Zhang, Peng Li, Yang Liu, and Diyi Yang.
\newblock Dynamic llm-agent network: An llm-agent collaboration framework with agent team optimization.
\newblock \emph{arXiv preprint arXiv:2310.02170}, 2023.

\bibitem[Lou et~al.(2025)Lou, Yang, Yu, Fu, Han, Huang, and Yu]{lou2025cxragent}
Jinhui Lou, Yan Yang, Zhou Yu, Zhenqi Fu, Weidong Han, Qingming Huang, and Jun Yu.
\newblock Cxragent: Director-orchestrated multi-stage reasoning for chest x-ray interpretation.
\newblock \emph{arXiv preprint arXiv:2510.21324}, 2025.

\bibitem[Lu et~al.(2024)Lu, Chen, Williamson, Chen, Liang, Ding, Jaume, Odintsov, Le, Gerber, et~al.]{lu2024visual}
Ming~Y Lu, Bowen Chen, Drew~FK Williamson, Richard~J Chen, Ivy Liang, Tong Ding, Guillaume Jaume, Igor Odintsov, Long~Phi Le, Georg Gerber, et~al.
\newblock A visual-language foundation model for computational pathology.
\newblock \emph{Nature medicine}, 30\penalty0 (3):\penalty0 863--874, 2024.

\bibitem[Lu et~al.(2026)Lu, Lin, Shi, Tamo, Zhao, Wang, and Wang]{lu2026clinenv}
Yuxing Lu, Yushuhong Lin, Wenqi Shi, J~Ben Tamo, Xukai Zhao, Jinzhuo Wang, and May~Dongmei Wang.
\newblock Clinenv: An interactive multi-stage long horizon ehr environment for agents.
\newblock \emph{arXiv preprint arXiv:2606.02568}, 2026.

\bibitem[Luo et~al.(2025)Luo, Zhang, Yuan, Zhao, Yang, Gu, Wu, Chen, Qiao, Long, et~al.]{luo2025large}
Junyu Luo, Weizhi Zhang, Ye~Yuan, Yusheng Zhao, Junwei Yang, Yiyang Gu, Bohan Wu, Binqi Chen, Ziyue Qiao, Qingqing Long, et~al.
\newblock Large language model agent: A survey on methodology, applications and challenges.
\newblock \emph{arXiv preprint arXiv:2503.21460}, 2025.

\bibitem[Lyu et~al.(2025)Lyu, Liang, Chen, Ding, Yang, Huang, Zhang, He, and Shen]{lyu2025wsi}
Xinheng Lyu, Yuci Liang, Wenting Chen, Meidan Ding, Jiaqi Yang, Guolin Huang, Daokun Zhang, Xiangjian He, and Linlin Shen.
\newblock Wsi-agents: A collaborative multi-agent system for multi-modal whole slide image analysis.
\newblock \emph{arXiv preprint arXiv:2507.14680}, 2025.

\bibitem[Ma et~al.(2024)Ma, He, Li, Han, You, and Wang]{ma2024segment}
Jun Ma, Yuting He, Feifei Li, Lin Han, Chenyu You, and Bo~Wang.
\newblock Segment anything in medical images.
\newblock \emph{Nature Communications}, 15:\penalty0 654, 2024.

\bibitem[Madaan et~al.(2023)Madaan, Tandon, Gupta, Hallinan, Gao, Wiegreffe, Alon, Dziri, Prabhumoye, Yang, et~al.]{madaan2023self}
Aman Madaan, Niket Tandon, Prakhar Gupta, Skyler Hallinan, Luyu Gao, Sarah Wiegreffe, Uri Alon, Nouha Dziri, Shrimai Prabhumoye, Yiming Yang, et~al.
\newblock Self-refine: Iterative refinement with self-feedback.
\newblock \emph{Advances in neural information processing systems}, 36:\penalty0 46534--46594, 2023.

\bibitem[Madavan et~al.(2025)Madavan, Kaimal, Faisal, and Chandrakala]{madavan2025med}
Rakesh~Raj Madavan, Akshat Kaimal, Hashim Faisal, and S~Chandrakala.
\newblock Med-grim: enhanced zero-shot medical vqa using prompt-embedded multimodal graph rag.
\newblock \emph{arXiv preprint arXiv:2508.06496}, 2025.

\bibitem[Mahdavi et~al.(2026)Mahdavi, Khodakaramimaghsoud, Khaloo, Taleshani, Hashemi, Kaleybar, and Manzari]{mahdavi2026med}
Zahra Mahdavi, Zahra Khodakaramimaghsoud, Hooman Khaloo, Sina~Bakhshandeh Taleshani, Erfan Hashemi, Javad~Mirzapour Kaleybar, and Omid~Nejati Manzari.
\newblock Med-vcd: Mitigating hallucination for medical large vision language models through visual contrastive decoding.
\newblock \emph{Computers in Biology and Medicine}, 200:\penalty0 111347, 2026.

\bibitem[Maksudov et~al.(2026)Maksudov, Kurenkov, Curran, and Mileo]{maksudov2026abra}
Bulat Maksudov, Vladislav Kurenkov, Kathleen~M Curran, and Alessandra Mileo.
\newblock Abra: Agent benchmark for radiology applications.
\newblock \emph{arXiv preprint arXiv:2605.11224}, 2026.

\bibitem[Mallinar et~al.(2026)Mallinar, Heydari, Liu, Faranesh, Winslow, Hammerquist, Graef, Speed, Malhotra, Patel, et~al.]{mallinar2026scalable}
Neil Mallinar, A~Ali Heydari, Xin Liu, Anthony~Z Faranesh, Brent Winslow, Nova Hammerquist, Benjamin Graef, Cathy Speed, Mark Malhotra, Shwetak Patel, et~al.
\newblock A scalable framework for evaluating health language models.
\newblock \emph{npj Digital Medicine}, 2026.

\bibitem[Malpure et~al.(2025)Malpure, Kanungo, Loomis, and Reza]{malpure2025securing}
Saurabh Malpure, Rajesh Kanungo, Benjamin~R Loomis, and SM~Salim Reza.
\newblock Securing ai and agentic systems in medical devices: Methods, risks, and defenses.
\newblock In \emph{2025 Third International Conference on Cyber Physical Systems, Power Electronics and Electric Vehicles (ICPEEV)}, pages 1--8. IEEE, 2025.

\bibitem[Mao et~al.(2026)Mao, Xu, Qin, and Gao]{mao2025ct}
Yuren Mao, Wenyi Xu, Yuyang Qin, and Yunjun Gao.
\newblock Ct-agent: a multimodal-llm agent for 3d ct radiology question answering.
\newblock \emph{Science China Information Sciences}, 69\penalty0 (5):\penalty0 150107, 2026.

\bibitem[Masayoshi et~al.(2025)Masayoshi, Hashimoto, Yokoyama, Toda, Uwamino, Fukuda, Namkoong, and Jinzaki]{masayoshi2025ehr}
Kanato Masayoshi, Masahiro Hashimoto, Ryoichi Yokoyama, Naoki Toda, Yoshifumi Uwamino, Shogo Fukuda, Ho~Namkoong, and Masahiro Jinzaki.
\newblock Ehr-mcp: Real-world evaluation of clinical information retrieval by large language models via model context protocol.
\newblock \emph{arXiv preprint arXiv:2509.15957}, 2025.

\bibitem[Mei et~al.(2024)Mei, Zhu, Xu, Hua, Jin, Li, Xu, Ye, Ge, and Zhang]{mei2024aios}
Kai Mei, Xi~Zhu, Wujiang Xu, Wenyue Hua, Mingyu Jin, Zelong Li, Shuyuan Xu, Ruosong Ye, Yingqiang Ge, and Yongfeng Zhang.
\newblock Aios: Llm agent operating system.
\newblock \emph{arXiv preprint arXiv:2403.16971}, 2024.

\bibitem[Mei et~al.(2025)Mei, Yao, Ge, Wang, Bi, Cai, Liu, Li, Li, Zhang, et~al.]{mei2025survey}
Lingrui Mei, Jiayu Yao, Yuyao Ge, Yiwei Wang, Baolong Bi, Yujun Cai, Jiazhi Liu, Mingyu Li, Zhong-Zhi Li, Duzhen Zhang, et~al.
\newblock A survey of context engineering for large language models.
\newblock \emph{arXiv preprint arXiv:2507.13334}, 2025.

\bibitem[Meng et~al.(2026)Meng, Wang, Chen, Wang, Lu, Wu, Gao, Wu, and Hu]{meng2026agent}
Qianyu Meng, Yanan Wang, Liyi Chen, Qimeng Wang, Chengqiang Lu, Wei Wu, Yan Gao, Yi~Wu, and Yao Hu.
\newblock Agent harness for large language model agents: A survey.
\newblock 2026.

\bibitem[Meng et~al.(2025)Meng, Hao, Dai, Feng, Liu, Feng, Wu, Gai, Zhu, Hu, et~al.]{meng2025dentvlm}
Zijie Meng, Jin Hao, Xiwei Dai, Yang Feng, Jiaxiang Liu, Bin Feng, Huikai Wu, Xiaotang Gai, Hengchuan Zhu, Tianxiang Hu, et~al.
\newblock Dentvlm: A multimodal vision-language model for comprehensive dental diagnosis and enhanced clinical practice.
\newblock \emph{arXiv preprint arXiv:2509.23344}, 2025.

\bibitem[Miculicich et~al.(2025)Miculicich, Parmar, Palangi, Dvijotham, Montanari, Pfister, and Le]{miculicich2025veriguard}
Lesly Miculicich, Mihir Parmar, Hamid Palangi, Krishnamurthy~Dj Dvijotham, Mirko Montanari, Tomas Pfister, and Long~T Le.
\newblock Veriguard: Enhancing llm agent safety via verified code generation.
\newblock \emph{arXiv preprint arXiv:2510.05156}, 2025.

\bibitem[Moon and Lee(2022)]{moon2022moma}
Sehwan Moon and Hyunju Lee.
\newblock Moma: a multi-task attention learning algorithm for multi-omics data interpretation and classification.
\newblock \emph{Bioinformatics}, 38\penalty0 (8):\penalty0 2287--2296, 2022.

\bibitem[Muhetaer et~al.(2025)Muhetaer, Yusupu, Yifan, Mutalipu, and Hao]{muhetaer2025medical}
Muretijiang Muhetaer, Ailimulati Yusupu, Wang Yifan, Munire Mutalipu, and Fan Hao.
\newblock Medical qa dialogue datasets in rag systems performance evaluation and chatgpt optimization.
\newblock \emph{Scientific Reports}, 15\penalty0 (1):\penalty0 44467, 2025.

\bibitem[Muti et~al.(2026)Muti, Dulout, and Fu]{muti2026medcase}
Valentina~Bui Muti, Eug{\'e}nie Dulout, and Ziquan Fu.
\newblock Medcase-structured: A text-to-fhir dataset for benchmarking diagnostic reasoning in clinically realistic ehr settings.
\newblock \emph{arXiv preprint arXiv:2605.30295}, 2026.

\bibitem[Naidu et~al.(2025)Naidu, Lakkshmanan, Krishna, Elamathi, Reddy, et~al.]{naidu2025federated}
U~Ganesh Naidu, Ajanthaa Lakkshmanan, JSV~Gopala Krishna, E~Elamathi, T~Sreenivasula Reddy, et~al.
\newblock Federated ai framework for privacy-preserving differential diagnosis across distributed medical networks.
\newblock In \emph{2025 6th International Conference on Inventive Research in Computing Applications (ICIRCA)}, pages 932--940. IEEE, 2025.

\bibitem[Nannini et~al.(2026)Nannini, Smith, Maggini, Panai, Feliciano, Tiulkanov, Maran, Gealy, and Bisconti]{nannini2026ai}
Luca Nannini, Adam~Leon Smith, Michele~Joshua Maggini, Enrico Panai, Sandra Feliciano, Aleksandr Tiulkanov, Elena Maran, James Gealy, and Piercosma Bisconti.
\newblock Ai agents under eu law.
\newblock \emph{arXiv preprint arXiv:2604.04604}, 2026.

\bibitem[Nguyen et~al.(2025)Nguyen, Ho, Ta, Nguyen, Chen, Rav, Dang, Ramchandre, Phung, Liao, et~al.]{nguyen2025localizing}
Dung Nguyen, Minh~Khoi Ho, Huy Ta, Thanh~Tam Nguyen, Qi~Chen, Kumar Rav, Quy~Duong Dang, Satwik Ramchandre, Son~Lam Phung, Zhibin Liao, et~al.
\newblock Localizing before answering: A benchmark for grounded medical visual question answering.
\newblock In \emph{Thirty-Fourth International Joint Conference on Artificial Intelligence (IJCAI-25)}. International Joint Conferences on Artificial Intelligence Organization, 2025.

\bibitem[Obermeyer et~al.(2019)Obermeyer, Powers, Vogeli, and Mullainathan]{obermeyer2019dissecting}
Ziad Obermeyer, Brian Powers, Christine Vogeli, and Sendhil Mullainathan.
\newblock Dissecting racial bias in an algorithm used to manage the health of populations.
\newblock \emph{Science}, 366\penalty0 (6464):\penalty0 447--453, 2019.

\bibitem[Ostmeier et~al.(2024)Ostmeier, Xu, Chen, Varma, Blankemeier, Bluethgen, Md, Moseley, Langlotz, Chaudhari, et~al.]{ostmeier2024green}
Sophie Ostmeier, Justin Xu, Zhihong Chen, Maya Varma, Louis Blankemeier, Christian Bluethgen, Arne Edward~Michalson Md, Michael Moseley, Curtis Langlotz, Akshay~S Chaudhari, et~al.
\newblock Green: Generative radiology report evaluation and error notation.
\newblock In \emph{Findings of the association for computational linguistics: EMNLP 2024}, pages 374--390, 2024.

\bibitem[Packer et~al.(2023)Packer, Fang, Patil, Lin, Wooders, and Gonzalez]{packer2023memgpt}
Charles Packer, Vivian Fang, Shishir\_G Patil, Kevin Lin, Sarah Wooders, and Joseph\_E Gonzalez.
\newblock Memgpt: towards llms as operating systems.
\newblock 2023.

\bibitem[Pal et~al.(2022)Pal, Umapathi, and Sankarasubbu]{pal2022medmcqa}
Ankit Pal, Logesh~Kumar Umapathi, and Malaikannan Sankarasubbu.
\newblock Medmcqa: A large-scale multi-subject multi-choice dataset for medical domain question answering.
\newblock In \emph{Conference on health, inference, and learning}, pages 248--260. PMLR, 2022.

\bibitem[Palaniappan et~al.(2024)Palaniappan, Lin, and Vogel]{palaniappan2024global}
Kavitha Palaniappan, Elaine Yan~Ting Lin, and Silke Vogel.
\newblock Global regulatory frameworks for the use of artificial intelligence (ai) in the healthcare services sector.
\newblock In \emph{Healthcare}, volume~12, page 562. MDPI, 2024.

\bibitem[Pan et~al.(2025{\natexlab{a}})Pan, Liu, Wu, Liu, Zhu, Li, Chen, Ouyang, and Rueckert]{pan2025medvlm}
Jiazhen Pan, Che Liu, Junde Wu, Fenglin Liu, Jiayuan Zhu, Hongwei~Bran Li, Chen Chen, Cheng Ouyang, and Daniel Rueckert.
\newblock Medvlm-r1: Incentivizing medical reasoning capability of vision-language models (vlms) via reinforcement learning.
\newblock In \emph{International Conference on Medical Image Computing and Computer-Assisted Intervention}, pages 337--347. Springer, 2025{\natexlab{a}}.

\bibitem[Pan et~al.(2026)Pan, Zou, Guo, Ni, and Zheng]{pan2026natural}
Linyue Pan, Lexiao Zou, Shuo Guo, Jingchen Ni, and Hai-Tao Zheng.
\newblock Natural-language agent harnesses.
\newblock \emph{arXiv preprint arXiv:2603.25723}, 2026.

\bibitem[Pan et~al.(2025{\natexlab{b}})Pan, Bai, Zou, Zhou, Zhou, Fu, Tham, and Liu]{pan2025eh}
Xiaoyu Pan, Yang Bai, Ke~Zou, Yang Zhou, Jun Zhou, Huazhu Fu, Yih-Chung Tham, and Yong Liu.
\newblock Eh-benchmark: Ophthalmic hallucination benchmark and agent-driven top-down traceable reasoning workflow.
\newblock \emph{Information Fusion}, page 103631, 2025{\natexlab{b}}.

\bibitem[Pandit et~al.(2025)Pandit, Xu, Hong, Wang, Chen, Xu, and Ding]{pandit2025medhallu}
Shrey Pandit, Jiawei Xu, Junyuan Hong, Zhangyang Wang, Tianlong Chen, Kaidi Xu, and Ying Ding.
\newblock Medhallu: A comprehensive benchmark for detecting medical hallucinations in large language models.
\newblock In \emph{Proceedings of the 2025 Conference on Empirical Methods in Natural Language Processing}, pages 2858--2873, 2025.

\bibitem[Patil et~al.(2025)Patil, Mao, Yan, Ji, Suresh, Stoica, and Gonzalez]{patil2025berkeley}
Shishir~G Patil, Huanzhi Mao, Fanjia Yan, Charlie Cheng-Jie Ji, Vishnu Suresh, Ion Stoica, and Joseph~E Gonzalez.
\newblock The berkeley function calling leaderboard (bfcl): From tool use to agentic evaluation of large language models.
\newblock In \emph{Forty-second International Conference on Machine Learning}, 2025.

\bibitem[P{\'e}rez-Garc{\'\i}a et~al.()P{\'e}rez-Garc{\'\i}a, Sharma, Bond-Taylor, Bouzid, Salvatelli, Ilse, et~al.]{perezrad}
F~P{\'e}rez-Garc{\'\i}a, H~Sharma, S~Bond-Taylor, K~Bouzid, V~Salvatelli, M~Ilse, et~al.
\newblock Rad-dino: Exploring scalable medical image encoders beyond text supervision; 2024.

\bibitem[Pesapane et~al.(2026)Pesapane, De~Cecco, Wang, Hauglid, and Sardanelli]{pesapane2026artificial}
Filippo Pesapane, Carlo De~Cecco, Hao Wang, Mathias~K Hauglid, and Francesco Sardanelli.
\newblock Artificial intelligence as medical device in radiology in 2025: the regulatory scenario in the eu, usa, and china.
\newblock \emph{European Radiology}, pages 1--12, 2026.

\bibitem[Qi et~al.(2025)Qi, Liu, Iong, Lai, Sun, Sun, Yang, Yang, Yao, Xu, et~al.]{qi2025webrl}
Zehan Qi, Xiao Liu, Iat~Long Iong, Hanyu Lai, Xueqiao Sun, Jiadai Sun, Xinyue Yang, Yu~Yang, Shuntian Yao, Wei Xu, et~al.
\newblock Webrl: Training llm web agents via self-evolving online curriculum reinforcement learning.
\newblock In \emph{International Conference on Learning Representations}, volume 2025, pages 79791--79821, 2025.

\bibitem[Qian et~al.(2025)Qian, Xie, Wang, Liu, Zhu, Xia, Dang, Du, Chen, Yang, et~al.]{qian2025scaling}
Chen Qian, Zihao Xie, Yifei Wang, Wei Liu, Kunlun Zhu, Hanchen Xia, Yufan Dang, Zhuoyun Du, Weize Chen, Cheng Yang, et~al.
\newblock Scaling large language model-based multi-agent collaboration.
\newblock In \emph{International Conference on Learning Representations}, volume 2025, pages 41488--41505, 2025.

\bibitem[Qian et~al.(2023)Qian, Han, Fung, Qin, Liu, and Ji]{qian2023creator}
Cheng Qian, Chi Han, Yi~Fung, Yujia Qin, Zhiyuan Liu, and Heng Ji.
\newblock Creator: Tool creation for disentangling abstract and concrete reasoning of large language models.
\newblock In \emph{Findings of the Association for Computational Linguistics: EMNLP 2023}, pages 6922--6939, 2023.

\bibitem[Qiao et~al.(2023)Qiao, Li, Zhang, He, Kang, Zhang, Yang, Dong, Zhang, Wang, et~al.]{qiao2023taskweaver}
Bo~Qiao, Liqun Li, Xu~Zhang, Shilin He, Yu~Kang, Chaoyun Zhang, Fangkai Yang, Hang Dong, Jue Zhang, Lu~Wang, et~al.
\newblock Taskweaver: A code-first agent framework.
\newblock \emph{arXiv preprint arXiv:2311.17541}, 2023.

\bibitem[Qiao et~al.(2026)Qiao, Liu, Shen, Wang, Gu, Chu, and Ren]{qiao2026ehr}
Yitong Qiao, Lei Liu, Yue Shen, Jian Wang, Jinjie Gu, Zhixuan Chu, and Kui Ren.
\newblock Ehr-complex: Benchmarking medical agents for complex clinical reasoning.
\newblock \emph{arXiv preprint arXiv:2606.23301}, 2026.

\bibitem[Qin et~al.(2024)Qin, Liang, Ye, Zhu, Yan, Lu, Lin, Cong, Tang, Qian, et~al.]{qin2024toolllm}
Yujia Qin, Shihao Liang, Yining Ye, Kunlun Zhu, Lan Yan, Yaxi Lu, Yankai Lin, Xin Cong, Xiangru Tang, Bill Qian, et~al.
\newblock Toolllm: Facilitating large language models to master 16000+ real-world apis.
\newblock In \emph{International Conference on Learning Representations}, 2024.

\bibitem[Qiu et~al.(2025)Qiu, Wu, Liu, Zhao, Chen, Gu, Peng, Zhang, Wang, and Xie]{qiu2025quantifying}
Pengcheng Qiu, Chaoyi Wu, Shuyu Liu, Weike Zhao, Zhuoxia Chen, Hongfei Gu, Chuanjin Peng, Ya~Zhang, Yanfeng Wang, and Weidi Xie.
\newblock Quantifying the reasoning abilities of llms on real-world clinical cases.
\newblock \emph{arXiv preprint arXiv:2503.04691}, 2025.

\bibitem[Quang et~al.(2025)Quang, Binh$^1$, Nguyen, Le~Thien Phuc~Nguyen, and Bagci]{quang2025gmat}
Ngoc Bui~Lam Quang, Nam Le~Nguyen Binh$^1$, Thanh-Huy Nguyen, Quan~Nguyen$^1$ Le~Thien Phuc~Nguyen, and Ulas Bagci.
\newblock Gmat: Grounded multi-agent clinical.
\newblock In \emph{Emerging LLM/LMM Applications in Medical Imaging: First International Workshop, ELAMI 2025, Held in Conjunction with MICCAI 2025, Daejeon, South Korea, September 27, 2025, Proceedings}, page~1. Springer Nature, 2025.

\bibitem[Radosevich and Halloran(2025)]{radosevich2025mcp}
Brandon Radosevich and John Halloran.
\newblock Mcp safety audit: Llms with the model context protocol allow major security exploits.
\newblock \emph{arXiv preprint arXiv:2504.03767}, 2025.

\bibitem[Rafailov et~al.(2023)Rafailov, Sharma, Mitchell, Manning, Ermon, and Finn]{rafailov2023direct}
Rafael Rafailov, Archit Sharma, Eric Mitchell, Christopher~D Manning, Stefano Ermon, and Chelsea Finn.
\newblock Direct preference optimization: Your language model is secretly a reward model.
\newblock \emph{Advances in neural information processing systems}, 36:\penalty0 53728--53741, 2023.

\bibitem[Rempe et~al.(2025)Rempe, Heine, Seibold, H{\"o}rst, and Kleesiek]{rempe2025identification}
Moritz Rempe, Lukas Heine, Constantin Seibold, Fabian H{\"o}rst, and Jens Kleesiek.
\newblock De-identification of medical imaging data: a comprehensive tool for ensuring patient privacy.
\newblock \emph{European radiology}, pages 1--10, 2025.

\bibitem[Roschewitz et~al.(2026)Roschewitz, Styppa, Tao, Sohn, Delbrouck, Gundersen, Deperrois, Bluethgen, Vogt, Menze, et~al.]{roschewitz2026radagent}
M{\'e}lanie Roschewitz, Kenneth Styppa, Yitian Tao, Jiwoong Sohn, Jean-Benoit Delbrouck, Benjamin Gundersen, Nicolas Deperrois, Christian Bluethgen, Julia Vogt, Bjoern Menze, et~al.
\newblock Radagent: A tool-using ai agent for stepwise interpretation of chest computed tomography.
\newblock \emph{arXiv preprint arXiv:2604.15231}, 2026.

\bibitem[Rose et~al.(2025)Rose, Hung, Lepri, Alqassem, Gashteovski, and Lawrence]{rose2025meddxagent}
Daniel~Philip Rose, Chia-Chien Hung, Marco Lepri, Israa Alqassem, Kiril Gashteovski, and Carolin Lawrence.
\newblock Meddxagent: A unified modular agent framework for explainable automatic differential diagnosis.
\newblock In \emph{Proceedings of the 63rd Annual Meeting of the Association for Computational Linguistics (Volume 1: Long Papers)}, pages 13803--13826, 2025.

\bibitem[Russell(2010)]{russell2010artificial}
Stuart~J Russell.
\newblock \emph{Artificial intelligence a modern approach}.
\newblock Pearson Education, Inc., 2010.

\bibitem[Saab et~al.(2024)Saab, Tu, Weng, Tanno, Stutz, Wulczyn, Zhang, Strother, Park, Vedadi, et~al.]{saab2024capabilities}
Khaled Saab, Tao Tu, Wei-Hung Weng, Ryutaro Tanno, David Stutz, Ellery Wulczyn, Fan Zhang, Tim Strother, Chunjong Park, Elahe Vedadi, et~al.
\newblock Capabilities of gemini models in medicine.
\newblock \emph{arXiv preprint arXiv:2404.18416}, 2024.

\bibitem[Sampa et~al.(2026)Sampa, Abdul~Aziz, Rahman, Ab.~Aziz, Besar, and Ghazali]{sampa2026reinforcement}
Masuda~Begum Sampa, Nor~Hidayati Abdul~Aziz, Md~Siddikur Rahman, Nor~Azlina Ab.~Aziz, Rosli Besar, and Anith~Khairunnisa Ghazali.
\newblock Reinforcement learning for medical image analysis: a systematic review of algorithms, engineering challenges, and clinical deployment.
\newblock \emph{Computer Assisted Surgery}, 31\penalty0 (1):\penalty0 2597553, 2026.

\bibitem[Sardana et~al.(2023)Sardana, Portes, Doubov, and Frankle]{sardana2023beyond}
Nikhil Sardana, Jacob Portes, Sasha Doubov, and Jonathan Frankle.
\newblock Beyond chinchilla-optimal: Accounting for inference in language model scaling laws.
\newblock \emph{arXiv preprint arXiv:2401.00448}, 2023.

\bibitem[Schick et~al.(2023)Schick, Dwivedi-Yu, Dess{\`\i}, Raileanu, Lomeli, Hambro, Zettlemoyer, Cancedda, and Scialom]{schick2023toolformer}
Timo Schick, Jane Dwivedi-Yu, Roberto Dess{\`\i}, Roberta Raileanu, Maria Lomeli, Eric Hambro, Luke Zettlemoyer, Nicola Cancedda, and Thomas Scialom.
\newblock Toolformer: Language models can teach themselves to use tools.
\newblock \emph{Advances in neural information processing systems}, 36:\penalty0 68539--68551, 2023.

\bibitem[Schmidgall et~al.(2024)Schmidgall, Ziaei, Harris, Reis, Jopling, and Moor]{schmidgall2024agentclinic}
Samuel Schmidgall, Rojin Ziaei, Carl Harris, Eduardo Reis, Jeffrey Jopling, and Michael Moor.
\newblock Agentclinic: a multimodal agent benchmark to evaluate ai in simulated clinical environments.
\newblock \emph{arXiv preprint arXiv:2405.07960}, 2024.

\bibitem[Schmidhuber(2007)]{schmidhuber2007godel}
J{\"u}rgen Schmidhuber.
\newblock G{\"o}del machines: Fully self-referential optimal universal self-improvers.
\newblock In \emph{Artificial general intelligence}, pages 199--226. Springer, 2007.

\bibitem[Schulman et~al.(2017)Schulman, Wolski, Dhariwal, Radford, and Klimov]{schulman2017proximal}
John Schulman, Filip Wolski, Prafulla Dhariwal, Alec Radford, and Oleg Klimov.
\newblock Proximal policy optimization algorithms, 2017.

\bibitem[Sellergren et~al.(2025)Sellergren, Kazemzadeh, Jaroensri, Kiraly, Traverse, Kohlberger, Xu, Jamil, Hughes, Lau, et~al.]{sellergren2025medgemma}
Andrew Sellergren, Sahar Kazemzadeh, Tiam Jaroensri, Atilla Kiraly, Madeleine Traverse, Timo Kohlberger, Shawn Xu, Fayaz Jamil, C{\'\i}an Hughes, Charles Lau, et~al.
\newblock Medgemma technical report.
\newblock \emph{arXiv preprint arXiv:2507.05201}, 2025.

\bibitem[Seyyed-Kalantari et~al.(2021)Seyyed-Kalantari, Zhang, McDermott, Chen, and Ghassemi]{seyyed2021underdiagnosis}
Laleh Seyyed-Kalantari, Haoran Zhang, Matthew~BA McDermott, Irene~Y Chen, and Marzyeh Ghassemi.
\newblock Underdiagnosis bias of artificial intelligence algorithms applied to chest radiographs in under-served patient populations.
\newblock \emph{Nature medicine}, 27\penalty0 (12):\penalty0 2176--2182, 2021.

\bibitem[Shao et~al.(2024)Shao, Wang, Zhu, Xu, Song, Bi, Zhang, Zhang, Li, Wu, et~al.]{shao2024deepseekmath}
Zhihong Shao, Peiyi Wang, Qihao Zhu, Runxin Xu, Junxiao Song, Xiao Bi, Haowei Zhang, Mingchuan Zhang, YK~Li, Yang Wu, et~al.
\newblock Deepseekmath: Pushing the limits of mathematical reasoning in open language models.
\newblock \emph{arXiv preprint arXiv:2402.03300}, 2024.

\bibitem[Shen et~al.(2025)Shen, Liu, Li, Fang, Ma, Liao, Shen, Zhang, Zhao, Zhang, et~al.]{shen2025vlm}
Haozhan Shen, Peng Liu, Jingcheng Li, Chunxin Fang, Yibo Ma, Jiajia Liao, Qiaoli Shen, Zilun Zhang, Kangjia Zhao, Qianqian Zhang, et~al.
\newblock Vlm-r1: A stable and generalizable r1-style large vision-language model.
\newblock \emph{arXiv preprint arXiv:2504.07615}, 2025.

\bibitem[Shen et~al.(2026{\natexlab{a}})Shen, Hu, Liu, Wu, Zhu, Shen, Xu, Jin, Wiestler, Rueckert, et~al.]{shen2026medopenclaw}
Weixiang Shen, Yanzhu Hu, Che Liu, Junde Wu, Jiayuan Zhu, Chengzhi Shen, Min Xu, Yueming Jin, Benedikt Wiestler, Daniel Rueckert, et~al.
\newblock Medopenclaw: Auditable medical imaging agents reasoning over uncurated full studies.
\newblock \emph{arXiv preprint arXiv:2603.24649}, 2026{\natexlab{a}}.

\bibitem[Shen et~al.(2026{\natexlab{b}})Shen, Jian, Li, Liu, Moll, Hu, Rueckert, Li, and Pan]{shen2026evo}
Weixiang Shen, Bailiang Jian, Jun Li, Che Liu, Johannes Moll, Xiaobin Hu, Daniel Rueckert, Hongwei~Bran Li, and Jiazhen Pan.
\newblock Evo-medagent: Beyond one-shot diagnosis with agents that remember, reflect, and improve.
\newblock \emph{arXiv preprint arXiv:2604.14475}, 2026{\natexlab{b}}.

\bibitem[Shen et~al.(2026{\natexlab{c}})Shen, Shen, Hu, Liu, Wu, Zhu, Han, Li, Wu, Xu, Xu, Jin, Wiestler, Rueckert, and Pan]{shen2026medopenclawmedflowbenchauditingmedical}
Weixiang Shen, Chengzhi Shen, Yanzhu Hu, Che Liu, Junde Wu, Jiayuan Zhu, Xiao Han, Zongyue Li, Jingpei Wu, Min Xu, Daguang Xu, Yueming Jin, Benedikt Wiestler, Daniel Rueckert, and Jiazhen Pan.
\newblock Medopenclaw and medflowbench: Auditing medical agents in full-study workflows, 2026{\natexlab{c}}.
\newblock URL \url{https://arxiv.org/abs/2603.24649}.

\bibitem[Shen et~al.(2023)Shen, Song, Tan, Li, Lu, and Zhuang]{shen2023hugginggpt}
Yongliang Shen, Kaitao Song, Xu~Tan, Dongsheng Li, Weiming Lu, and Yueting Zhuang.
\newblock Hugginggpt: Solving ai tasks with chatgpt and its friends in hugging face.
\newblock \emph{Advances in Neural Information Processing Systems}, 36:\penalty0 38154--38180, 2023.

\bibitem[Shi et~al.(2026)Shi, Chen, Yan, Zhang, Xu, Yang, Chen, Huang, Liu, Wu, Xie, Gao, Wu, Lin, Jin, Gong, Tham, Zhang, Dong, Zhang, Yam, Jin, Ding, Zou, Zheng, Ge, and He]{shi2026eyeagentagenticaimultimodal}
Danli Shi, Xiaolan Chen, Bingjie Yan, Weiyi Zhang, Pusheng Xu, Jiancheng Yang, Ruoyu Chen, Siyu Huang, Bowen Liu, Xinyuan Wu, Meng Xie, Ziyu Gao, Yue Wu, Senlin Lin, Kai Jin, Xia Gong, Yih~Chung Tham, Xiujuan Zhang, Li~Dong, Yuzhou Zhang, Jason Yam, Guangming Jin, Xiaohu Ding, Haidong Zou, Yalin Zheng, Zongyuan Ge, and Mingguang He.
\newblock Eyeagent: An agentic ai system for multimodal clinical decision support in ophthalmology, 2026.
\newblock URL \url{https://arxiv.org/abs/2511.09394}.

\bibitem[Shi et~al.(2024)Shi, Xu, Zhuang, Yu, Zhang, Wu, Zhu, Ho, Yang, and Wang]{shi2024ehragent}
Wenqi Shi, Ran Xu, Yuchen Zhuang, Yue Yu, Jieyu Zhang, Hang Wu, Yuanda Zhu, Joyce~C Ho, Carl Yang, and May~Dongmei Wang.
\newblock Ehragent: Code empowers large language models for few-shot complex tabular reasoning on electronic health records.
\newblock In \emph{Proceedings of the 2024 Conference on Empirical Methods in Natural Language Processing}, pages 22315--22339, 2024.

\bibitem[Shimgekar et~al.(2025)Shimgekar, Vassef, Goyal, Saha, Zonooz, and Kumar]{shimgekar2025agentic}
Soorya~Ram Shimgekar, Shayan Vassef, Abhay Goyal, Koustuv Saha, Pi~Zonooz, and Navin Kumar.
\newblock Agentic ai framework for end-to-end medical data inference.
\newblock In \emph{2025 IEEE International Conference on Bioinformatics and Biomedicine (BIBM)}, pages 7807--7810. IEEE, 2025.

\bibitem[Shinn et~al.(2023)Shinn, Cassano, Gopinath, Narasimhan, and Yao]{shinn2023reflexion}
Noah Shinn, Federico Cassano, Ashwin Gopinath, Karthik Narasimhan, and Shunyu Yao.
\newblock Reflexion: Language agents with verbal reinforcement learning.
\newblock \emph{Advances in neural information processing systems}, 36:\penalty0 8634--8652, 2023.

\bibitem[Shumailov et~al.(2024)Shumailov, Shumaylov, Zhao, Papernot, Anderson, and Gal]{shumailov2024ai}
Ilia Shumailov, Zakhar Shumaylov, Yiren Zhao, Nicolas Papernot, Ross Anderson, and Yarin Gal.
\newblock Ai models collapse when trained on recursively generated data.
\newblock \emph{Nature}, 631\penalty0 (8022):\penalty0 755--759, 2024.

\bibitem[Silveira et~al.(2025)Silveira, da~Rosa~Righi, and Da~Costa]{silveira2025multi}
Andre~Lehdermann Silveira, Rodrigo da~Rosa~Righi, and Cristiano~Andre Da~Costa.
\newblock Multi-agent systems for clinical decision support: A systematic review.
\newblock \emph{Applied Soft Computing}, page 114447, 2025.

\bibitem[Singh et~al.(2023)Singh, Co-Reyes, Agarwal, Anand, Patil, Garcia, Liu, Harrison, Lee, Xu, et~al.]{singh2023beyond}
Avi Singh, John~D Co-Reyes, Rishabh Agarwal, Ankesh Anand, Piyush Patil, Xavier Garcia, Peter~J Liu, James Harrison, Jaehoon Lee, Kelvin Xu, et~al.
\newblock Beyond human data: Scaling self-training for problem-solving with language models.
\newblock \emph{arXiv preprint arXiv:2312.06585}, 2023.

\bibitem[Smuha(2025)]{smuha2025regulation}
Nathalie~A Smuha.
\newblock Regulation 2024/1689 of the eur. parl. \& council of june 13, 2024 (eu artificial intelligence act).
\newblock \emph{International Legal Materials}, 64\penalty0 (5):\penalty0 1234--1381, 2025.

\bibitem[Snell et~al.(2024)Snell, Lee, Xu, and Kumar]{snell2024scaling}
Charlie Snell, Jaehoon Lee, Kelvin Xu, and Aviral Kumar.
\newblock Scaling llm test-time compute optimally can be more effective than scaling model parameters, 2024.

\bibitem[Song and Zheng(2026)]{song2026survey}
Mingyang Song and Mao Zheng.
\newblock A survey of on-policy distillation for large language models.
\newblock \emph{arXiv preprint arXiv:2604.00626}, 2026.

\bibitem[Song et~al.(2026)Song, Chang, Dong, Zhu, Wen, and Dou]{song2026envscaler}
Xiaoshuai Song, Haofei Chang, Guanting Dong, Yutao Zhu, Ji-Rong Wen, and Zhicheng Dou.
\newblock Envscaler: Scaling tool-interactive environments for llm agent via programmatic synthesis.
\newblock \emph{arXiv preprint arXiv:2601.05808}, 2026.

\bibitem[Song et~al.(2024)Song, Yin, Yue, Huang, Li, and Lin]{song2024trial}
Yifan Song, Da~Yin, Xiang Yue, Jie Huang, Sujian Li, and Bill~Yuchen Lin.
\newblock Trial and error: Exploration-based trajectory optimization of llm agents.
\newblock In \emph{Proceedings of the 62nd Annual Meeting of the Association for Computational Linguistics (Volume 1: Long Papers)}, pages 7584--7600, 2024.

\bibitem[Su et~al.(2025)Su, Choudhuri, Gao, Planche, Nguyen, Zheng, Shen, Innanje, Chen, Elhamifar, et~al.]{su2025medgrpo}
Yuhao Su, Anwesa Choudhuri, Zhongpai Gao, Benjamin Planche, Van~Nguyen Nguyen, Meng Zheng, Yuhan Shen, Arun Innanje, Terrence Chen, Ehsan Elhamifar, et~al.
\newblock Medgrpo: Multi-task reinforcement learning for heterogeneous medical video understanding.
\newblock \emph{arXiv preprint arXiv:2512.06581}, 2025.

\bibitem[Sun et~al.(2026{\natexlab{a}})Sun, Yu, Tan, Chen, Cheng, Joshi, and Xiong]{sun2026linking}
Liwen Sun, Xiang Yu, Ming Tan, Zhuohao Chen, Anqi Cheng, Ashutosh Joshi, and Chenyan Xiong.
\newblock Linking knowledge to care: Knowledge graph-augmented medical follow-up question generation.
\newblock In \emph{Findings of the Association for Computational Linguistics: EACL 2026}, pages 846--853, 2026{\natexlab{a}}.

\bibitem[Sun et~al.(2026{\natexlab{b}})Sun, Si, Zhu, Zhang, Shui, Ding, Lin, and Yang]{sun2026cpathagent}
Yuxuan Sun, Yixuan Si, Chenglu Zhu, Kai Zhang, Zhongyi Shui, Bowen Ding, Tao Lin, and Lin Yang.
\newblock Cpathagent: An agent-based foundation model for interpretable high-resolution pathology image analysis mimicking pathologists' diagnostic logic.
\newblock \emph{Advances in Neural Information Processing Systems}, 38:\penalty0 101673--101731, 2026{\natexlab{b}}.

\bibitem[Sun et~al.(2026{\natexlab{c}})Sun, Jagtiani, Yim, Xia, Gunn, Yetisgen, and Abacha]{sun2026radar}
Zhaoyi Sun, Minal Jagtiani, Wen-wai Yim, Fei Xia, Martin Gunn, Meliha Yetisgen, and Asma~Ben Abacha.
\newblock Radar: A multimodal benchmark for 3d image-based radiology report review.
\newblock \emph{arXiv preprint arXiv:2603.06681}, 2026{\natexlab{c}}.

\bibitem[Sur{\'\i}s et~al.(2023)Sur{\'\i}s, Menon, and Vondrick]{suris2023vipergpt}
D{\'\i}dac Sur{\'\i}s, Sachit Menon, and Carl Vondrick.
\newblock Vipergpt: Visual inference via python execution for reasoning.
\newblock In \emph{Proceedings of the IEEE/CVF international conference on computer vision}, pages 11888--11898, 2023.

\bibitem[Sviridov et~al.(2025)Sviridov, Miftakhova, Tereshchenko, Zubkova, Blinov, and Savchenko]{Sviridov_2025}
Ivan Sviridov, Amina Miftakhova, Artemiy Tereshchenko, Galina Zubkova, Pavel Blinov, and Andrey Savchenko.
\newblock 3mdbench: Medical multimodal multi-agent dialogue benchmark.
\newblock In \emph{Proceedings of the 2025 Conference on Empirical Methods in Natural Language Processing}, page 26625–26665. Association for Computational Linguistics, 2025.
\newblock \doi{10.18653/v1/2025.emnlp-main.1353}.
\newblock URL \url{http://dx.doi.org/10.18653/v1/2025.emnlp-main.1353}.

\bibitem[Tan and Tian(2026)]{tan2026medresearchbench}
Shuping Tan and Zhanxiao Tian.
\newblock Medresearchbench: A multi-domain benchmark for evaluating ai research agents on clinical medical research.
\newblock \emph{medRxiv}, pages 2026--03, 2026.

\bibitem[Tang et~al.(2024)Tang, Zou, Zhang, Li, Zhao, Zhang, Cohan, and Gerstein]{tang2024medagents}
Xiangru Tang, Anni Zou, Zhuosheng Zhang, Ziming Li, Yilun Zhao, Xingyao Zhang, Arman Cohan, and Mark Gerstein.
\newblock Medagents: Large language models as collaborators for zero-shot medical reasoning.
\newblock In \emph{Findings of the Association for Computational Linguistics: ACL 2024}, pages 599--621, 2024.

\bibitem[Tang et~al.(2025)Tang, Shao, Sohn, Chen, Zhang, Xiang, Wu, Zhao, Wu, Shi, et~al.]{tang2025medagentsbench}
Xiangru Tang, Daniel Shao, Jiwoong Sohn, Jiapeng Chen, Jiayi Zhang, Jinyu Xiang, Fang Wu, Yilun Zhao, Chenglin Wu, Wenqi Shi, et~al.
\newblock Medagentsbench: Benchmarking thinking models and agent frameworks for complex medical reasoning.
\newblock \emph{arXiv preprint arXiv:2503.07459}, 2025.

\bibitem[Touvron et~al.(2023)Touvron, Martin, Stone, Albert, Almahairi, Babaei, Bashlykov, Batra, Bhargava, Bhosale, et~al.]{touvron2023llama}
Hugo Touvron, Louis Martin, Kevin Stone, Peter Albert, Amjad Almahairi, Yasmine Babaei, Nikolay Bashlykov, Soumya Batra, Prajjwal Bhargava, Shruti Bhosale, et~al.
\newblock Llama 2: Open foundation and fine-tuned chat models, 2023.

\bibitem[Tu et~al.(2024)Tu, Palepu, Schaekermann, Saab, Freyberg, Tanno, Wang, Li, Amin, Tomasev, et~al.]{tu2024towards}
Tao Tu, Anil Palepu, Mike Schaekermann, Khaled Saab, Jan Freyberg, Ryutaro Tanno, Amy Wang, Brenna Li, Mohamed Amin, Nenad Tomasev, et~al.
\newblock Towards conversational diagnostic ai.
\newblock \emph{arXiv preprint arXiv:2401.05654}, 2024.

\bibitem[Tzanis and Klontzas(2025)]{tzanis2025maistro}
Eleftherios Tzanis and Michail~E Klontzas.
\newblock maistro: An open-source multi-agent system for automated end-to-end development of radiomics and deep learning models for medical imaging.
\newblock \emph{European Journal of Radiology Artificial Intelligence}, page 100044, 2025.

\bibitem[Wan et~al.(2023)Wan, Wang, Liu, Alam, Zheng, Liu, Qu, Yan, Zhu, Zhang, et~al.]{wan2023efficient}
Zhongwei Wan, Xin Wang, Che Liu, Samiul Alam, Yu~Zheng, Jiachen Liu, Zhongnan Qu, Shen Yan, Yi~Zhu, Quanlu Zhang, et~al.
\newblock Efficient large language models: A survey.
\newblock \emph{arXiv preprint arXiv:2312.03863}, 2023.

\bibitem[Wang et~al.(2023{\natexlab{a}})Wang, Xie, Jiang, Mandlekar, Xiao, Zhu, Fan, and Anandkumar]{wang2023voyager}
Guanzhi Wang, Yuqi Xie, Yunfan Jiang, Ajay Mandlekar, Chaowei Xiao, Yuke Zhu, Linxi Fan, and Anima Anandkumar.
\newblock Voyager: An open-ended embodied agent with large language models.
\newblock \emph{arXiv preprint arXiv:2305.16291}, 2023{\natexlab{a}}.

\bibitem[Wang et~al.(2023{\natexlab{b}})Wang, Xu, Lan, Hu, Lan, Lee, and Lim]{wang2023plan}
Lei Wang, Wanyu Xu, Yihuai Lan, Zhiqiang Hu, Yunshi Lan, Roy Ka-Wei Lee, and Ee-Peng Lim.
\newblock Plan-and-solve prompting: Improving zero-shot chain-of-thought reasoning by large language models.
\newblock In \emph{Proceedings of the 61st annual meeting of the association for computational linguistics (volume 1: long papers)}, pages 2609--2634, 2023{\natexlab{b}}.

\bibitem[Wang et~al.(2024{\natexlab{a}})Wang, Ma, Feng, Zhang, Yang, Zhang, Chen, Tang, Chen, Lin, et~al.]{wang2024survey}
Lei Wang, Chen Ma, Xueyang Feng, Zeyu Zhang, Hao Yang, Jingsen Zhang, Zhiyuan Chen, Jiakai Tang, Xu~Chen, Yankai Lin, et~al.
\newblock A survey on large language model based autonomous agents.
\newblock \emph{Frontiers of Computer Science}, 18\penalty0 (6):\penalty0 186345, 2024{\natexlab{a}}.

\bibitem[Wang et~al.(2025{\natexlab{a}})Wang, Ye, Naseem, and Kim]{mrgagents2025}
Pengyu Wang, Shuchang Ye, Usman Naseem, and Jinman Kim.
\newblock Mrgagents: A multi-agent framework for improved medical report generation with med-lvlms.
\newblock In \emph{2025 International Conference on Digital Image Computing: Techniques and Applications (DICTA)}, pages 1--7. IEEE, 2025{\natexlab{a}}.

\bibitem[Wang et~al.(2025{\natexlab{b}})Wang, Ye, Naseem, and Kim]{wang2025mrgr1}
Pengyu Wang, Shuchang Ye, Usman Naseem, and Jinman Kim.
\newblock Mrg-r1: Reinforcement learning for clinically aligned medical report generation.
\newblock \emph{arXiv preprint arXiv:2512.16145}, 2025{\natexlab{b}}.

\bibitem[Wang et~al.(2025{\natexlab{c}})Wang, Luo, Tu, Chou, and Wu]{wang2025systematic}
Ting-Wei Wang, Wei-Ting Luo, Yu-Kang Tu, Yu-Bai Chou, and Yu-Te Wu.
\newblock Systematic review and meta-analysis of regulator-approved deep learning systems for fundus diabetic retinopathy detections.
\newblock \emph{npj Digital Medicine}, 2025{\natexlab{c}}.

\bibitem[Wang et~al.(2024{\natexlab{b}})Wang, Chen, Yuan, Zhang, Li, Peng, and Ji]{wang2024executable}
Xingyao Wang, Yangyi Chen, Lifan Yuan, Yizhe Zhang, Yunzhu Li, Hao Peng, and Heng Ji.
\newblock Executable code actions elicit better llm agents.
\newblock In \emph{Forty-first International Conference on Machine Learning}, 2024{\natexlab{b}}.

\bibitem[Wang et~al.(2025{\natexlab{d}})Wang, Li, Song, Xu, Tang, Zhuge, Pan, Song, Li, Singh, et~al.]{wang2025openhands}
Xingyao Wang, Boxuan Li, Yufan Song, Frank~F Xu, Xiangru Tang, Mingchen Zhuge, Jiayi Pan, Yueqi Song, Bowen Li, Jaskirat Singh, et~al.
\newblock Openhands: An open platform for ai software developers as generalist agents.
\newblock In \emph{International Conference on Learning Representations}, volume 2025, pages 65882--65919, 2025{\natexlab{d}}.

\bibitem[Wang et~al.(2023{\natexlab{c}})Wang, Kordi, Mishra, Liu, Smith, Khashabi, and Hajishirzi]{wang2023self}
Yizhong Wang, Yeganeh Kordi, Swaroop Mishra, Alisa Liu, Noah~A Smith, Daniel Khashabi, and Hannaneh Hajishirzi.
\newblock Self-instruct: Aligning language models with self-generated instructions.
\newblock In \emph{Proceedings of the 61st annual meeting of the association for computational linguistics (volume 1: long papers)}, pages 13484--13508, 2023{\natexlab{c}}.

\bibitem[Wang et~al.(2024{\natexlab{c}})Wang, Fried, and Neubig]{wang2024trove}
Zhiruo Wang, Daniel Fried, and Graham Neubig.
\newblock Trove: Inducing verifiable and efficient toolboxes for solving programmatic tasks.
\newblock \emph{arXiv preprint arXiv:2401.12869}, 2024{\natexlab{c}}.

\bibitem[Wang et~al.(2026)Wang, Wang, Feng, Yang, Wang, Zhang, Lin, Ji, and Yang]{wang2026deepmed}
Zihan Wang, Hao Wang, Shi Feng, Xiaocui Yang, Daling Wang, Yiqun Zhang, Jinghao Lin, Xiaozhong Ji, and Haihua Yang.
\newblock Deepmed: Building a medical deepresearch agent via multi-hop med-search data and turn-controlled agentic training \& inference.
\newblock In \emph{Findings of the Association for Computational Linguistics: ACL 2026}, pages 18160--18178, 2026.

\bibitem[Wang et~al.(2025{\natexlab{e}})Wang, Zhu, Zhao, Zheng, Sui, Wang, Tang, Wang, Harrison, Pan, et~al.]{wang2025colacare}
Zixiang Wang, Yinghao Zhu, Huiya Zhao, Xiaochen Zheng, Dehao Sui, Tianlong Wang, Wen Tang, Yasha Wang, Ewen Harrison, Chengwei Pan, et~al.
\newblock Colacare: Enhancing electronic health record modeling through large language model-driven multi-agent collaboration.
\newblock In \emph{Proceedings of the ACM on Web Conference 2025}, pages 2250--2261, 2025{\natexlab{e}}.

\bibitem[Wang et~al.(2025{\natexlab{f}})Wang, Wu, Cai, Low, Yang, Li, and Jin]{wang2025medagent}
Ziyue Wang, Junde Wu, Linghan Cai, Chang~Han Low, Xihong Yang, Qiaxuan Li, and Yueming Jin.
\newblock Medagent-pro: Towards evidence-based multi-modal medical diagnosis via reasoning agentic workflow.
\newblock \emph{arXiv preprint arXiv:2503.18968}, 2025{\natexlab{f}}.

\bibitem[Wang et~al.(2024{\natexlab{d}})Wang, Mao, Fried, and Neubig]{wang2024agent}
Zora~Zhiruo Wang, Jiayuan Mao, Daniel Fried, and Graham Neubig.
\newblock Agent workflow memory.
\newblock \emph{arXiv preprint arXiv:2409.07429}, 2024{\natexlab{d}}.

\bibitem[Wei et~al.(2024)Wei, Qiu, Yu, and Yuan]{wei2024medco}
Hao Wei, Jianing Qiu, Haibao Yu, and Wu~Yuan.
\newblock Medco: Medical education copilots based on a multi-agent framework.
\newblock In \emph{European Conference on Computer Vision}, pages 119--135. Springer, 2024.

\bibitem[Wei et~al.(2022)Wei, Wang, Schuurmans, Bosma, Xia, Chi, Le, Zhou, et~al.]{wei2022chain}
Jason Wei, Xuezhi Wang, Dale Schuurmans, Maarten Bosma, Fei Xia, Ed~Chi, Quoc~V Le, Denny Zhou, et~al.
\newblock Chain-of-thought prompting elicits reasoning in large language models.
\newblock \emph{Advances in neural information processing systems}, 35:\penalty0 24824--24837, 2022.

\bibitem[Wei et~al.(2026)Wei, Duchenne, Copet, Carbonneaux, Zhang, Fried, Synnaeve, Singh, and Wang]{wei2026swe}
Yuxiang Wei, Olivier Duchenne, Jade Copet, Quentin Carbonneaux, Lingming Zhang, Daniel Fried, Gabriel Synnaeve, Rishabh Singh, and Sida Wang.
\newblock Swe-rl: Advancing llm reasoning via reinforcement learning on open software evolution.
\newblock \emph{Advances in Neural Information Processing Systems}, 38:\penalty0 78500--78525, 2026.

\bibitem[Weishaupt et~al.(2025)Weishaupt, Chen, Williamson, Chen, Jaume, Ding, Chen, Vaidya, Le, Lu, et~al.]{weishaupt2025evidence}
Luca~L Weishaupt, Chengkuan Chen, Drew~FK Williamson, Richard~J Chen, Guillaume Jaume, Tong Ding, Bowen Chen, Anurag Vaidya, Long~Phi Le, Ming~Y Lu, et~al.
\newblock Evidence-based diagnostic reasoning with multi-agent copilot for human pathology.
\newblock \emph{arXiv preprint arXiv:2506.20964}, 2025.

\bibitem[Wen et~al.(2025)Wen, Liu, Zheng, Ye, Wu, Wang, Xu, Liang, Li, Miao, et~al.]{wen2025reinforcement}
Xumeng Wen, Zihan Liu, Shun Zheng, Shengyu Ye, Zhirong Wu, Yang Wang, Zhijian Xu, Xiao Liang, Junjie Li, Ziming Miao, et~al.
\newblock Reinforcement learning with verifiable rewards implicitly incentivizes correct reasoning in base llms.
\newblock \emph{arXiv preprint arXiv:2506.14245}, 2025.

\bibitem[Wu et~al.(2025)Wu, Deng, Li, Liu, Mi, Peng, Xu, Liu, Cho, Choi, et~al.]{wu2025medreason}
Juncheng Wu, Wenlong Deng, Xingxuan Li, Sheng Liu, Taomian Mi, Yifan Peng, Ziyang Xu, Yi~Liu, Hyunjin Cho, Chang-In Choi, et~al.
\newblock Medreason: Eliciting factual medical reasoning steps in llms via knowledge graphs.
\newblock \emph{arXiv preprint arXiv:2504.00993}, 2025.

\bibitem[Wu et~al.(2026)Wu, Zhang, Wang, Tu, Chen, Wang, Xie, and Zhou]{wu2026clinseekagent}
Juncheng Wu, Letian Zhang, Yuhan Wang, Haoqin Tu, Hardy Chen, Zijun Wang, Cihang Xie, and Yuyin Zhou.
\newblock Clinseekagent: Automating multimodal evidence seeking for agentic clinical reasoning.
\newblock \emph{arXiv preprint arXiv:2605.20176}, 2026.

\bibitem[Wu et~al.(2023)Wu, Bansal, Zhang, Wu, Zhang, Zhu, Li, Jiang, Zhang, and Wang]{wu2023autogen}
Qingyun Wu, Gagan Bansal, Jieyu Zhang, Yiran Wu, Shaokun Zhang, Erkang Zhu, Beibin Li, Li~Jiang, Xiaoyun Zhang, and Chi Wang.
\newblock Autogen: Enabling next-gen llm applications via multi-agent conversation framework.
\newblock \emph{arXiv preprint arXiv:2308.08155}, 3\penalty0 (4), 2023.

\bibitem[Xi et~al.()Xi, Ding, Chen, Hong, Guo, Wang, Yang, Liao, Guo, He, et~al.]{xi2406agentgym}
Z~Xi, Y~Ding, W~Chen, B~Hong, H~Guo, J~Wang, D~Yang, C~Liao, X~Guo, W~He, et~al.
\newblock Agentgym: Evolving large language model-based agents across diverse environments. arxiv 2024.
\newblock \emph{arXiv preprint arXiv:2406.04151}.

\bibitem[Xi et~al.(2025)Xi, Chen, Guo, He, Ding, Hong, Zhang, Wang, Jin, Zhou, et~al.]{xi2025rise}
Zhiheng Xi, Wenxiang Chen, Xin Guo, Wei He, Yiwen Ding, Boyang Hong, Ming Zhang, Junzhe Wang, Senjie Jin, Enyu Zhou, et~al.
\newblock The rise and potential of large language model based agents: A survey.
\newblock \emph{Science China Information Sciences}, 68\penalty0 (2):\penalty0 121101, 2025.

\bibitem[Xia et~al.(2024{\natexlab{a}})Xia, Chen, Tian, Gong, Hou, Xu, Wu, Fan, Zhou, Zhu, Zheng, Wang, Wang, Zhang, Bansal, Niethammer, Huang, Zhu, Li, Sun, Ge, Li, Zou, and Yao]{xia2024carescomprehensivebenchmarktrustworthiness}
Peng Xia, Ze~Chen, Juanxi Tian, Yangrui Gong, Ruibo Hou, Yue Xu, Zhenbang Wu, Zhiyuan Fan, Yiyang Zhou, Kangyu Zhu, Wenhao Zheng, Zhaoyang Wang, Xiao Wang, Xuchao Zhang, Chetan Bansal, Marc Niethammer, Junzhou Huang, Hongtu Zhu, Yun Li, Jimeng Sun, Zongyuan Ge, Gang Li, James Zou, and Huaxiu Yao.
\newblock Cares: A comprehensive benchmark of trustworthiness in medical vision language models, 2024{\natexlab{a}}.
\newblock URL \url{https://arxiv.org/abs/2406.06007}.

\bibitem[Xia et~al.(2024{\natexlab{b}})Xia, Chen, Tian, Gong, Hou, Xu, Wu, Fan, Zhou, Zhu, et~al.]{xia2024cares}
Peng Xia, Ze~Chen, Juanxi Tian, Yangrui Gong, Ruibo Hou, Yue Xu, Zhenbang Wu, Zhiyuan Fan, Yiyang Zhou, Kangyu Zhu, et~al.
\newblock Cares: A comprehensive benchmark of trustworthiness in medical vision language models.
\newblock \emph{Advances in Neural Information Processing Systems}, 37:\penalty0 140334--140365, 2024{\natexlab{b}}.

\bibitem[Xia et~al.(2024{\natexlab{c}})Xia, Zhu, Li, Zhu, Li, Li, Zhang, and Yao]{xia2024rule}
Peng Xia, Kangyu Zhu, Haoran Li, Hongtu Zhu, Yun Li, Gang Li, Linjun Zhang, and Huaxiu Yao.
\newblock Rule: Reliable multimodal rag for factuality in medical vision language models.
\newblock In \emph{Proceedings of the 2024 Conference on Empirical Methods in Natural Language Processing}, pages 1081--1093, 2024{\natexlab{c}}.

\bibitem[Xia et~al.(2025)Xia, Wang, Peng, Zeng, Dong, Wu, Tang, Zhu, Li, Zhang, et~al.]{xia2025mmedagent}
Peng Xia, Jinglu Wang, Yibo Peng, Kaide Zeng, Zihan Dong, Xian Wu, Xiangru Tang, Hongtu Zhu, Yun Li, Linjun Zhang, et~al.
\newblock Mmedagent-rl: Optimizing multi-agent collaboration for multimodal medical reasoning.
\newblock \emph{arXiv preprint arXiv:2506.00555}, 2025.

\bibitem[Xiang et~al.(2024)Xiang, Zheng, Li, Hong, Li, Xie, Zhang, Xiong, Xie, Yang, et~al.]{xiang2024guardagent}
Zhen Xiang, Linzhi Zheng, Yanjie Li, Junyuan Hong, Qinbin Li, Han Xie, Jiawei Zhang, Zidi Xiong, Chulin Xie, Carl Yang, et~al.
\newblock Guardagent: Safeguard llm agents by a guard agent via knowledge-enabled reasoning.
\newblock \emph{arXiv preprint arXiv:2406.09187}, 2024.

\bibitem[Xiao et~al.(2025)Xiao, Huang, He, Xiao, Mousavi, Liu, Li, Chen, and Zhang]{xiao2025amqa}
Ying Xiao, Jie Huang, Ruijuan He, Jing Xiao, Mohammad~Reza Mousavi, Yepang Liu, Kezhi Li, Zhenpeng Chen, and Jie~M Zhang.
\newblock Amqa: An adversarial dataset for benchmarking bias of llms in medicine and healthcare.
\newblock \emph{arXiv preprint arXiv:2505.19562}, 2025.

\bibitem[Xie et~al.(2026)Xie, Han, Xiao, Cui, Wu, Zhang, Shu, Lu, Hu, and Yang]{xie2026ehrbench}
Yuzhang Xie, Keqi Han, Yunpeng Xiao, Hejie Cui, Guanchen Wu, Ziyang Zhang, Kai Shu, Jiaying Lu, Xiao Hu, and Carl Yang.
\newblock Ehrbench: An automated and reliable ehr-based benchmark for clinical decision making with llms.
\newblock \emph{arXiv preprint arXiv:2605.30637}, 2026.

\bibitem[Xu et~al.(2025{\natexlab{a}})Xu, Li, Wang, Fan, Zheng, Shi, Fan, Song, and Yang]{xu2025multiagentreasoningsystemscollaborative}
Baixuan Xu, Chunyang Li, Weiqi Wang, Wei Fan, Tianshi Zheng, Haochen Shi, Tao Fan, Yangqiu Song, and Qiang Yang.
\newblock Towards multi-agent reasoning systems for collaborative expertise delegation: An exploratory design study, 2025{\natexlab{a}}.
\newblock URL \url{https://arxiv.org/abs/2505.07313}.

\bibitem[Xu et~al.(2023)Xu, Sun, Zheng, Geng, Zhao, Feng, Tao, and Jiang]{xu2023wizardlm}
Can Xu, Qingfeng Sun, Kai Zheng, Xiubo Geng, Pu~Zhao, Jiazhan Feng, Chongyang Tao, and Daxin Jiang.
\newblock Wizardlm: Empowering large language models to follow complex instructions.
\newblock \emph{arXiv e-prints}, pages arXiv--2304, 2023.

\bibitem[Xu et~al.(2025{\natexlab{b}})Xu, Zhuang, Zhong, Yu, Wang, Tang, Wu, Wang, Ruan, Yang, et~al.]{xu2025medagentgym}
Ran Xu, Yuchen Zhuang, Yishan Zhong, Yue Yu, Zifeng Wang, Xiangru Tang, Hang Wu, May~D Wang, Peifeng Ruan, Donghan Yang, et~al.
\newblock Medagentgym: A scalable agentic training environment for code-centric reasoning in biomedical data science.
\newblock \emph{arXiv preprint arXiv:2506.04405}, 2025{\natexlab{b}}.

\bibitem[Xu et~al.(2025{\natexlab{c}})Xu, Chan, Li, Aljunied, Yuan, Wang, Xiao, Chen, Liu, Li, et~al.]{xu2025lingshu}
Weiwen Xu, Hou~Pong Chan, Long Li, Mahani Aljunied, Ruifeng Yuan, Jianyu Wang, Chenghao Xiao, Guizhen Chen, Chaoqun Liu, Zhaodonghui Li, et~al.
\newblock Lingshu: A generalist foundation model for unified multimodal medical understanding and reasoning.
\newblock \emph{arXiv preprint arXiv:2506.07044}, 2025{\natexlab{c}}.

\bibitem[Xu et~al.(2026)Xu, Liang, Mei, Gao, Tan, and Zhang]{xu2026mem}
Wujiang Xu, Zujie Liang, Kai Mei, Hang Gao, Juntao Tan, and Yongfeng Zhang.
\newblock A-mem: Agentic memory for llm agents.
\newblock \emph{Advances in Neural Information Processing Systems}, 38:\penalty0 17577--17604, 2026.

\bibitem[Yan et~al.(2026)Yan, Liu, Wu, Chen, Wang, Chai, and Wang]{yan2026clinicallab}
Weixiang Yan, Haitian Liu, Tengxiao Wu, Qian Chen, Wen Wang, Haoyuan Chai, and Jiayi Wang.
\newblock Clinicallab: Aligning agents for multi-departmental clinical diagnostics in the real world.
\newblock \emph{Advances in Neural Information Processing Systems}, 38, 2026.

\bibitem[Yang et~al.(2026{\natexlab{a}})Yang, Zhou, Xiao, Dong, Zhuang, Zhang, Wang, Hong, Yuan, Xiang, et~al.]{yang2026graph}
Chang Yang, Chuang Zhou, Yilin Xiao, Su~Dong, Luyao Zhuang, Yujing Zhang, Zhu Wang, Zijin Hong, Zheng Yuan, Zhishang Xiang, et~al.
\newblock Graph-based agent memory: Taxonomy, techniques, and applications.
\newblock \emph{arXiv preprint arXiv:2602.05665}, 2026{\natexlab{a}}.

\bibitem[Yang et~al.(2024{\natexlab{a}})Yang, Wang, Lu, Liu, Le, Zhou, and Chen]{yang2024large}
Chengrun Yang, Xuezhi Wang, Yifeng Lu, Hanxiao Liu, Quoc~V Le, Denny Zhou, and Xinyun Chen.
\newblock Large language models as optimizers.
\newblock In \emph{International Conference on Learning Representations}, volume 2024, pages 12028--12068, 2024{\natexlab{a}}.

\bibitem[Yang et~al.(2017)Yang, Cambias, Cleary, Daimler, Drake, Dupont, Hata, Kazanzides, Martel, Patel, et~al.]{yang2017medical}
Guang-Zhong Yang, James Cambias, Kevin Cleary, Eric Daimler, James Drake, Pierre~E Dupont, Nobuhiko Hata, Peter Kazanzides, Sylvain Martel, Rajni~V Patel, et~al.
\newblock Medical robotics—regulatory, ethical, and legal considerations for increasing levels of autonomy, 2017.

\bibitem[Yang et~al.(2024{\natexlab{b}})Yang, Jimenez, Wettig, Lieret, Yao, Narasimhan, and Press]{yang2024swe}
John Yang, Carlos Jimenez, Alexander Wettig, Kilian Lieret, Shunyu Yao, Karthik Narasimhan, and Ofir Press.
\newblock Swe-agent: Agent-computer interfaces enable automated software engineering.
\newblock \emph{Advances in Neural Information Processing Systems}, 37:\penalty0 50528--50652, 2024{\natexlab{b}}.

\bibitem[Yang et~al.(2025{\natexlab{a}})Yang, Zhou, Qi, Zhen, Sun, Shi, Su, and Yang]{yang2025aligning}
Lingrui Yang, Yuxing Zhou, Jun Qi, Xiantong Zhen, Li~Sun, Shan Shi, Qinghua Su, and Xuedong Yang.
\newblock Aligning large language models with radiologists by reinforcement learning from ai feedback for chest ct reports.
\newblock \emph{European Journal of Radiology}, 184:\penalty0 111984, 2025{\natexlab{a}}.

\bibitem[Yang et~al.(2025{\natexlab{b}})Yang, Wang, Liu, Sun, Wang, Chellappa, Zhou, Yuille, Zhu, Zhang, et~al.]{yang2025medical}
Yijun Yang, Zhao-Yang Wang, Qiuping Liu, Shuwen Sun, Kang Wang, Rama Chellappa, Zongwei Zhou, Alan Yuille, Lei Zhu, Yu-Dong Zhang, et~al.
\newblock Medical world model: Generative simulation of tumor evolution for treatment planning.
\newblock \emph{arXiv preprint arXiv:2506.02327}, 2025{\natexlab{b}}.

\bibitem[Yang et~al.(2023)Yang, Li, Wang, Lin, Azarnasab, Ahmed, Liu, Liu, Zeng, and Wang]{yang2023mm}
Zhengyuan Yang, Linjie Li, Jianfeng Wang, Kevin Lin, Ehsan Azarnasab, Faisal Ahmed, Zicheng Liu, Ce~Liu, Michael Zeng, and Lijuan Wang.
\newblock Mm-react: Prompting chatgpt for multimodal reasoning and action.
\newblock \emph{arXiv preprint arXiv:2303.11381}, 2023.

\bibitem[Yang et~al.(2026{\natexlab{b}})Yang, Janghorbani, Zhang, Han, Qian, Ressler~II, Lyng, Batra, and Tillman]{yang2026health}
Zhichao Yang, Sepehr Janghorbani, Dongxu Zhang, Jun Han, Qian Qian, Andrew Ressler~II, Gregory~D Lyng, Sanjit~Singh Batra, and Robert~E Tillman.
\newblock Health-score: Towards scalable rubrics for improving health-llms.
\newblock \emph{arXiv preprint arXiv:2601.18706}, 2026{\natexlab{b}}.

\bibitem[Yang et~al.(2025{\natexlab{c}})Yang, Luo, Han, Xu, and Li]{yang2025mitigating}
Zhihe Yang, Xufang Luo, Dongqi Han, Yunjian Xu, and Dongsheng Li.
\newblock Mitigating hallucinations in large vision-language models via dpo: On-policy data hold the key.
\newblock In \emph{Proceedings of the IEEE/CVF Conference on Computer Vision and Pattern Recognition}, pages 10610--10620, 2025{\natexlab{c}}.

\bibitem[Yao et~al.(2022)Yao, Zhao, Yu, Du, Shafran, Narasimhan, and Cao]{yao2022react}
Shunyu Yao, Jeffrey Zhao, Dian Yu, Nan Du, Izhak Shafran, Karthik Narasimhan, and Yuan Cao.
\newblock React: Synergizing reasoning and acting in language models.
\newblock \emph{arXiv preprint arXiv:2210.03629}, 2022.

\bibitem[Yao et~al.(2023)Yao, Yu, Zhao, Shafran, Griffiths, Cao, and Narasimhan]{yao2023tree}
Shunyu Yao, Dian Yu, Jeffrey Zhao, Izhak Shafran, Tom Griffiths, Yuan Cao, and Karthik Narasimhan.
\newblock Tree of thoughts: Deliberate problem solving with large language models.
\newblock \emph{Advances in neural information processing systems}, 36:\penalty0 11809--11822, 2023.

\bibitem[Ye and Tang(2025)]{ye2025multimodal}
Jiarui Ye and Hao Tang.
\newblock Multimodal large language models for medicine: A comprehensive survey.
\newblock \emph{arXiv preprint arXiv:2504.21051}, 2025.

\bibitem[Yi et~al.(2025)Yi, Liu, Albert, and Xiao]{yi2025multi}
Ziruo Yi, Jinyu Liu, Mark~V Albert, and Ting Xiao.
\newblock A multi-agent system for complex reasoning in radiology visual question answering.
\newblock In \emph{2025 ACM/IEEE Joint Conference on Digital Libraries (JCDL)}, pages 139--147. IEEE, 2025.

\bibitem[Yin et~al.(2024)Yin, Wang, Pan, Lin, Wan, and Wang]{yin2024g}
Xunjian Yin, Xinyi Wang, Liangming Pan, Li~Lin, Xiaojun Wan, and William~Yang Wang.
\newblock G$\backslash$" odel agent: A self-referential agent framework for recursive self-improvement.
\newblock \emph{arXiv preprint arXiv:2410.04444}, 2024.

\bibitem[Yu et~al.(2025)Yu, Yao, Liu, Chen, Yin, Wang, Liao, Ye, Li, Yue, et~al.]{yu2025medresearcher}
Ailing Yu, Lan Yao, Jingnan Liu, Zhe Chen, Jiajun Yin, Yuan Wang, Xinhao Liao, Zhiling Ye, Ji~Li, Yun Yue, et~al.
\newblock Medresearcher-r1: Expert-level medical deep researcher via a knowledge-informed trajectory synthesis framework.
\newblock \emph{arXiv preprint arXiv:2508.14880}, 2025.

\bibitem[Yu et~al.(2023)Yu, Endo, Krishnan, Pan, Tsai, Reis, Fonseca, Lee, Shakeri, Ng, et~al.]{yu2023radiology}
Feiyang Yu, Mark Endo, Rayan Krishnan, Ian Pan, Andy Tsai, Eduardo~Pontes Reis, EKU Fonseca, Henrique Lee, Zahra Shakeri, Andrew Ng, et~al.
\newblock Radiology report expert evaluation (rexval) dataset, 2023.

\bibitem[Yu et~al.(2026{\natexlab{a}})Yu, Zhang, Zhu, Yuan, Zuo, Yue, Dai, Fan, Liu, Liu, et~al.]{yu2026dapo}
Qiying Yu, Zheng Zhang, Ruofei Zhu, Yufeng Yuan, Xiaochen Zuo, Yu~Yue, Weinan Dai, Tiantian Fan, Gaohong Liu, Lingjun Liu, et~al.
\newblock Dapo: An open-source llm reinforcement learning system at scale.
\newblock \emph{Advances in Neural Information Processing Systems}, 38:\penalty0 113222--113244, 2026{\natexlab{a}}.

\bibitem[Yu et~al.(2026{\natexlab{b}})Yu, Yao, Xie, Tan, Feng, Li, and Wu]{yu2026agenticmemorylearningunified}
Yi~Yu, Liuyi Yao, Yuexiang Xie, Qingquan Tan, Jiaqi Feng, Yaliang Li, and Libing Wu.
\newblock Agentic memory: Learning unified long-term and short-term memory management for large language model agents, 2026{\natexlab{b}}.
\newblock URL \url{https://arxiv.org/abs/2601.01885}.

\bibitem[Yue et~al.(2026)Yue, Bhandari, Ko, Patel, Lin, Zhou, Gao, Chen, and Pan]{yue2026static}
Ling Yue, Kushal~Raj Bhandari, Ching-Yun Ko, Dhaval Patel, Shuxin Lin, Nianjun Zhou, Jianxi Gao, Pin-Yu Chen, and Shaowu Pan.
\newblock From static templates to dynamic runtime graphs: a survey of workflow optimization for llm agents.
\newblock \emph{arXiv preprint arXiv:2603.22386}, 2026.

\bibitem[Yuksekgonul et~al.(2024)Yuksekgonul, Bianchi, Boen, Liu, Huang, Guestrin, and Zou]{yuksekgonul2024textgrad}
Mert Yuksekgonul, Federico Bianchi, Joseph Boen, Sheng Liu, Zhi Huang, Carlos Guestrin, and James Zou.
\newblock Textgrad: Automatic" differentiation" via text.
\newblock \emph{arXiv preprint arXiv:2406.07496}, 2024.

\bibitem[Zeng et~al.(2024)Zeng, Lyu, Li, and Li]{radcouncil2024}
Fang Zeng, Zhiliang Lyu, Quanzheng Li, and Xiang Li.
\newblock Enhancing llms for impression generation in radiology reports through a multi-agent system.
\newblock \emph{arXiv preprint arXiv:2412.06828}, 2024.

\bibitem[Zeng et~al.(2025)Zeng, Wei, Brown, Frunza, Nevmyvaka, Zhao, and Hong]{zeng2025reinforcing}
Siliang Zeng, Quan Wei, William Brown, Oana Frunza, Yuriy Nevmyvaka, Yang~Katie Zhao, and Mingyi Hong.
\newblock Reinforcing multi-turn reasoning in llm agents via turn-level credit assignment.
\newblock In \emph{ICML 2025 Workshop on Computer Use Agents}, 2025.

\bibitem[Zhang et~al.(2025{\natexlab{a}})Zhang, Lazuka, and Murag]{zhang2025equipping}
Barry Zhang, Keith Lazuka, and Mahesh Murag.
\newblock Equipping agents for the real world with agent skills.
\newblock \emph{Anthropic Engineering Blog}, 2025{\natexlab{a}}.

\bibitem[Zhang et~al.(2025{\natexlab{b}})Zhang, Geng, Yu, Yin, Zhang, Tan, Zhou, Li, Xue, Li, et~al.]{zhang2025landscape}
Guibin Zhang, Hejia Geng, Xiaohang Yu, Zhenfei Yin, Zaibin Zhang, Zelin Tan, Heng Zhou, Zhongzhi Li, Xiangyuan Xue, Yijiang Li, et~al.
\newblock The landscape of agentic reinforcement learning for llms: A survey.
\newblock \emph{arXiv preprint arXiv:2509.02547}, 2025{\natexlab{b}}.

\bibitem[Zhang et~al.(2025{\natexlab{c}})Zhang, Niu, Fang, Wang, Bai, and Wang]{zhang2025multi}
Guibin Zhang, Luyang Niu, Junfeng Fang, Kun Wang, Lei Bai, and Xiang Wang.
\newblock Multi-agent architecture search via agentic supernet.
\newblock \emph{arXiv preprint arXiv:2502.04180}, 2025{\natexlab{c}}.

\bibitem[Zhang et~al.(2025{\natexlab{d}})Zhang, Liu, Lv, Sun, Jing, Iong, Hou, Qi, Lai, Xu, et~al.]{zhang2025agentrl}
Hanchen Zhang, Xiao Liu, Bowen Lv, Xueqiao Sun, Bohao Jing, Iat~Long Iong, Zhenyu Hou, Zehan Qi, Hanyu Lai, Yifan Xu, et~al.
\newblock Agentrl: Scaling agentic reinforcement learning with a multi-turn, multi-task framework.
\newblock \emph{arXiv preprint arXiv:2510.04206}, 2025{\natexlab{d}}.

\bibitem[Zhang et~al.(2026{\natexlab{a}})Zhang, Hu, Lu, Lange, and Clune]{zhang2026darwin}
Jenny Zhang, Shengran Hu, Cong Lu, Robert Lange, and Jeff Clune.
\newblock Darwin g{\"o}del machine: open-ended evolution of self-improving agents.
\newblock In \emph{International Conference on Learning Representations}, volume 2026, pages 104223--104294, 2026{\natexlab{a}}.

\bibitem[Zhang et~al.(2025{\natexlab{e}})Zhang, Barrett, Kim, Sun, Taghavi, and Kenthapadi]{zhang2025radagents}
Kai Zhang, Corey~D Barrett, Jangwon Kim, Lichao Sun, Tara Taghavi, and Krishnaram Kenthapadi.
\newblock Radagents: Multimodal agentic reasoning for chest x-ray interpretation with radiologist-like workflows.
\newblock \emph{arXiv preprint arXiv:2509.20490}, 2025{\natexlab{e}}.

\bibitem[Zhang et~al.(2025{\natexlab{f}})Zhang, Hu, Upasani, Ma, Hong, Kamanuru, Rainton, Wu, Ji, Li, et~al.]{zhang2025agentic}
Qizheng Zhang, Changran Hu, Shubhangi Upasani, Boyuan Ma, Fenglu Hong, Vamsidhar Kamanuru, Jay Rainton, Chen Wu, Mengmeng Ji, Hanchen Li, et~al.
\newblock Agentic context engineering: Evolving contexts for self-improving language models.
\newblock \emph{arXiv preprint arXiv:2510.04618}, 2025{\natexlab{f}}.

\bibitem[Zhang et~al.(2026{\natexlab{b}})Zhang, Hu, Upasani, Ma, Hong, Kamanuru, Rainton, Wu, Ji, Li, et~al.]{zhang2026agentic}
Qizheng Zhang, Changran Hu, Shubhangi Upasani, Boyuan Ma, Fenglu Hong, Vamsidhar Kamanuru, Jay Rainton, Chen Wu, Mengmeng Ji, Hanchen Li, et~al.
\newblock Agentic context engineering: Evolving contexts for self-improving language models.
\newblock In \emph{International Conference on Learning Representations}, volume 2026, pages 86069--86100, 2026{\natexlab{b}}.

\bibitem[Zhang et~al.(2023)Zhang, Xu, Usuyama, Bagber, Tinn, Preston, Dragan, Mueen, Naumann, Wong, et~al.]{zhang2023biomedclip}
Sheng Zhang, Yanbo Xu, Naoto Usuyama, Jaspreet Bagber, Robert Tinn, Sam Preston, Rajesh Dragan, Abdullah Mueen, Tristan Naumann, Cliff Wong, et~al.
\newblock Biomedclip: A multimodal biomedical foundation model pretrained from fifteen million scientific image-text pairs, 2023.

\bibitem[Zhang et~al.(2026{\natexlab{c}})Zhang, Guo, Zhang, Zhang, Chen, Zhang, Zhang, Yi, and Bu]{zhang2026patho}
Wenchuan Zhang, Jingru Guo, Hengzhe Zhang, Penghao Zhang, Jie Chen, Shuwan Zhang, Zhang Zhang, Yuhao Yi, and Hong Bu.
\newblock Patho-agenticrag: towards multimodal agentic retrieval-augmented generation for pathology vlms via reinforcement learning.
\newblock In \emph{Proceedings of the AAAI Conference on Artificial Intelligence}, volume~40, pages 29921--29929, 2026{\natexlab{c}}.

\bibitem[Zhang et~al.(2026{\natexlab{d}})Zhang, Zhang, Guo, Cheng, Chen, Zhang, Zhang, Yi, and Bu]{zhang2026pathor1}
Wenchuan Zhang, Penghao Zhang, Jingru Guo, Tao Cheng, Jie Chen, Shuwan Zhang, Zhang Zhang, Yuhao Yi, and Hong Bu.
\newblock Patho-r1: A multimodal reinforcement learning-based pathology expert reasoner.
\newblock In \emph{Proceedings of the AAAI Conference on Artificial Intelligence}, volume~40, pages 28418--28426, 2026{\natexlab{d}}.

\bibitem[Zhang et~al.(2025{\natexlab{g}})Zhang, Li, Zhou, and Liu]{zhang2025r2med}
Xiangxu Zhang, Lei Li, Xiao Zhou, and Zheng Liu.
\newblock R2med: A benchmark for reasoning-driven medical retrieval.
\newblock \emph{arXiv preprint arXiv:2505.14558}, 2025{\natexlab{g}}.

\bibitem[Zhao et~al.(2024)Zhao, Huang, Xu, Lin, Liu, and Huang]{zhao2024expel}
Andrew Zhao, Daniel Huang, Quentin Xu, Matthieu Lin, Yong-Jin Liu, and Gao Huang.
\newblock Expel: Llm agents are experiential learners.
\newblock In \emph{Proceedings of the AAAI Conference on Artificial Intelligence}, volume~38, pages 19632--19642, 2024.

\bibitem[Zhao et~al.(2026)Zhao, Wu, Wu, Xu, Yue, Lin, Wang, Wu, Zheng, and Huang]{zhao2026absolute}
Andrew Zhao, Yiran Wu, Tong Wu, Quentin Xu, Yang Yue, Matthieu Lin, Shenzhi Wang, Qingyun Wu, Zilong Zheng, and Gao Huang.
\newblock Absolute zero: Reinforced self-play reasoning with zero data.
\newblock \emph{Advances in Neural Information Processing Systems}, 38:\penalty0 105816--105879, 2026.

\bibitem[Zheng et~al.(2025)Zheng, Liu, Li, Chen, Yu, Gao, Dang, Liu, Men, Yang, et~al.]{zheng2025group}
Chujie Zheng, Shixuan Liu, Mingze Li, Xiong-Hui Chen, Bowen Yu, Chang Gao, Kai Dang, Yuqiong Liu, Rui Men, An~Yang, et~al.
\newblock Group sequence policy optimization.
\newblock \emph{arXiv preprint arXiv:2507.18071}, 2025.

\bibitem[Zheng et~al.(2024)Zheng, Wang, Wang, and An]{zheng2024synapse}
Longtao Zheng, Rundong Wang, Xinrun Wang, and Bo~An.
\newblock Synapse: Trajectory-as-exemplar prompting with memory for computer control.
\newblock In \emph{International Conference on Learning Representations}, volume 2024, pages 19036--19066, 2024.

\bibitem[Zhong et~al.(2026)Zhong, Xia, Zhang, Yue, Xia, Shi, et~al.]{zhong2026neuroagent}
Lujia Zhong, Yihao Xia, Jianwei Zhang, Jiaxin Yue, Mingyang Xia, Yonggang Shi, et~al.
\newblock Neuroagent: Llm agents for multimodal neuroimaging analysis and research.
\newblock \emph{arXiv preprint arXiv:2605.06584}, 2026.

\bibitem[Zhou et~al.(2022)Zhou, Sch{\"a}rli, Hou, Wei, Scales, Wang, Schuurmans, Cui, Bousquet, Le, et~al.]{zhou2022least}
Denny Zhou, Nathanael Sch{\"a}rli, Le~Hou, Jason Wei, Nathan Scales, Xuezhi Wang, Dale Schuurmans, Claire Cui, Olivier Bousquet, Quoc Le, et~al.
\newblock Least-to-most prompting enables complex reasoning in large language models.
\newblock \emph{arXiv preprint arXiv:2205.10625}, 2022.

\bibitem[Zhou et~al.(2024)Zhou, Pujara, Ren, Chen, Cheng, Le, Zhou, Mishra, Zheng, et~al.]{zhou2024self}
Pei Zhou, Jay Pujara, Xiang Ren, Xinyun Chen, Heng-Tze Cheng, Quoc~V Le, Denny Zhou, Swaroop Mishra, Huaixiu~S Zheng, et~al.
\newblock Self-discover: Large language models self-compose reasoning structures.
\newblock \emph{Advances in Neural Information Processing Systems}, 37:\penalty0 126032--126058, 2024.

\bibitem[Zhou et~al.(2026)Zhou, Liang, Yang, Chen, Wu, Yao, and Wang]{zhou2026escrl}
Qin Zhou, Guoyan Liang, Qianyi Yang, Jingyuan Chen, Sai Wu, Chang Yao, and Zhe Wang.
\newblock Enhancing reinforcement learning for radiology report generation with evidence-aware rewards and self-correcting preference learning, 2026.
\newblock URL \url{https://arxiv.org/abs/2604.13598}.

\bibitem[Zhou et~al.(2025)Zhou, Song, and Shen]{zhou2025mam}
Yucheng Zhou, Lingran Song, and Jianbing Shen.
\newblock Mam: Modular multi-agent framework for multi-modal medical diagnosis via role-specialized collaboration.
\newblock In \emph{Findings of the Association for Computational Linguistics: ACL 2025}, pages 25319--25333, 2025.

\bibitem[Zhu et~al.(2026{\natexlab{a}})Zhu, Lin, Chen, Wang, and Lin]{zhu2026medeyes}
Chunzheng Zhu, Yangfang Lin, Shen Chen, Yijun Wang, and Jianxin Lin.
\newblock Medeyes: Learning dynamic visual focus for medical progressive diagnosis.
\newblock In \emph{Proceedings of the AAAI Conference on Artificial Intelligence}, volume~40, pages 13916--13924, 2026{\natexlab{a}}.

\bibitem[Zhu et~al.(2026{\natexlab{b}})Zhu, Zeng, Jiang, Lin, and Wang]{zhu2026medsynapse}
Chunzheng Zhu, Jiaqi Zeng, Junyu Jiang, Jianxin Lin, and Yijun Wang.
\newblock Medsynapse-v: Bridging visual perception and clinical intuition via latent memory evolution.
\newblock \emph{arXiv preprint arXiv:2604.26283}, 2026{\natexlab{b}}.

\bibitem[Zhu et~al.(2025{\natexlab{a}})Zhu, Wang, Chen, Liu, Ye, Gu, Tian, Duan, Su, Shao, et~al.]{zhu2025internvl3}
Jinguo Zhu, Weiyun Wang, Zhe Chen, Zhaoyang Liu, Shenglong Ye, Lixin Gu, Hao Tian, Yuchen Duan, Weijie Su, Jie Shao, et~al.
\newblock Internvl3: Exploring advanced training and test-time recipes for open-source multimodal models.
\newblock \emph{arXiv preprint arXiv:2504.10479}, 2025{\natexlab{a}}.

\bibitem[Zhu et~al.(2025{\natexlab{b}})Zhu, Li, Wu, Xing, Ma, Tang, Liu, Yang, Liu, Jiang, et~al.]{zhu2025scaling}
King Zhu, Hanhao Li, Siwei Wu, Tianshun Xing, Dehua Ma, Xiangru Tang, Minghao Liu, Jian Yang, Jiaheng Liu, Yuchen~Eleanor Jiang, et~al.
\newblock Scaling test-time compute for llm agents.
\newblock \emph{arXiv preprint arXiv:2506.12928}, 2025{\natexlab{b}}.

\bibitem[Zhu et~al.(2026{\natexlab{c}})Zhu, He, Hu, Zheng, Zhang, Wang, Gao, Ma, and Yu]{zhu2026medagentboard}
Yinghao Zhu, Ziyi He, Haoran Hu, Xiaochen Zheng, Xichen Zhang, Wang Wang, Junyi Gao, Liantao Ma, and Lequan Yu.
\newblock Medagentboard: Benchmarking multi-agent collaboration with conventional methods for diverse medical tasks.
\newblock \emph{Advances in Neural Information Processing Systems}, 38, 2026{\natexlab{c}}.

\bibitem[Zhu et~al.(2026{\natexlab{d}})Zhu, Liu, Yu, and Zhang]{zhu2026llm}
Yiwen Zhu, Lihe Liu, Jiaqian Yu, and Di~Zhang.
\newblock Llm-based multi-agent orchestration: A survey of frameworks, communication protocols, and emerging patterns.
\newblock 2026{\natexlab{d}}.

\bibitem[Zhuge et~al.(2024)Zhuge, Wang, Kirsch, Faccio, Khizbullin, and Schmidhuber]{zhuge2024gptswarm}
Mingchen Zhuge, Wenyi Wang, Louis Kirsch, Francesco Faccio, Dmitrii Khizbullin, and J{\"u}rgen Schmidhuber.
\newblock Gptswarm: Language agents as optimizable graphs.
\newblock In \emph{Forty-first International Conference on Machine Learning}, 2024.

\bibitem[Zong et~al.(2026)Zong, Chen, Li, Yi, Zhou, Li, Qian, Chen, and Jiang]{zong20262}
Zefang Zong, Dingwei Chen, Yang Li, Qi~Yi, Bo~Zhou, Chengming Li, Bo~Qian, Peng Chen, and Jie Jiang.
\newblock At$^2$po: Agentic turn-based policy optimization via tree search.
\newblock \emph{arXiv preprint arXiv:2601.04767}, 2026.

\bibitem[Zou and Topol(2025)]{zou2025rise}
James Zou and Eric~J Topol.
\newblock The rise of agentic ai teammates in medicine.
\newblock \emph{The Lancet}, 405\penalty0 (10477):\penalty0 457, 2025.

\bibitem[Zuo and Jiang(2024)]{zuo2024medhallbench}
Kaiwen Zuo and Yirui Jiang.
\newblock Medhallbench: A new benchmark for assessing hallucination in medical large language models.
\newblock \emph{arXiv preprint arXiv:2412.18947}, 2024.

\bibitem[Zuo et~al.(2025)Zuo, Qu, Li, Chen, Zhu, Hua, Zhang, Ding, and Zhou]{zuo2025medxpertqa}
Yuxin Zuo, Shang Qu, Yifei Li, Zhangren Chen, Xuekai Zhu, Ermo Hua, Kaiyan Zhang, Ning Ding, and Bowen Zhou.
\newblock Medxpertqa: Benchmarking expert-level medical reasoning and understanding.
\newblock \emph{arXiv preprint arXiv:2501.18362}, 2025.

\bibitem[Zuo et~al.(2026)Zuo, Zhang, Sheng, Qu, Cui, Zhu, Li, Long, Hua, Qi, et~al.]{zuo2026ttrl}
Yuxin Zuo, Kaiyan Zhang, Li~Sheng, Shang Qu, Ganqu Cui, Xuekai Zhu, Haozhan Li, Xinwei Long, Ermo Hua, Biqing Qi, et~al.
\newblock Ttrl: Test-time reinforcement learning.
\newblock \emph{Advances in Neural Information Processing Systems}, 38:\penalty0 131459--131483, 2026.

\bibitem[Zweiger et~al.(2026)Zweiger, Pari, Guo, Kim, and Agrawal]{zweiger2026self}
Adam Zweiger, Jyo Pari, Han Guo, Yoon Kim, and Pulkit Agrawal.
\newblock Self-adapting language models.
\newblock \emph{Advances in Neural Information Processing Systems}, 38:\penalty0 74084--74115, 2026.

\end{thebibliography}

\end{document}